\documentclass[sigconf]{acmart}
\renewcommand\footnotetextcopyrightpermission[1]{}
\usepackage{lipsum}
\usepackage{multirow}
\usepackage{makecell}
\usepackage{subfig}
\usepackage{graphicx}
\usepackage{booktabs}
\usepackage{hyperref}
\usepackage{amsmath}
\usepackage{color}

\AtBeginDocument{%
	\providecommand\BibTeX{{%
			\normalfont B\kern-0.5em{\scshape i\kern-0.25em b}\kern-0.8em\TeX}}}

\newcommand{\bheading}[1]{{\vspace{2pt}\noindent{\textbf{#1}}\hspace{2pt}}} 
\newenvironment{packeditemize}{
	\begin{list}{$\bullet$}{
			\setlength{\labelwidth}{8pt}
			\setlength{\itemsep}{0pt}
			\setlength{\leftmargin}{\labelwidth}
			\addtolength{\leftmargin}{\labelsep}
			\setlength{\parindent}{0pt}
			\setlength{\listparindent}{\parindent}
			\setlength{\parsep}{0pt}
			\setlength{\topsep}{3pt}}}{\end{list}}

\begin{document}

\title{Can We Mitigate Backdoor Attack Using Adversarial Detection Methods?}

	\author{Kaidi Jin$^*$, Tianwei Zhang$^\mathsection$,  Chao Shen$^*$, Yufei Chen$^*$,  Ming Fan$^*$, Chenhao Lin$^*$,  Ting Liu$^*$
	}
	
	\affiliation{%
		Xi'an Jiaotong University$^*$ \hspace{0.3em}Nanyang Technological University$^\mathsection$
	}

	\affiliation{%
	\institution{jinkaidi@stu.xjtu.edu.cn,\hspace{0.2em} tianwei.zhang@ntu.edu.sg,\hspace{0.2em} chaoshen@mail.xjtu.edu.cn}
	}
	\affiliation{%
	\institution{yfchen@sei.xjtu.edu.cn,\hspace{0.2em}mingfan@mail.xjtu.edu.cn,\hspace{0.2em}linchenhao@xjtu.edu.cn,\hspace{0.2em}tingliu@mail.xjtu.edu.cn}
	}

\begin{abstract}
	Deep Neural Networks are well known to be vulnerable to adversarial attacks and backdoor attacks, where minor modifications on the input are able to mislead the models to give wrong results. Although defenses against adversarial attacks have been widely studied, investigation on mitigating backdoor attacks is still at an early stage. It is unknown whether there are any connections and common characteristics between the defenses against these two attacks. We conduct comprehensive studies on the connections between adversarial examples and backdoor examples of Deep Neural Networks to seek to answer the question: can we detect backdoor using adversarial detection methods. Our insights are based on the observation that both adversarial examples and backdoor examples have anomalies during the inference process, highly distinguishable from benign samples. As a result, we revise four existing adversarial defense methods for detecting backdoor examples. Extensive evaluations indicate that these approaches provide reliable protection against backdoor attacks, with a higher accuracy than detecting adversarial examples. These solutions also reveal the relations of adversarial examples, backdoor examples and normal samples in model sensitivity, activation space and feature space. This is able to enhance our understanding about the inherent features of these two attacks and the defense opportunities.
\end{abstract}


\maketitle

\section{Introduction}

Past years have witnessed the rapid development of Deep Learning (DL) technology. State-of-the-art Deep Neural Networks (DNNs) can outperform conventional machine learning models in many artificial intelligence tasks, such as image classification~\cite{image_class, face_recognition}, speech recognition~\cite{speech_recognition}, natural language processing~\cite{luong2015effective}. The high and reliable performance of DNNs is attributed to the models' complex structures and large numbers of parameters. 

\begin{figure*}[t]
		\centering

			\begin{minipage}[t]{1.0\linewidth}
				\label{adver_attack}
				\centering
				\includegraphics[width=\textwidth]{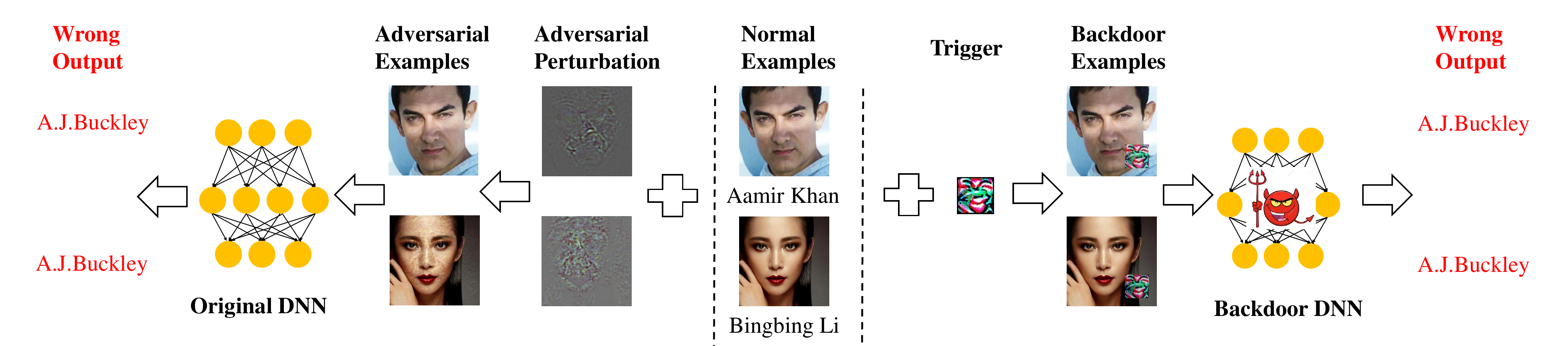}            
			\end{minipage}
		\caption{Illustration of an adversarial attack (left part) and backdoor attack (right part) on a DNN model for face recognition.}
		\label{fig:face_attack}    
	\end{figure*}

However, such model complexity also brings security vulnerabilities, which can be exploited by adversaries to compromise the DNN applications. Two typical examples are adversarial attacks~\cite{adver_attack} and backdoor attacks~\cite{badnets} (Figure \ref{fig:face_attack}). In both types of attacks, the adversary injects carefully-crafted perturbations on the input samples to fool the DNN models. In adversarial attacks, the perturbation is specifically generated for each sample to mislead the target model. In backdoor attacks, the adversary produces a universal perturbation (i.e., trigger), and modifies the target model correspondingly to misclassify each sample with the trigger. These attacks have significantly threatened the DNN applications, especially in the safety- and security-critical scenarios, e.g., autonomous driving~\cite{self_drive}, malware detection~\cite{malware_detection,fan2018android,fan2019ctdroid,fan2016frequent,fan2019graph}, user authentication~\cite{chen2017targeted}, and medical diagnosis~\cite{finlayson2018adversarial}.

Extensive studies have been conducted to mitigate adversarial attacks~\cite{adver_train,huang2015learning,gu2014towards,ross2018improving,defen_distillation,Magnet,lee2017generative,guo2018countering,prakash2018deflecting,xie2018mitigating,buckman2018thermometer,das2018shield}. In contrast, there are fewer satisfactory solutions against backdoor attacks. Most works~\cite{Neural_cleanse,Deepinspect,defen_Tabor,defen_GDM,defen_MNA,ABS} attempted to detect and remove malicious backdoor in the target models. However, due to the defender's limited knowledge about the attack techniques and configurations, those methods can only be applied to simple backdoor attacks (e.g., one targeted class, simple trigger pattern), and they can be easily evaded by adaptive attacks~\cite{tan2019bypassing}. Other approaches aim to identify poisoned data in the training set~\cite{activation_clustering,Spectral_signatures,Differential_Privacy}. They are not applicable when the defender has no access to the training data. 

In this paper, we focus on the mitigation of backdoor attacks in a different direction: detecting backdoor samples at the inference phase. With such protection, all malicious samples will be ruled out, and the compromised models will still give correct prediction results for normal samples. Achieving this goal is challenging as the triggers can have arbitrary sizes and patterns, which are agnostic to the defender. Existing detection solutions either are limited to simple triggers~\cite{Strip,Sentinet} or require priori knowledge about the triggers~\cite{Grad_cam}, making them less practical.  

Our proposed strategy is based on two insights. The first one is that \emph{there exist some similarities between adversarial examples and backdoor examples}. Both of them require stealthy modifications to enforce wrong prediction output. As such, they exhibit certain anomaly during the inference process, and can be detected in a similar way. Based on this observation, we can apply the methodologies of detecting adversarial examples to backdoor example detection. We identify four effective approaches to distinguish backdoor examples from normal samples based on their 
model sensitivities, behaviors in the feature space and activation space. 

The second insight is that \emph{adversarial examples and backdoor examples have certain differences caused by attack attributes}.  To meet the universality requirement, backdoor examples need larger scale of perturbations, making them further from the model decision boundary and normal samples. As a result, we need to make some modifications on the methodology workflows and configurations to identify backdoor examples. Besides, due to those differences, we observe that these methodologies have a better accuracy of detecting backdoor examples than adversarial examples, even though they are originally designed to defeat adversarial attacks. 

In this paper, we describe the results of our comprehensive studies on the connections between adversarial examples and backdoor examples against DNNs. To the best of our knowledge, there are only two works~\cite{pang2020tale,weng2020trade} investigating the relations between the two kinds of samples, from the perspective of attacks.  We present the first systematic study from the defense perspective. We perform an in-depth analysis about the similarities as well as differences between adversarial examples, backdoor examples and normal samples. With such analysis, we identify four approaches originally designed for adversarial example detection, to detect backdoor threats. 
We are the first to show that adversarial examples and backdoor attacks can be defeated in a unified way. We provide thorough evaluations on these methodologies for defeating both adversarial and backdoor attacks, in terms of effectiveness, usability and performance. Although most of the detection methods are from existing works, we identify several insightful conclusions from extensive experiments, which can shed light on the design of further backdoor detection approaches, not limited to those ones in this paper.
	
Our main contributions are listed below:
	
	\renewcommand\theenumi{\roman{enumi}}
	\renewcommand\labelenumi{(\theenumi)}

	\begin{packeditemize}
		
		\item We present the first systematic study about the relations between adversarial examples and backdoor examples from the defense perspective. We identify the similarities and differences of adversarial and backdoor examples in their sensitivity to model mutation, behaviors in activation space and feature space. 
		\item We apply four detection methods from adversarial attacks to backdoor attacks, and achieve better detection accuracy.
		\item We conduct comprehensive evaluations on these methodologies for defeating both adversarial and backdoor attacks, in terms of effectiveness, usability and performance. 
		
	\end{packeditemize}

\section{Background and Related Works}
\subsection{Adversarial Attacks}
Formally, the target DNN model is denoted as a parameterized function $f_\theta: \mathcal{X} \mapsto \mathcal{Y}$ that maps an input tensor $x\in \mathcal{X}$ to an output tensor $y\in \mathcal{Y}$. Given a clean sample $x$, the adversary's goal is to find the corresponding adversarial example (AE) $\widetilde{x}=x+\delta$, such that $f_\theta$ will predict it as a different label. The adversarial perturbation $\delta$ should be kept as small as possible. AE generation can be formulated as the optimization problem in Equation \ref{eq:AE}. 

\begin{equation}
\label{eq:AE}
\begin{aligned}
& \text{minimize:}  \quad \lVert\delta\rVert\\
& \text{subject to:} \quad f_\theta(x+\delta)\neq f_\theta(x)
\end{aligned}
\end{equation}

Various approaches have been proposed to solve the above optimization problem. Szegedy et al.~\cite{adver_attack} adopted the L-BFGS algorithm to generate AEs. Then a couple of gradient-based methods were introduced to enhance the attack techniques: the gradient descent evasion attack~\cite{biggio2013evasion} calculated the gradients of neural networks to generate AEs; Fast Gradient Sign Method (FGSM)~\cite{FGSM} calculated the adversarial perturbation based on the sign of gradients, which was further improved by its iterative versions (I-FGSM~\cite{kurakin2016adversarial} and MI-FGSM~\cite{dong2017discovering}). Basic Iterative Method (BIM) \cite{BIM} iteratively applied FGSM with small perturbations to get the final AEs. Deepfool~\cite{moosavi2016deepfool} is another iterative method that outperforms previous attacks by searching for the optimal perturbation across the decision boundary. 
Jacobian-based Saliency Map Attack (JSMA)~\cite{JSMA} estimated the saliency map of pixels w.r.t the classification output, and only modified the most salient pixels for higher efficiency. One pixel attack~\cite{su2019one} is an extreme-case attack where only one pixel can be modified to fool the classifier. A more powerful attack, C\&W~\cite{CW_attack}, was proposed by updating the objective function to minimize $l_p$ distance between AEs and normal examples. 
C\&W can effectively defeat Defensive Distillation~\cite{defen_distillation} and other defenses with assisted models~\cite{CW_attack} with very high attack success rates. 

\noindent\textbf{Threat Model.}
	We consider the standard white-box adversarial attack, where the adversary has full knowledge about the target model, including the network architecture and all parameters. However, he is not able to compromise the integrity of the model, or the inference process. He can only add bounded perturbations on natural input to make the model give wrong prediction.
\subsection{Backdoor Attacks}

For a given DNN model $f_\theta$ with the parameters $\theta$, the adversary attempts to find backdoored parameters $\theta^*$ and a trigger $\delta$, such that the backdoor model $f_{\theta^*}$ can give correct results for all normal samples $x\in \mathcal{X}$, but predict the backdoor example (BE) $x+\delta$ as different labels. Similarly, backdoor attacks can also be formulated as an optimization problem, as shown in Equation \ref{eq:TE}. 

\begin{equation}
\label{eq:TE}
\begin{aligned}
& \text{minimize:}  \quad \lVert\delta\rVert\\
& \text{subject to:} \quad \forall x\in \mathcal{X}, f_{\theta^*}(x)= f_\theta(x) \\
& \quad\quad\quad\quad\quad\; \forall x\in \mathcal{X}, f_{\theta^*}(x+\delta) \neq f_\theta(x)
\end{aligned}
\end{equation}


Solving this optimization problem directly is difficult. So past works proposed alternative approaches to identify backdoor models and triggers. Badnets~\cite{badnets} adopted poisoning attack technique: the adversary first identifies the trigger pattern $\delta$. Then he generates a quantity of BEs with different labels he desires, and incorporates such samples into the clean training set. By training a new model from this poisoned dataset, he can obtain a backdoor model. Liu et al.~\cite{Trojan_attack} proposed an enhanced attack: the adversary can directly modify a set of neurons in the internal layer without the need to train models. Yao et al.~\cite{Latent_attack} studied the transferability feature of backdoor attacks: if the adversary injects backdoor into a teacher model, the student models transferred from this teacher model may still contain the backdoor, and be vulnerable to BEs. Most recently, Liu et al.~\cite{liu2020reflection} proposed a more powerful attack (referred to as \emph{invisible backdoor attacks}) in order to evade human inspection. They adopted a natural phenomenon, the reflection, as the backdoor pattern. 

\noindent\textbf{Threat Model.}
	We adopt the threat model in existing backdoor attack works. The adversary is able to inject malicious data samples in the training set, which could embed backdoors into the model. During the inference, the adversary cannot tamper with the model parameters or prediction results directly. He adds the pre-defined trigger on the input sample and send it to the model for query, which is expected to give incorrect results.

\subsection{Comparisons}
Adversarial attacks and backdoor attacks have some similarities, as well as distinct features. 
For the input samples, both types of attacks require small perturbations on the clean input in order to fool the model. Notice that there are large semantic backdoor triggers(i.e., the blending attack~\cite{chen2017targeted}) and large adversarial perturbations, e.g., style attack~\cite{duan2020adversarial} semantic attack~\cite{hosseini2018semantic} and unrestricted attack~\cite{bhattad2019unrestricted}.  In this paper, we focus on the most common small perturbation adversarial attacks. Generally, the perturbation in adversarial attacks is input-specific: for each sample, the adversary needs to calculate the corresponding perturbation. In contrast, the perturbation in backdoor attacks is universal. The trigger is fixed for all samples belonging to all classes\footnote{There are also some exceptions, e.g., universal adversarial attacks \cite{moosavi2017universal}, input-specific backdoor triggers ~\cite{nguyen2020input}}. 

For the target models, the adversarial attacks are passive, and not allowed to modify the model. Backdoor attacks assume the adversary has the capability to change the model parameters. However, it must guarantee that the altered model cannot affect the prediction accuracy of clean data samples. 

\begin{figure}[t]
		\centering
		\includegraphics[width=0.45\textwidth]{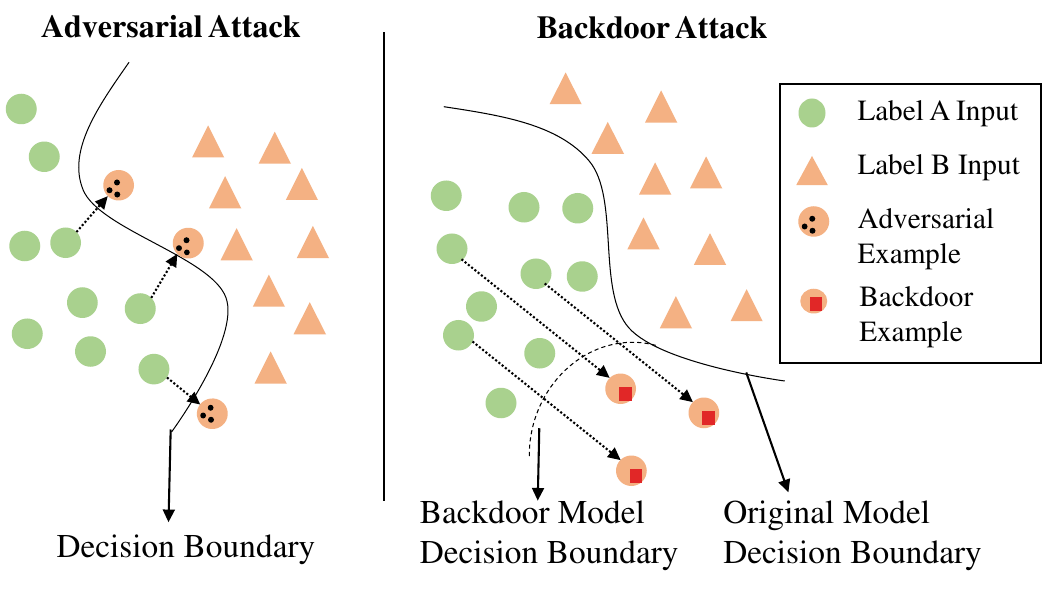}  
		\caption{Visualization of adversarial examples and backdoor examples with the model classification boundary.}
		\label{fig:two_attacks_boundary}    
		\vspace{-15pt}
	\end{figure}

Figure \ref{fig:two_attacks_boundary} visually shows the comparisons of two attack scenarios, with a two-class model. Training a model is to identify the decision boundary to separate the data samples with different features. Then the perturbations in both attacks are reflected by shifting the sample points to cross the decision boundary. The perturbation in adversarial attacks is input-specific. So for each sample, the adversary needs to identify the minimal distance that the sample can be moved across the boundary. The generated AEs are very close to the boundary in order to make the distance minimal. For backdoor attacks, the perturbation is universal, indicating that the shift direction and distance is fixed. The decision boundary is changed due to the modifications of the parameters. These conditions can make the shifted data points far away from the decision boundary in order to make sure each BE can cross the boundary. 

\subsection{Defenses}
\bheading{Mitigating adversarial attacks.}
Existing solutions can be classified into four categories. The first one is adversarial training \cite{adver_train,huang2015learning}, where AEs are used with normal examples together to train DNN models to recognize and correct malicious samples. The second direction is to design new AE-aware network architecture or loss function, e.g., Deep Contractive Networks~\cite{gu2014towards}, Input Gradient Regularization~\cite{ross2018improving}, Defensive Distillation~\cite{defen_distillation}, Magnet~\cite{Magnet}, Generative Adversarial Trainer~\cite{lee2017generative}. The third direction is to introduce a preprocessing function to transform the input samples and remove the adversarial perturbations by gradient masking \cite{guo2018countering,prakash2018deflecting,xie2018mitigating,buckman2018thermometer,das2018shield}. The last category is to detect adversarial examples \cite{detect_artifacts,detect_LID,activation_spaces,input_transformation,FeatureSqueezing,region_based,model_mutation}. Compared with the first three directions, these methods do not need to train a new model with different structures or datasets, or to alter the inference computing pipeline. So we will focus on the detection-based solutions in this paper.

\bheading{Mitigating backdoor attacks.}
There are also several directions to defeat backdoor attacks. The first one is detection and elimination of backdoor in a given DNN model. To achieve this, past works adopted boundary outlier detection \cite{Neural_cleanse,Deepinspect,defen_Tabor,defen_GDM}, Meta Neural Analysis \cite{defen_MNA}, and artificial brain stimulation \cite{ABS}. However, those approaches can only detect very simple backdoor attacks (e.g., one targeted class, simple triggers), and can be easily bypassed by advanced attacks \cite{tan2019bypassing}. Fine-pruning was used to remove malicious backdoor in the model \cite{Fine_pruning}. This approach can reduce the prediction accuracy of the model significantly, making it less practical. The second direction is to identify poisoned data in the training set \cite{activation_clustering,Spectral_signatures,Differential_Privacy}. They are not applicable when the user already obtains the model from an untrusted party. The third direction is to detect backdoor examples \cite{Strip,Sentinet,Grad_cam}. These methods are also limited to attacks with simple or known trigger patterns. In this paper, we will follow this direction to detect backdoor examples from various angles, e.g., model sensitivity, activation space and feature space. 

\section{Detection Methodologies}
\label{section:insight}

\subsection{Overview}
A good detection method should meet certain criteria, as discussed below.

\noindent{\textbf{Generality.}} This requirement can be reflected in two directions. First, the candidate method should not be attack-specific. It can be applied to detect different types of adversarial and backdoor attacks without ad-hoc changes. Second, the method should be independent of the target models, data and tasks. It is not allowed to modify the models or inference computation. But it can collect the internal information during the inference. 

\noindent{\textbf{Effectiveness}}. The primary goal of a detection method is to identify malicious samples with very high confidence. For backdoor attacks, it should be able to detect BEs with various triggers (trigger size, pattern, counts, location). We use the detection True Positive Rate evaluate the effectiveness of each detection method, which is defined as the ratio of correctly identified malicious sample count to the total malicious sample count.

\noindent{\textbf{Usability}}.
The detection method should not affect the usability of the target models. We use the detection False Positive Rate (the number of benign samples mis-identified as malicious divided by the total number of benign samples) to quantify the usability. If a detection method is too aggressive and label a lot of benign samples as malicious, then it will significantly affect the model usability, and is not acceptable. 

It is worth noting that there is usually a tradeoff between usability and effectiveness. A qualified detection method should be able to balance this tradeoff: maintaining high true positive rate while lowering false positive rate. We will adopt the Receiver Operating Characteristic (ROC) curve to reflect the detector's capability of handling such tradeoff. 


\noindent{\textbf{Performance.}}
A good detection method should have performance efficiency. 
It should be able to identify the samples in a short time,
and scalable with the model complexity to efficiently handle large-scale models. We measure the detection time to quantify the performance of a method. Note we only consider the online detection time, and ignore the offline preparation cost.

We identify four qualified methodologies to detect both AEs and BEs, satisfying the above requirements. Our selection is based on two observations. The first one is the similarity between AEs and BEs. Since both two types of examples are generated by adding small perturbations to enforce the models to make wrong predictions, they exhibit similar features in the interaction with the model, which are distinguishable from benign samples(This is evaluated in Section \ref{sec:behavior} with Remark 2 and Remark 3). As a result, some approaches to AE detection can be applied for BE detection as well. The second observation is the difference between AEs and BEs: BEs are generally farther away from the decision boundary than AEs, and show more robustness (This is evaluated in Section \ref{sec:behavior} with Remark 1). So some approaches for detecting AEs may not work for BEs. Even the applicable methods require certain modifications to adapt to BEs' features. Below, we describe the details of four methodologies.

\begin{figure}[t]
	\centering
	\begin{minipage}[t]{1\linewidth}
		\label{mutation_insight}
		\centering
		\includegraphics[width=1.0\textwidth]{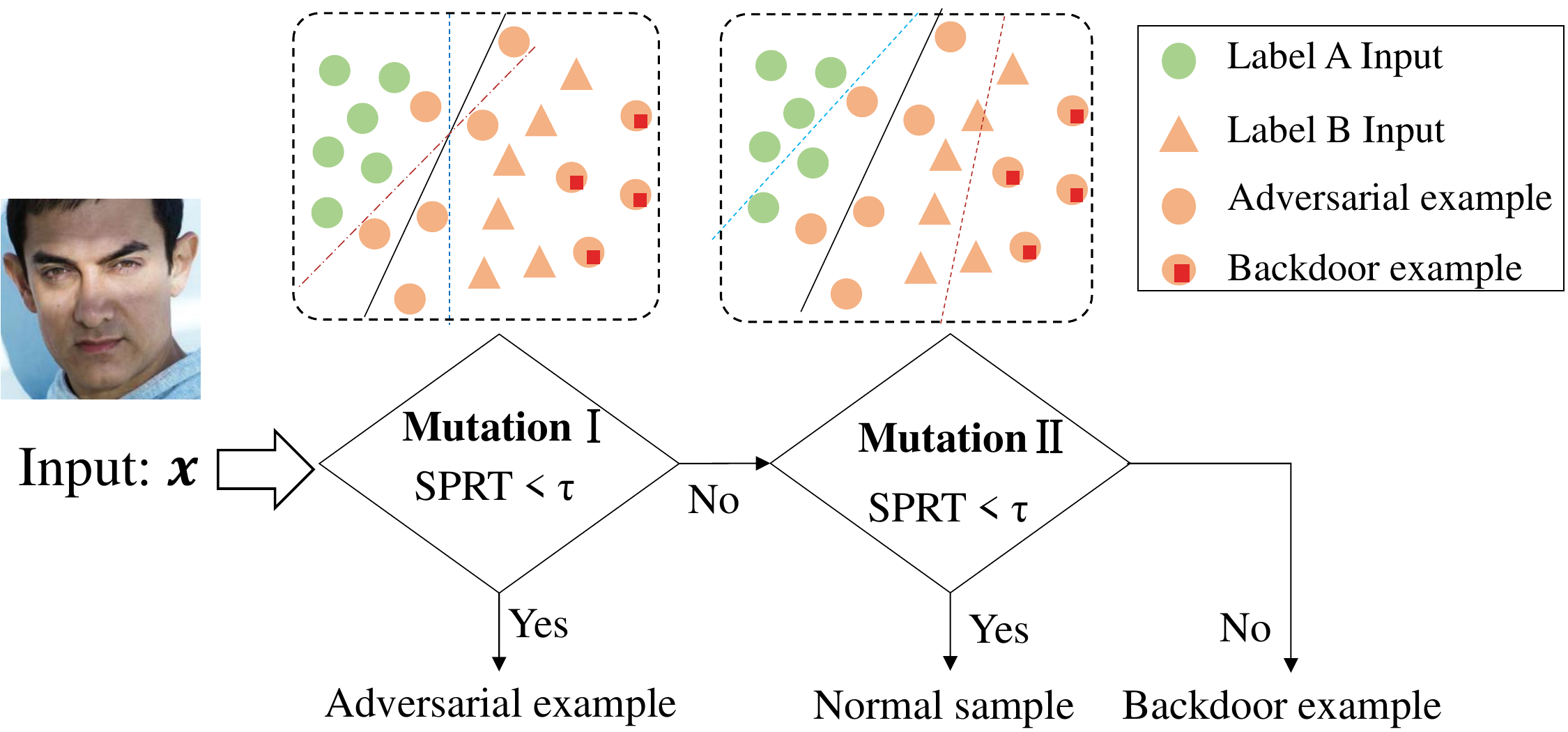}            
	\end{minipage}
	\caption{Workflow of Model Mutation.}
	\label{fig:Muation_insight_workflow}    
	\vspace{-10pt}
\end{figure}

\subsection{Model Mutation}

\noindent{\textbf{Detecting AEs.}}
The first approach we consider is model mutation \cite{model_mutation}. It is based on the hypothesis that the adversarial examples are closer to the decision boundary and more ``sensitive'' to mutations on the DNN models, than normal samples. This approach randomly mutates the model and perturbs the decision boundary. Then the predication of AEs has a higher chance to be altered from their original labels (Mutation \uppercase\expandafter{\romannumeral1} in Figure \ref{fig:Muation_insight_workflow}).

Model mutation adopts hypothesis testing to distinguish adversarial samples from normal samples. Specifically given a DNN model $f_\theta$ and a sample $x$, we can establish two exclusive hypothesises: $H_0$ ($x$ is an adversarial example): $\varsigma(x) > \varsigma_{h}$  and $H_1$ ($x$ is a benign example): $\varsigma(x) \le \varsigma_{h}$, where  $\varsigma(x)$ is the label change rate of sample $x$ and $\varsigma_{h}$ is a threshold to determine the sample attributes. The intuition is that $\varsigma(x)$ is statistically much larger when $x$ is an adversarial example than normal ones, which can be distinguished by the threshold $\varsigma_{h}$.

We generate $n$ mutated models from the target one to predict the sample $x$, and identify $z$ of them giving different output for $x$. Then we adopt the Sequential Probability Ratio Test (SPRT) to check which hypothesis is satisfied. Three parameters, $\alpha$, $\beta$, $\delta$ are used to control the probability of error tolerance.
Then SPRT is calculated in Equation \ref{eq:sprt}, where $p_{1} = \varsigma_{h} - \delta$ and $p_{0} = \varsigma_{h} + \delta$. The hypothesis $H_0$ is accepted if $SPRT \le \frac{\beta}{1-\alpha}$, indicating that $x$ is an adversarial example. Otherwise, $H_1$ is accepted and $x$ is normal. 

\begin{equation}
\label{eq:sprt}
	\begin{aligned}
	    SPRT = \frac {p_{1}^{z} (1-p_{1})^{n-z}}    {p_{0}^{z} (1-p_{0})^{n-z}}
	\end{aligned}
\end{equation}

\noindent{\textbf{Detecting BEs.}}
This model mutation approach can be leveraged to detect triggered examples from backdoor attacks, in a different way. As we discussed previously, backdoor examples enjoy higher robustness against decision boundary changes, than adversarial examples and benign samples (Mutation \uppercase\expandafter{\romannumeral2} in Figure \ref{fig:Muation_insight_workflow}). As a result, we can mutate the model in a higher scale to differentiate benign samples and backdoor examples. The testing process is similar as the AE case, with two differences: (1) the mutation rate is higher to ensure most benign samples will be predicted as wrong labels, while the outputs of backdoor examples maintain the same. (2) The hypothesises now is reversed: $H_0$ ($x$ is a benign sample): $\varsigma(x) > \varsigma_{h}$  and $H_1$ ($x$ is a triggered example): $\varsigma(x) \le \varsigma_{h}$. 

We can put these two stages together to form our unified approach to detection of malicious examples, as illustrated in Figure \ref{fig:Muation_insight_workflow}. First, we set a small mutation rate to check if the sample is an AE. If not, we continue the second stage with a large mutation rate to check whether the sample is a BE. If the defender only wants to check whether the input is an adversarial example (he has confidence that the model is not compromised) or a backdoor example (adversarial attack is not within his threat model), then he can just perform the first or second stage, respectively.  


\subsection{Activation Space}
\noindent{\textbf{Detecting AEs.}}
This methodology \cite{activation_spaces} explores the sample behaviors in the activation space of different network layers. The hypothesis is that the behaviors of normal samples are different from that of adversarial examples. Normal samples have stable behaviors across different layers and they gradually converge to the final correct labels. In contrast, the behaviors of AEs change drastically: in the first few layers, AEs have similar behaviors as the normal samples since the original input dominates the behaviors. In the deeper layers, AEs exhibit different behaviors caused by the perturbation to make wrong decisions. Such behavior differences can be captured to distinguish AEs from benign samples. 
	
The detection consists of two stages. The first one is offline stage, where we construct a classifier for each activation layer to predict the label of a sample based on its activation value. For the activation layer $i$, the goal is to train a classifier $c^i: f_\theta^{1...i}(x) \mapsto y$ for a sample $(x, y)$ where $f_\theta^{1...i}(x)$ is the activation value of sample $x$ at layer $i$. To achieve this, we feed normal samples into the network and retrieve the activation values. Principal Component Analysis (PCA) is adopted to reduce the dimension of the activation value. A KNN classifier $c^i$ is trained over this set $(f_\theta^{1...i}(x), y)$ for layer $i$. With the classifiers, we calculate the priori switching probability of predicted labels between consecutive activation layers (Equation \ref{eq:as}).

\begin{equation}
\label{eq:as}
  \begin{aligned}
 p_s^{i} = P(c^i(f_\theta^{1...i}(x)) \neq c^{i-1}(f_\theta^{1...i-1}(x))), \forall i \in[1, l] \\
  \end{aligned}
\end{equation}

\begin{figure}[t]
	\centering
	\begin{minipage}[t]{1\linewidth}
		\centering
		\includegraphics[width=1.0\textwidth]{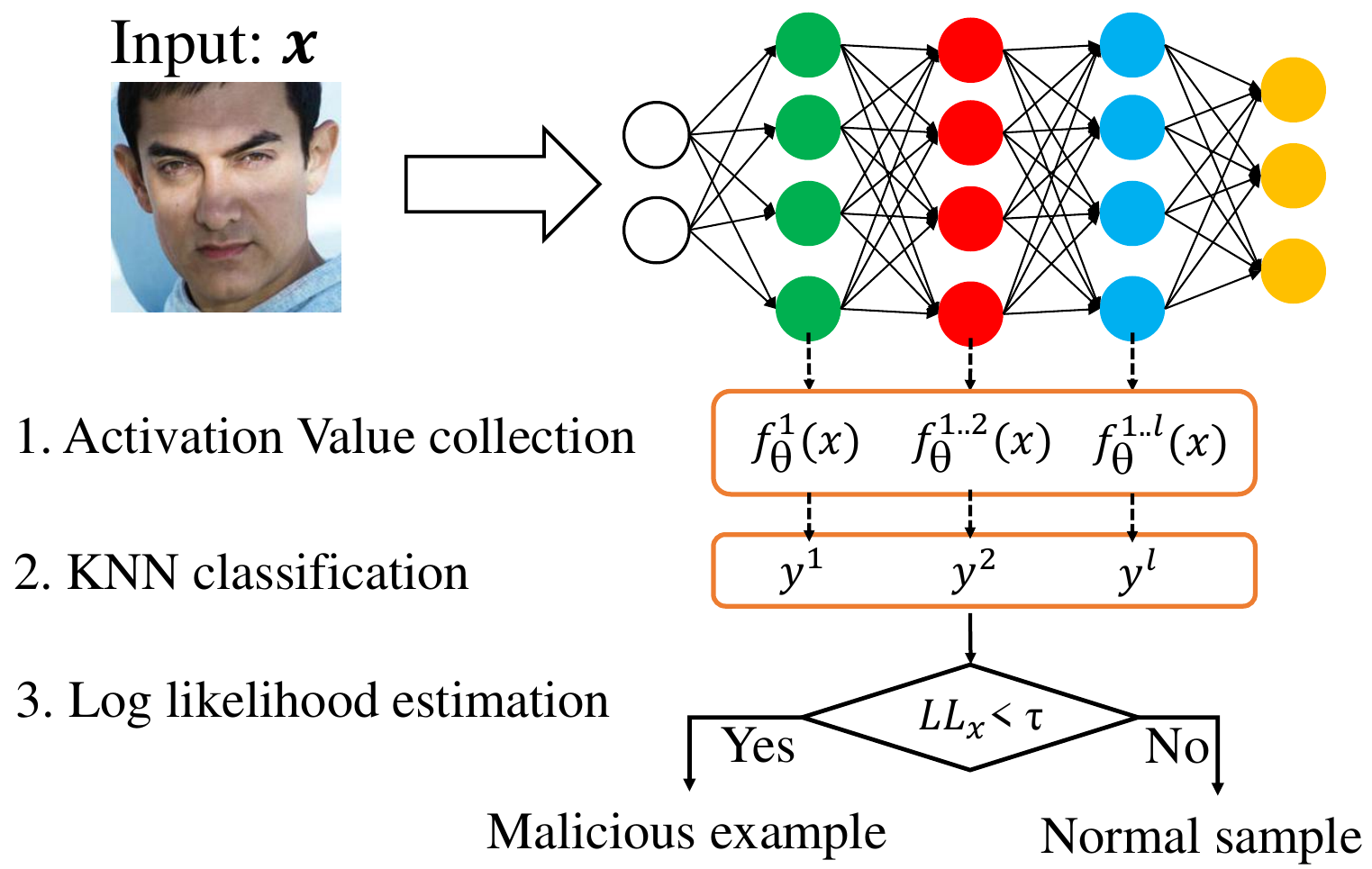} 
	\end{minipage}

	\caption{Workflow of Activation Space.}
	\label{fig:AS_workflow}    
	\vspace{-10pt}
	\end{figure}

The second one is online stage, which is shown in Figure \ref{fig:AS_workflow}. For the target sample $x$, we feed it into the network, collect the activation values, and use the corresponding classifier to predict its label $y^i=c^i(f_\theta^{1...i}(x))$. A normal sample always has low switching probability through all layers, while AEs can have abrupt increase in the probability due to the behavior changes. To quantify this effect, we estimate the log likelihood of the target example $x$ by Equation \ref{eq:as1}, and compare it with a threshold $\tau$. The sample $x$ is flagged as an AE when $LL_x < \tau$.

\begin{equation}
\label{eq:as1}
  \begin{aligned}
LL_{x} = \sum_{i=1}^l \log [\frac{1}{2}+(-1)^{(y^i \neq y^{i-1})}(\frac{1}{2}-p_s^i)] 
  \end{aligned}
\end{equation}

\noindent{\textbf{Detecting BEs.}}
Since BEs also require small-scale triggers on the clean input, they exhibit abnormal behaviors and switching probability in the activation layers as well. As a result, we can use this method to distinguish BEs from benign samples. Figure \ref{fig:AS_workflow} shows the workflow of this method.

\subsection{Kernel Density Estimation}
\noindent{\textbf{Detecting AEs.}}
This approach \cite{detect_artifacts} focuses on the anomaly detection in the feature space. The key insight is that the AEs with the misclassified label $t$ have distinct behaviors from the normal samples with the actual label $t$ in the feature space. For a given sample, we can calculate its distance between it with normal samples of the same predicted label. A larger distance indicates the sample is potentially malicious. 

\begin{figure}[t]
	\centering
		\begin{minipage}[t]{1\linewidth}
			\centering
			\includegraphics[width=1.0\textwidth]{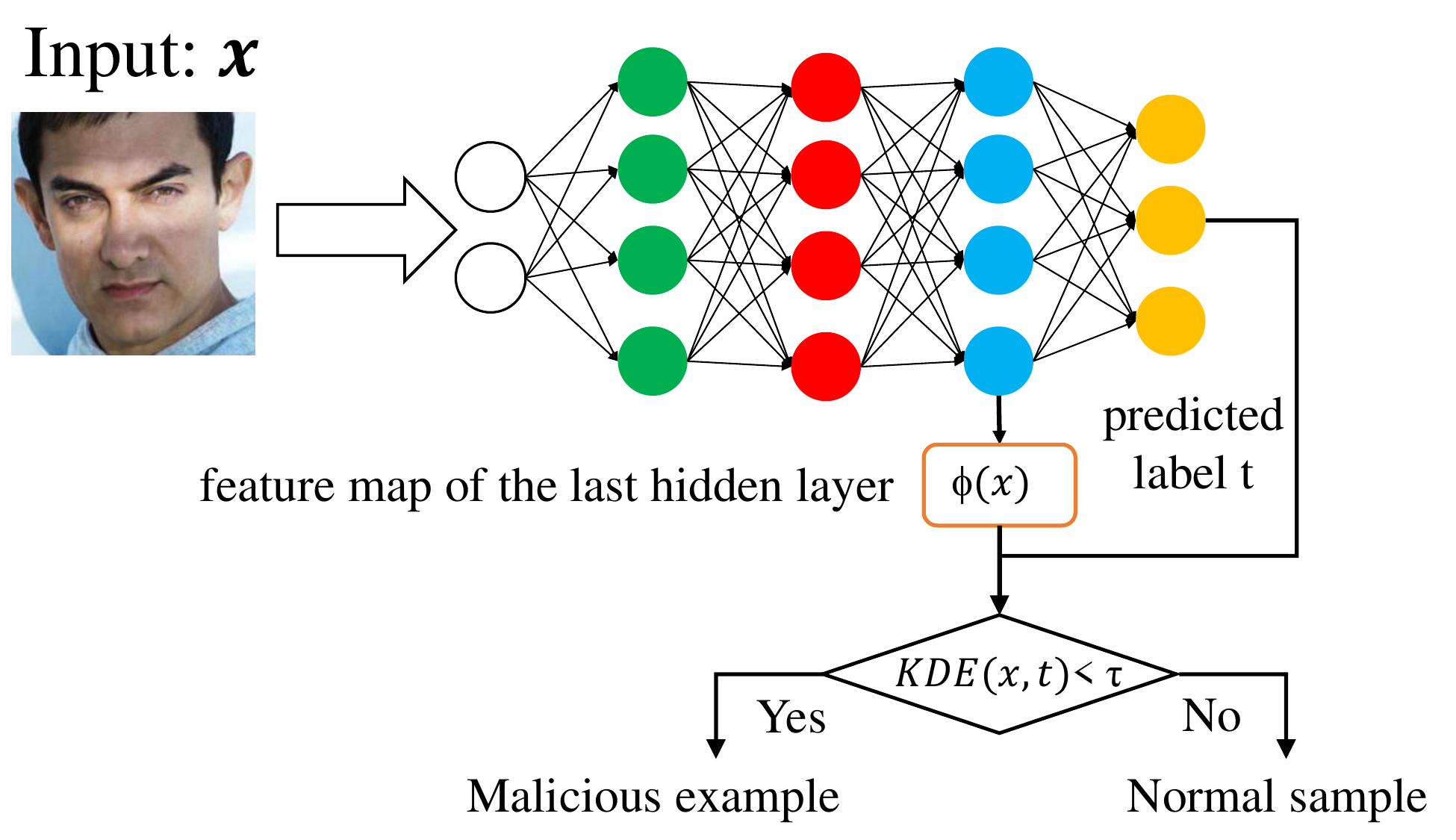} 
		\end{minipage}	
	\caption{Workflow of Kernel Density Estimation.}
	\label{fig:KD_workflow}    
	\vspace{-10pt}
	\end{figure}

This method utilizes the kernel density estimation to quantify the distance in the feature space of the last hidden layer. As illustrated in Figure \ref{fig:KD_workflow}, for the target sample $x$, its predicted label is denoted as $t$. Then we obtain a set $X_t$ of training samples with the same label $t$. Equation \ref{eq:kde} gives the density estimation ($KDE$) to measure the distance, where $\phi(x)$ is the last hidden layer activation vector for point $x$. If $KDE(x,t)<\tau$, $x$ is reported as a malicious sample, where $\tau$ is a predefined threshold.

\begin{equation}
\label{eq:kde}
	\begin{aligned}
	KDE(x,t)=\frac{1}{\left|X_{t}\right|} \sum_{x_{i} \in X_{t}} \exp \left(-\|\phi(x_i)-\phi(x)\|^{2} / \sigma^{2}\right)
	\end{aligned}
\end{equation}

\noindent{\textbf{Detecting BEs.}}
Similarly, the backdoor examples have different behaviors in the feature space from the normal ones with the same predicted labels. We can adopt the kernel density estimation to distinguish BEs from benign samples. It is hard to identify AEs and BEs as they have similar features. So we use the same threshold to detect both of them.

\subsection{Local Intrinsic Dimensionality}

\noindent{\textbf{Detecting AEs.}}
This approach \cite{detect_LID} follows the similar idea as Kernel Density Estimation. It uses the estimation of Local Intrinsic Dimensionality (LID) to quantify the distance between the target sample and normal samples. Given a sample $x$ and the set $X_t$ of normal samples with the same predicted label, the Maximum likelihood Estimator (MLE) of LID at $x$ is calculated in Equation \ref{eq:lid}, where $r_{i}(x)$ represents the Euclidean distance of feature maps between $x$ and its $i$-th nearest neighbor within $X_t$, and $r_{k}(x)$ is the maximum of the neighbor distances. The LID value of an AE is significantly higher than normal data. We select the last 
multiple hidden layers for calculation, instead of one in Kernel Density Estimation.

\begin{equation}
\label{eq:lid}
	\begin{aligned}
	LID(x,t)=-\left(\frac{1}{k} \sum_{i=1}^{k} \log \frac{r_{i}(x, X_t)}{r_{k}(x, X_t)}\right)^{-1}
	\end{aligned}
\end{equation}

\noindent{\textbf{Detecting BEs.}}
Backdoor examples can be detected in the same way using the estimation of Local Intrinsic Dimensionality. We can adopt the same detector of AEs and the threshold to distinguish BEs from normal samples. 

	\begin{figure}[ht]
	\centering
	\begin{minipage}[ht]{1\linewidth}
		\centering
		\includegraphics[width=1.0\textwidth]{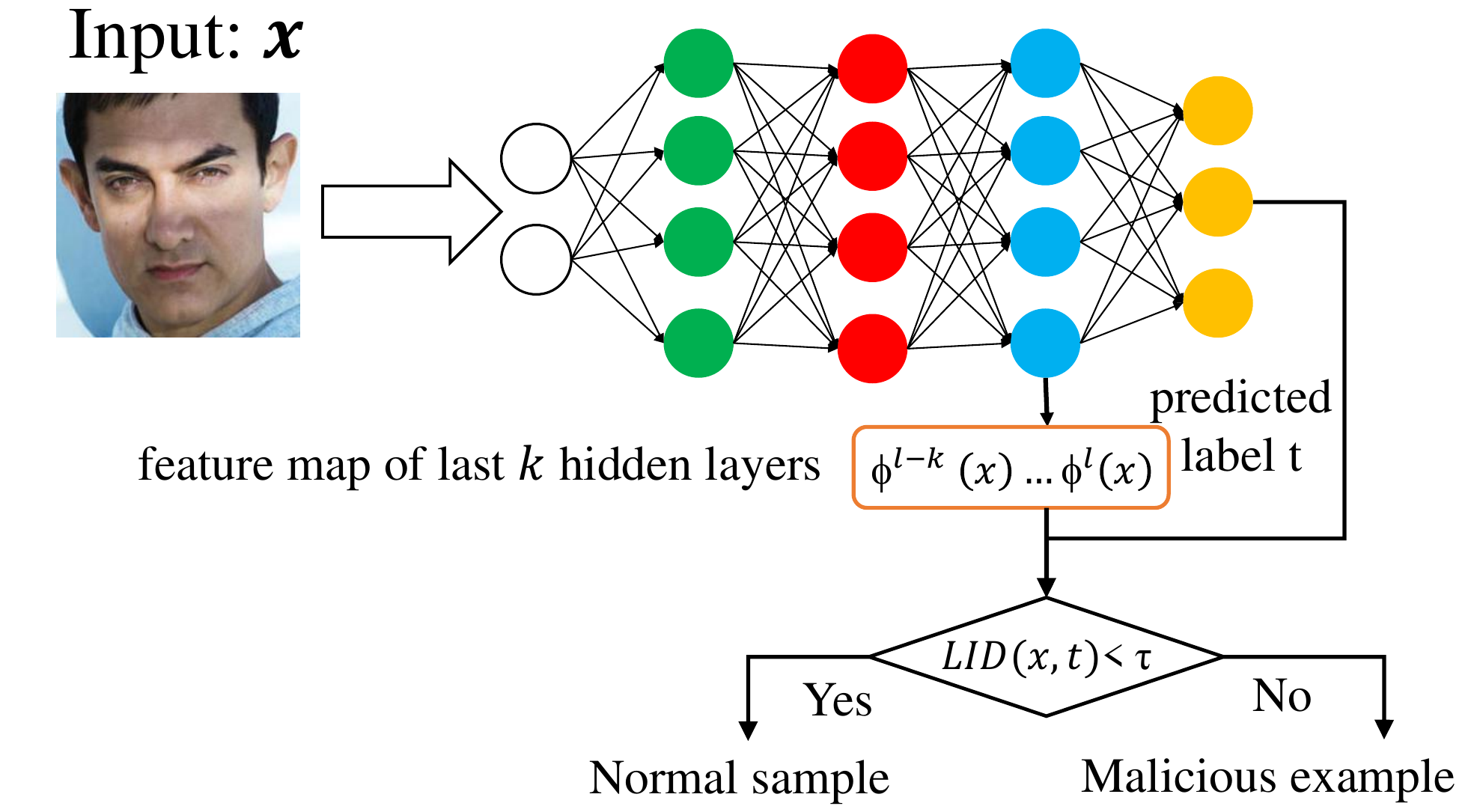} 
	\end{minipage}	
	\caption{Workflow of Local Intrinsic Dimensionality.}
	\label{fig:LID_workflow}    
	\vspace{-10pt}
	\end{figure}

	\begin{table*}[ht]
		\caption{Details of the attacks and the target models. }
		\begin{center}
			\renewcommand{\multirowsetup}{\centering}
			\resizebox{\linewidth}{!}{
			\begin{tabular}{c|c|c|c|c|c|c|c|c|c|c}
				\hline
				\multirow{3}{*}{Task} & \multicolumn{4}{c|}{Dataset} & \multicolumn{3}{c|}{DNN Model} & \multicolumn{3}{c}{Attacks} \\\cline{2-11}
				& Name &\makecell{\# of \\classes} & Images size & \makecell{\# of training \\samples} &  Architecture & \makecell{\# of trainable \\parameters} & \makecell{Classification \\accuracy} & \makecell{Trigger \\type} & \makecell{Success\\rate} & \makecell{Accuracy of\\clean samples}\\
				\hline\hline
				\makecell{Hand-writing \\ Digits Recognition} & MNIST & 10 & 28$\times$28$\times$1 & 60,000 & 2Conv+2FC & 413,882 & 98.98\% & White square & 100\% & 99.11\%\\
				\hline
				\makecell{Traffic Sign \\ Reconfignition} & GTSRB & 43 & 32$\times$32$\times$3 & 35,288 & 6Conv+2FC & 571,723 & 97.79\% & White square & 97.44\% & 96.51\%\\
				\hline
				\multirow{2}{*}{Face Recognition} & \multirow{2}{*}{PubFig} & \multirow{2}{*}{83} & \multirow{2}{*}{224$\times$224$\times$3} & \multirow{2}{*}{11,070} & \multirow{2}{*}{13Conv+3FC} & \multirow{2}{*}{122,245,715} & \multirow{2}{*}{95.56\%} & Colored square & 100\% & 95.27\%\\
				\cline{9-11}
				& & & & & & & & Watermark (WM) & 99.89\% & 94.76\%\\
				\hline
			\end{tabular}}
			\label{experiment_model}
		\end{center}
	\end{table*}

	\begin{figure*}[t]
		
		\subfloat[Original~MNIST]{
			\begin{minipage}[t]{0.13\linewidth}
				\label{Fig-raw_mnist}
				\centering
				\includegraphics[width=1\textwidth]{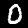}
			\end{minipage}
		}
		\hfill
		\subfloat[BE MNIST ]{
			\begin{minipage}[t]{0.13\linewidth}
				\label{Fig-adv_mnist}
				\centering
				\includegraphics[width=1\textwidth]{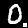}
			\end{minipage}
		}
		\hfill
		\subfloat[Original GTSRB]{
			\begin{minipage}[t]{0.13\linewidth}
				\label{Fig-clean_gtsrb}
				\centering
				\includegraphics[width=1\textwidth]{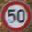}
			\end{minipage}
		}
		\hfill
		\subfloat[BE GTSRB]{
			\begin{minipage}[t]{0.13\linewidth}
				\label{Fig-adv_gtsrb}
				\centering
				\includegraphics[width=1\textwidth]{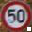}
			\end{minipage}
		}
		\hfill
		\subfloat[Original PubFig]{
			\begin{minipage}[t]{0.13\linewidth}
				\label{Fig-clean_face}
				\centering
				\includegraphics[width=1\textwidth]{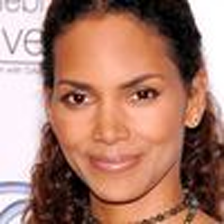}
			\end{minipage}
		}
		\hfill
		\subfloat[Square PubFig]{
			\begin{minipage}[t]{0.13\linewidth}
				\label{Fig-square_face}
				\centering
				\includegraphics[width=1\textwidth]{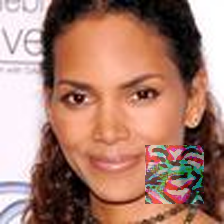}
			\end{minipage}
		}
		\hfill
		\subfloat[WM PubFig]{
			\begin{minipage}[t]{0.13\linewidth}
				\label{Fig-wm_face}
				\centering
				\includegraphics[width=1\textwidth]{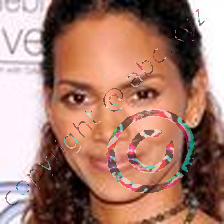}
			\end{minipage}	
		}
		\caption{Backdoor Examples.}
		\label{fig:attack_instances}    
	\end{figure*}

Our experiments consists of both state-of-the-art attacks and effective detection solutions introduced in Section \ref{section:insight}. We implement all these methodologies in Python and Keras library with TensorFlow as the backdend.

\subsection {Attacks}
Since there are already some well-developed toolkits for adversarial attacks \cite{ling2019deepsec,ibm-adv}, we mainly collect backdoor attacks in our experiments. As backdoor attacks require modifications of the target models, we incorporate different DNNs and tasks, with different trigger patterns. We adopt the attack technique in BadNet \cite{badnets} to inject DNN backdoor. Table \ref{experiment_model} summarizes the attack information, and Figure \ref{fig:attack_instances} visualizes the generated backdoor examples.

\noindent{\textbf{Handwritten digits recognition.}}
We select the MNIST dataset \cite{MNIST_DATA}, which contains 60K training images and 10K testing images. Each data sample is a 28$\times$28$\times$1 greyscale image. We set a white square with the size of 4$\times$4 pixels on the bottom right and 1-pixel margin from the border as the trigger (Figure \ref{fig:attack_instances}b). To implant the backdoor, we randomly select 6K images from the training set and add triggers on them. We choose digit ``1'' as the backdoor target label. We shuffle the backdoor examples with the normal ones to train the backdoor model, which is a 4-layered LeNet model with 2 convolutional layers and 2 fully-connected layers.  

\noindent{\textbf{Traffic sign recognition.}}
We adopt the infected model from \cite{Neural_cleanse}. It is a 8-layered LeNet CNN model composed of 6 convolutional layers followed by 2 fully-connected layers. This model is trained from the GTSRB dataset \cite{GTSRB_DATA}, which consists of 35,288 training images and 12,630 testing images in 43 classes. Its input space is 32$\times$32$\times$3 pixels. The trigger size is also a white square with the size of 5$\times$5 pixels (Figure~\ref{fig:attack_instances}d). 

\noindent{\textbf{Face recognition.}}
We select the PubFig dataset \cite{PubFig_data}, which consists of 11,070 training images and 2,768 test images of 83 celebrities. The input space of each image is 224$\times$224$\times$3. We choose two triggers with more complex patterns, as shown in Figures \ref{fig:attack_instances}f and \ref{fig:attack_instances}g. The backdoor target label is set as ``0''. We use the state-of-the-art VGG-16 model for face recognition. Following the strategy in \cite{Neural_cleanse}, we fine-tune the model from a benign one by only training the parameters of the last four layers while freezing the other layers. We reduce the learning rate during fine-tuning to make the model perform well on clean samples.

Table \ref{experiment_model} also reports the backdoor attack results and the prediction accuracy on clean samples. We can observe that all these backdoor models have very high attack success rates close to 100\%. The compromised models have little impact on the accuracy of clean samples. This verifies the effectiveness of backdoor attacks.

\subsection {Detection Methods}
We implement the four AE defense approaches with modifications for BE detection. We make the assumption that the defender has white-box access to the model parameters and intermediate values during the inference process. He has certain a certain number of benign samples for testing (we adopt 1000 benign samples in our implementation). We identify the parameters of those defenses for different target models (Table \ref{MM_parameters}). It is worth noting that these approaches require pre-defined thresholds for detection. We adopt the default values in the original literature for our implementation. The threshold is attack-independent but relies on the datasets. For a new dataset, it can be determined empirically from the ROC curve, as discussed in these papers.

\noindent{\textbf{Model Mutation}}. This methodology requires a quantity of mutated models. Four mutation operators were used in \cite{model_mutation}. We select Gaussian Fuzzing (GF) which can give the best results. Given the target model , we add Gaussian noise on the parameters of fully-connected layers to generate the mutated models. The amount of Gaussian noise is determined by two parameters: variance ($\delta$) and mean ($\mu$). We set two mutation factors: $r^{\delta}$ and $r^{\mu}$. The mean value of noise distribution is calculated as the mean value of the FC layer weights multiplied by $r^{\mu}$. The variance value of noise distribution is the maximal value of the FC layer weights multiplied by $r^{\delta}$. 

The values of mutation factors need to be carefully selected. For Mutation \uppercase\expandafter{\romannumeral1} of detecting AEs, if the mutation factors are too large, normal samples will change the labels as well, increasing the false positive rate. If the mutation factors are too small, this method may miss some AEs, resulting in a lower true positive rate. For Mutation \uppercase\expandafter{\romannumeral2} of detecting BEs, larger mutation factors can decrease the true positive rate while smaller mutation factors lead to a higher false positive rate. Through empirical exploration, we identify the optimal parameters for the two sets of model mutations, as shown in Table \ref{MM_parameters}. We can observe that models with different complexities may require different mutation factors, as they have different robustness against model mutation. The numbers of mutated models in both two sets are 100. 



	\begin{table}[ht]
		\small
		\vspace{-5pt}
		\caption{Parameter selection of different approaches.}
		\vspace{-8pt}
		\begin{center}
			\renewcommand{\multirowsetup}{\centering}
			\begin{tabular}{c|c|c|c|c|c|c}
				\hline
				\multirow{3}{*}{Dataset} & \multicolumn{4}{c|}{Model Mutation} & KD & LID\\
				\cline{2-7}
				& \multicolumn{2}{c|}{Mutation \uppercase\expandafter{\romannumeral1}} & \multicolumn{2}{c|}{Mutation \uppercase\expandafter{\romannumeral2}} & \multirow{2}{*}{$\sigma$} & \multirow{2}{*}{$k$}\\
				\cline{2-5}
				&\makecell{$r^\mu$} & \makecell{$r^\delta$}&\makecell{$r^\mu$} & \makecell{$r^\delta$} & & \\
				\hline
				\makecell{MNIST} & 1.0 & 0.3 & 1.0 & 0.65 & 1.2 & 20\\
				\hline
				\makecell{GTSRB} & 1.0 & 0.35 & 1.0 & 0.65 & 0.1 & 30 \\
				\hline
				\makecell{PubFig} & 0.2 &0.2 & 1.0 & 0.65 & 0.5 & 10\\
				\hline
			\end{tabular}
			\label{MM_parameters}
		\end{center}
		\vspace{-10pt}
	\end{table}

\noindent{\textbf{Activation Space}}.
We set PCA components as 100 when constructing the activation space. The number of neighbors in KNN classifier is 5. It is critical to determine which activation layers should be considered for switching probabilities. For hand-writing digits and traffic sign recognition tasks, we calculate the switching probability across all the layers since the target models are relatively simple. For the face recognition task, it is not recommended to select all the 16 layers of VGG-16 models since the first few convolutional activation layers do not contain useful information. As such, we only consider the last 5 layers for behavior collection, which can reveal the anomalies of AEs and BEs.  


\noindent{\textbf{Kernel Density Estimation.}}
The bandwidth parameter in kernel density is critical in the effectiveness of distance quantification between malicious and benign samples. Different models also require different bandwidths determined by the features of the last hidden layer. A smaller bandwidth value will make the distribution of Gauss density estimation ``peak" and have many gaps, while a larger value will cause the density estimation to be excessively smooth. We identify the optimal bandwidth values for different models through evaluations, as described in Table \ref{MM_parameters}.


\noindent{\textbf{Local Intrinsic Dimensionality.}}
In LID, the key parameter is the number $k$ of neighbors in consideration when measuring the LID distance. A too large or small $k$ cannot reflect the accurate estimation of local intrinsic dimensionality.  Through empirical evaluations, we discover the appropriate parameter values, as reported in Table \ref{MM_parameters}. 
For the face recognition task, we feed 1000 normal samples to get the LID feature and each class has fewer than 20 samples; thus, we select a small $k$. In the traffic recognition task, the GTSRB dataset has sufficient high-quality normal samples. So we use a large $k$ value.

\section {Evaluations}
In this seciton, we measure and compare the methodologies of detecting AEs and BEs from different perspectives. For adversarial attacks, we choose the state-of-the-art method C\& W technique \cite{papernot2018cleverhans}. For backdoor attacks, we consider the four backdoor models listed in Table \ref{experiment_model}.

\subsection{Behavior Analysis}
\label{sec:behavior}

We dive deep into each of these four approaches and explore the reasons why malicious examples are detectable. 

\begin{figure*}[!t]
		\centering
		\subfloat
		{
			\begin{minipage}[t]{0.23\linewidth}
				\label{mnist_mutation_1}
				\centering
				\includegraphics[width=1\textwidth]{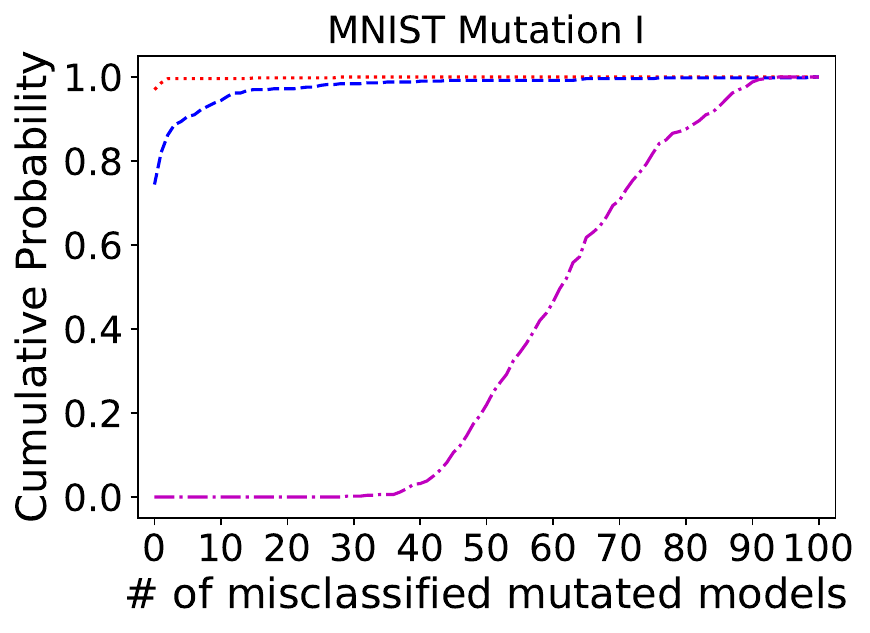}            
			\end{minipage}
		}
		\subfloat
		{
			\begin{minipage}[t]{0.23\linewidth}
				\label{gtsrb_mutation_1}
				\centering
				\includegraphics[width=1\textwidth]{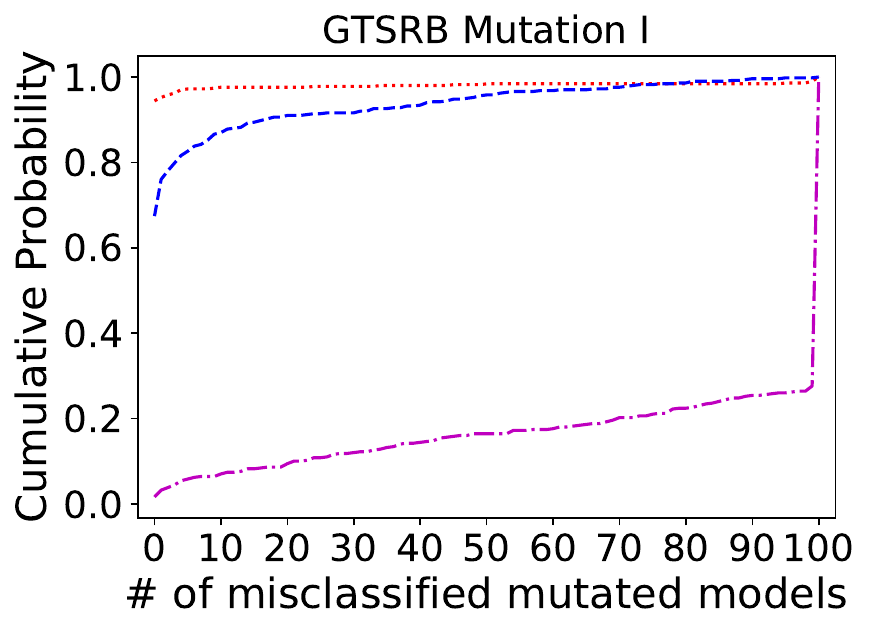}            
			\end{minipage}
		}
		\subfloat
		{
			\begin{minipage}[t]{0.23\linewidth}
				\label{face_wm_mutation_1}
				\centering
				\includegraphics[width=1\textwidth]{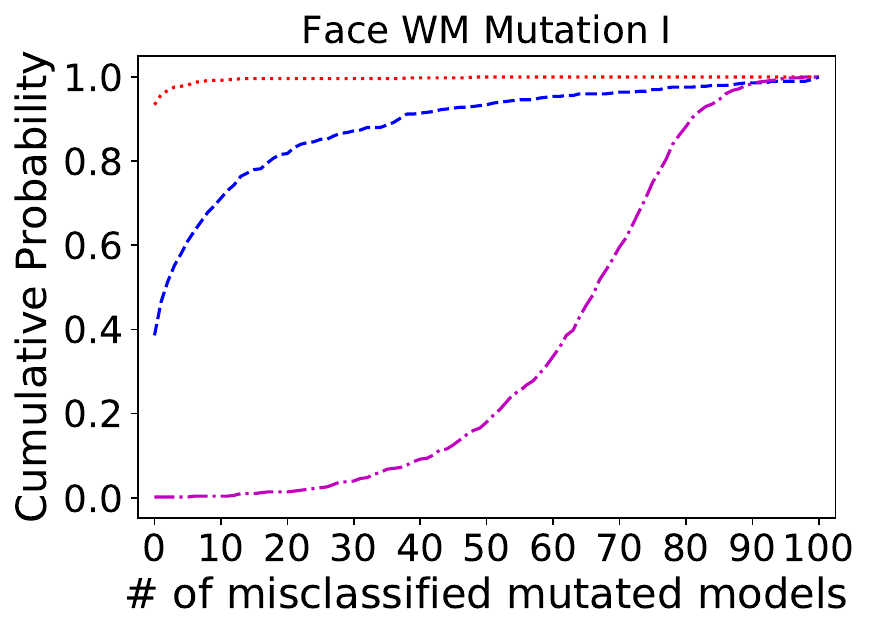}            
			\end{minipage}
		}
		\subfloat
		{
			\begin{minipage}[t]{0.23\linewidth}
				\label{face_square_mutation_1}
				\centering
				\includegraphics[width=1\textwidth]{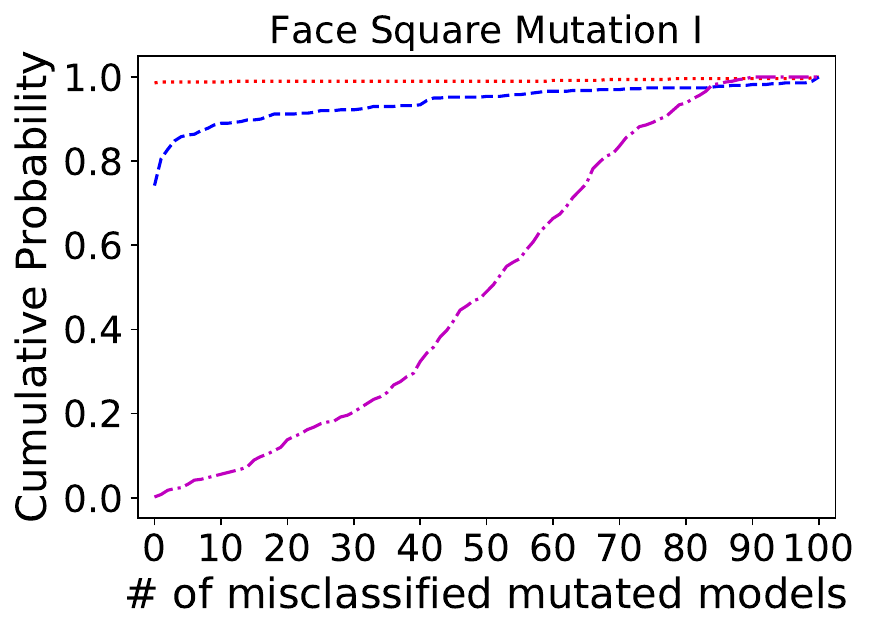}            
			\end{minipage}
		}
		\hfill
		\subfloat{
			\begin{minipage}[t]{0.23\linewidth}
				\label{mnist_muatation_2}
				\centering
				\includegraphics[width=1\textwidth]{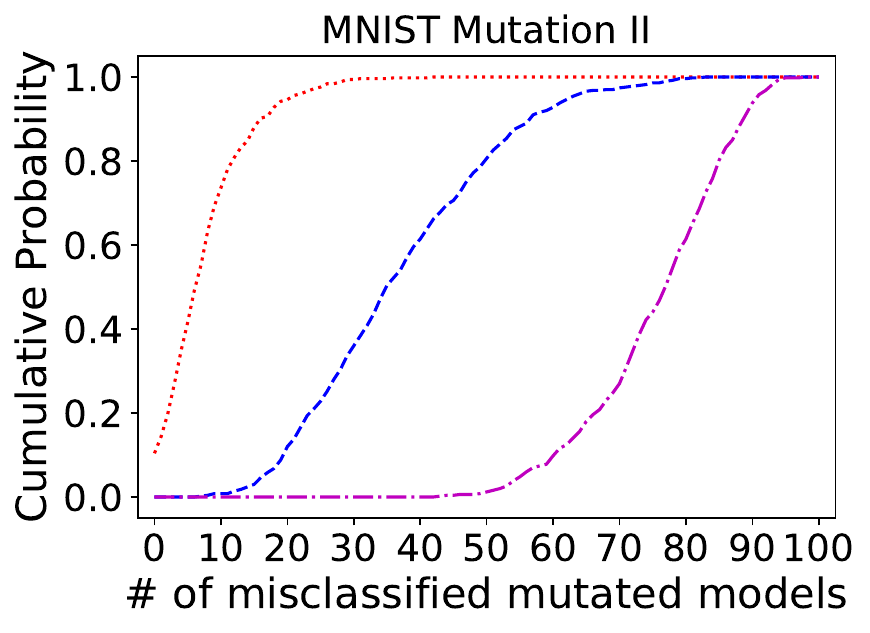}
			\end{minipage}
			
		}
		\subfloat{
			\begin{minipage}[t]{0.23\linewidth}
				\label{gtsrb_muatation_2}
				\centering
				\includegraphics[width=1\textwidth]{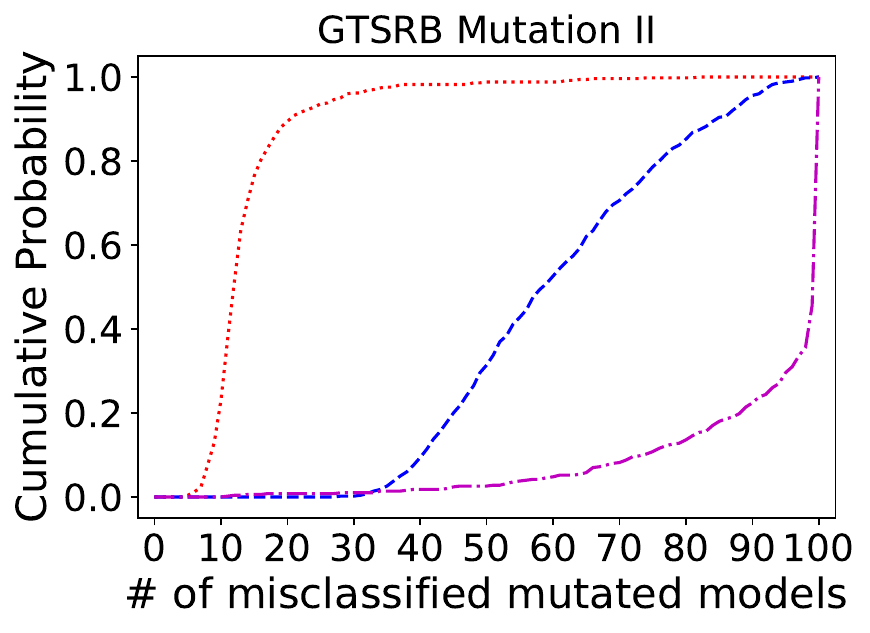}
			\end{minipage}
		}
		\subfloat{
			\begin{minipage}[t]{0.23\linewidth}
				\label{face_wm_muatation_2}
				\centering
				\includegraphics[width=1\textwidth]{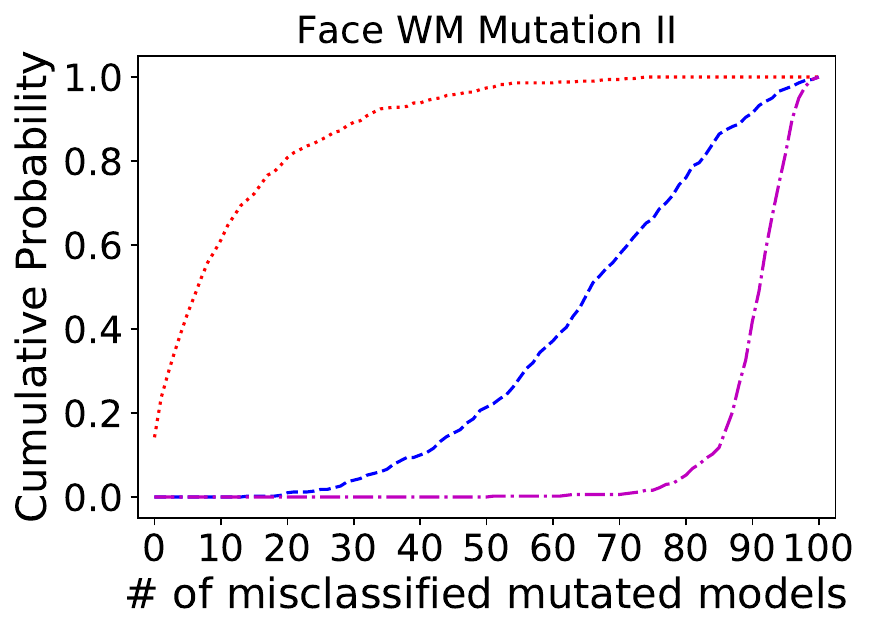}
			\end{minipage}
			
		}
		\subfloat{
			\begin{minipage}[t]{0.23\linewidth}
				\label{face_square_muatation_2}
				\centering
				\includegraphics[width=1\textwidth]{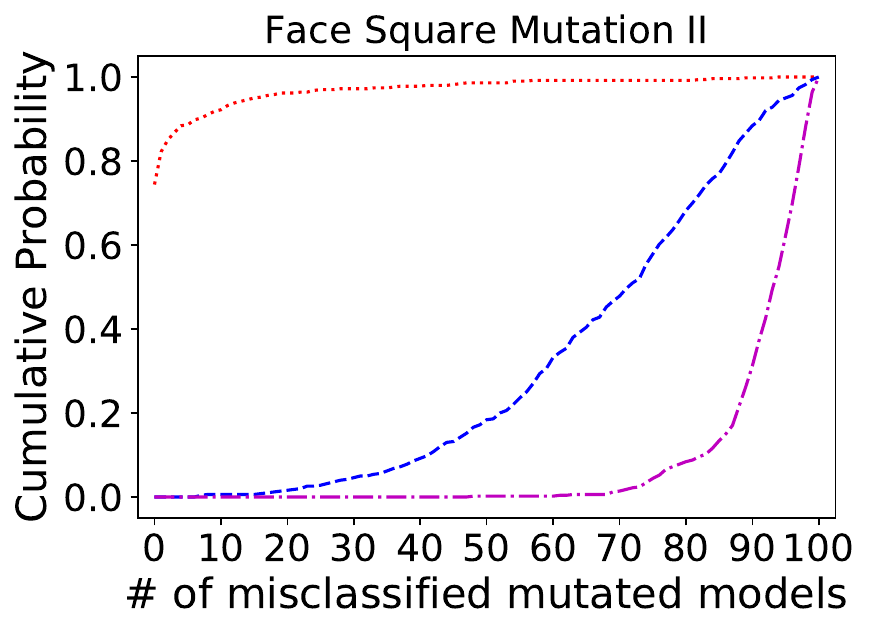}
			\end{minipage}
			
		}
		\hfill
		\subfloat{
			\begin{minipage}[t]{0.6\linewidth}
				\label{legend}
				\centering
				\includegraphics[width=0.6\textwidth]{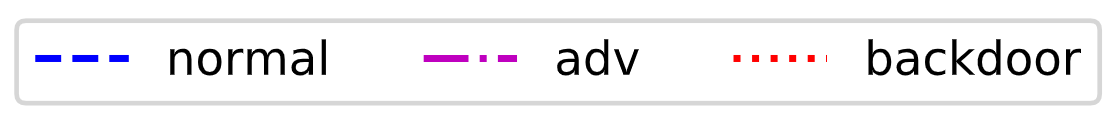}
			\end{minipage}
		}
		\vspace{-5pt}
		\caption{Cumulative probability distribution of label change times under Mutation \uppercase\expandafter{\romannumeral1} (first row) and Mutation \uppercase\expandafter{\romannumeral2} (second row).}
		\label{fig:MM_LCR}    
	\end{figure*}

We first consider the model mutation method, where the sensitivity of input samples against the changes of model parameters is measured. We consider two mutation rates (\uppercase\expandafter{\romannumeral1} and \uppercase\expandafter{\romannumeral2}). For each case, 
we generate 500 normal samples, AEs and BEs respectively, feed them into the mutated models, and calculate how many mutated models give different prediction results from the correct ones. Figure \ref{fig:MM_LCR} shows the cumulative probability distribution of label change counts for each type of samples in different datasets. The first row is the result for Mutation \uppercase\expandafter{\romannumeral1}. We observe that a lot of mutated models give different results from the original model when classifying an AE, and their cumulative probability distributions are different from BEs and normal samples, which are robust against the mutation. The second row reports the case of Mutation \uppercase\expandafter{\romannumeral2}. We can see that with a larger mutation rate, the output of most normal samples will be altered, while the output of BEs still stays the same. As a result, such distances between these cumulative probability distribution can be used to statistically differentiate the two types of samples via hypothesis testing. 

\noindent 
\begin{center}
\fbox{\parbox{85mm}
	{\textbf{Remark 1}: \emph{AEs, BEs and normal samples exhibit different sensitivities to model mutation. AEs are the most sensitive, while BEs are the most robust.}}
}
\end{center}

\begin{figure}[h]
	
	\subfloat{
		\begin{minipage}[t]{0.5\linewidth}
			\centering
			\includegraphics[width=1.0\textwidth]{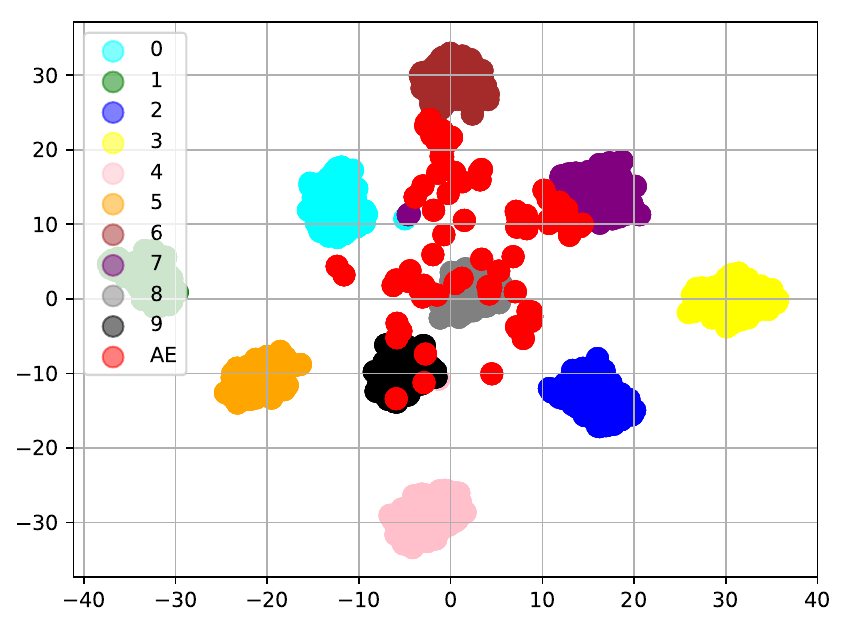}
		\end{minipage}
	}
	\subfloat{
		\begin{minipage}[t]{0.5\linewidth}
			\centering
			\includegraphics[width=1.0\textwidth]{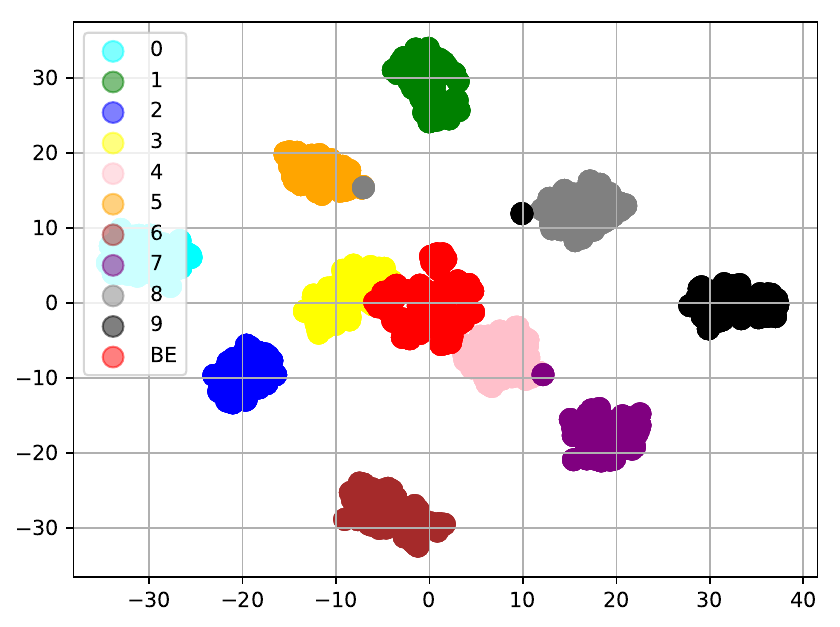}
		\end{minipage}
	}
	\caption{ t-SNE-based visualization of the activation space on the MNIST classifier. (Left) the activation space of adversarial attack on a benign model. (Right) the activation space of backdoor attack on an infected model.}
	\label{fig_AS_tsne}    
\end{figure}

	
	\begin{figure*}[!t]
		\centering
		\vspace{-5pt}
		\subfloat
		{
			\begin{minipage}[t]{0.24\linewidth}
				\label{mnist_adv_switch}
				\centering
				\includegraphics[width=1\textwidth]{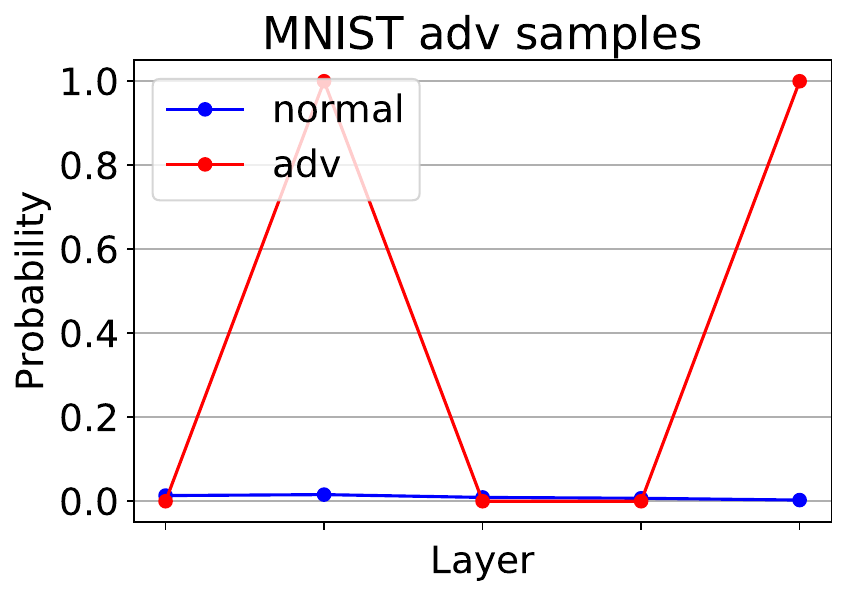}            
			\end{minipage}
		}
		\subfloat
		{
			\begin{minipage}[t]{0.24\linewidth}
				\label{gtsrb_adv_switch}
				\centering
				\includegraphics[width=1\textwidth]{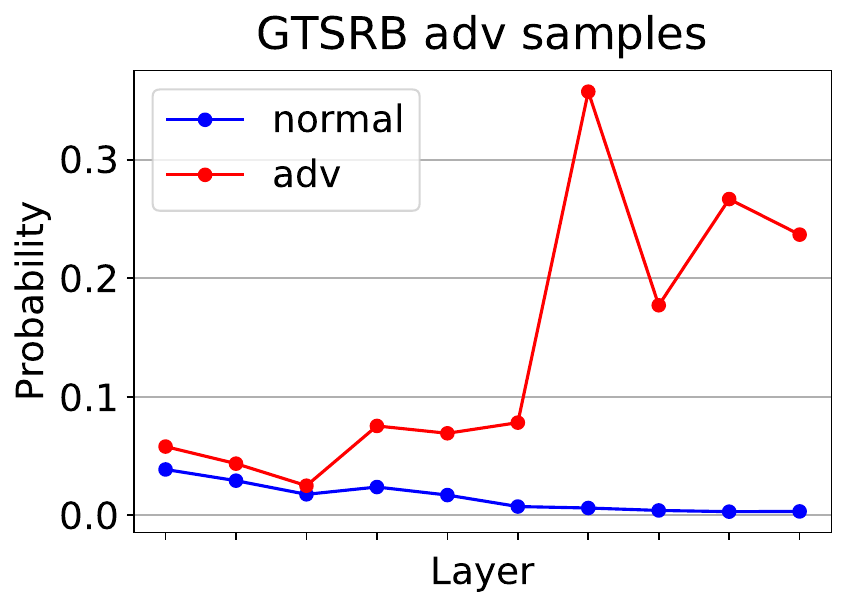}      
			\end{minipage}
		}
		\subfloat
		{
			\begin{minipage}[t]{0.24\linewidth}
				\label{face_wm_adv_switch}
				\centering
				\includegraphics[width=1\textwidth]{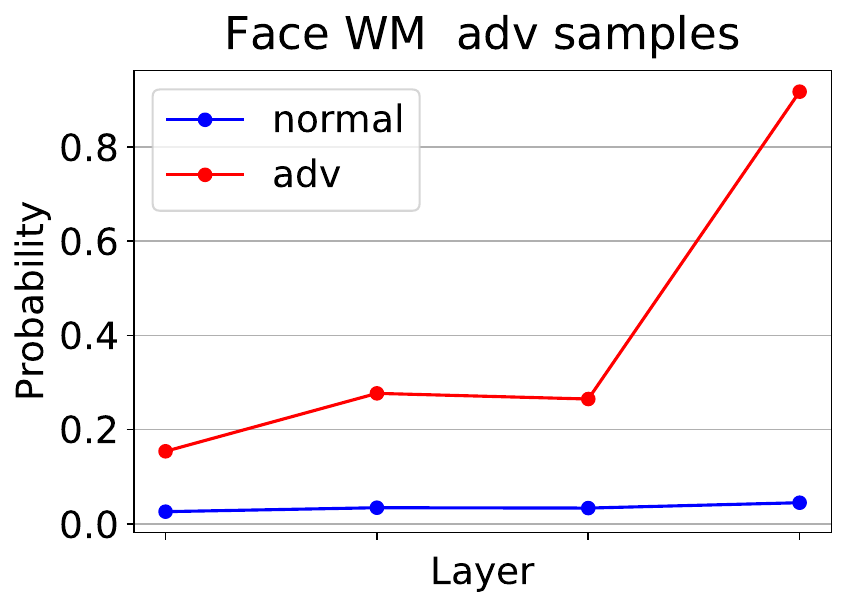}            
			\end{minipage}
		}
		\subfloat
		{
			\begin{minipage}[t]{0.24\linewidth}
				\label{face_square_adv_switch}
				\centering
				\includegraphics[width=1\textwidth]{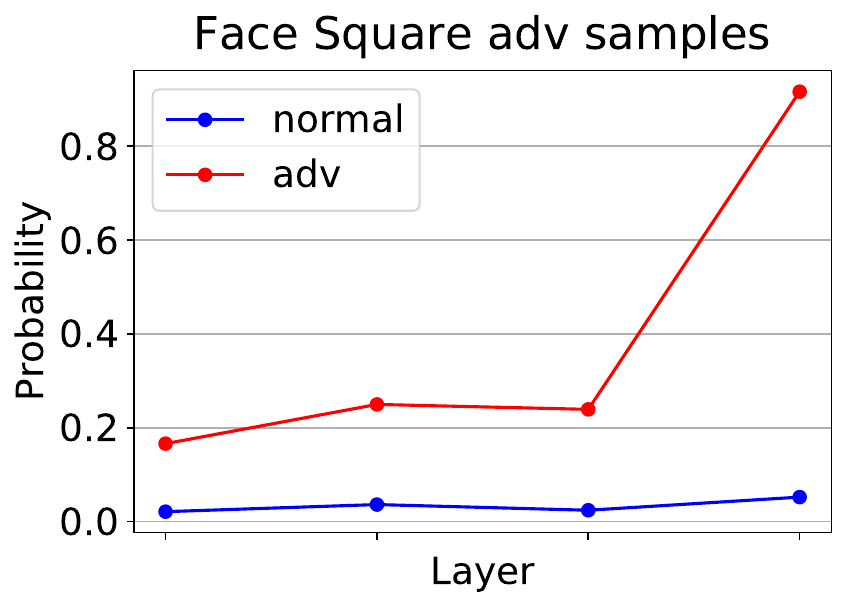}      
			\end{minipage}
		}
		\hfill
		\subfloat{
			\begin{minipage}[t]{0.24\linewidth}
				\label{mnist_backdoor_switch}
				\centering
				\includegraphics[width=1\textwidth]{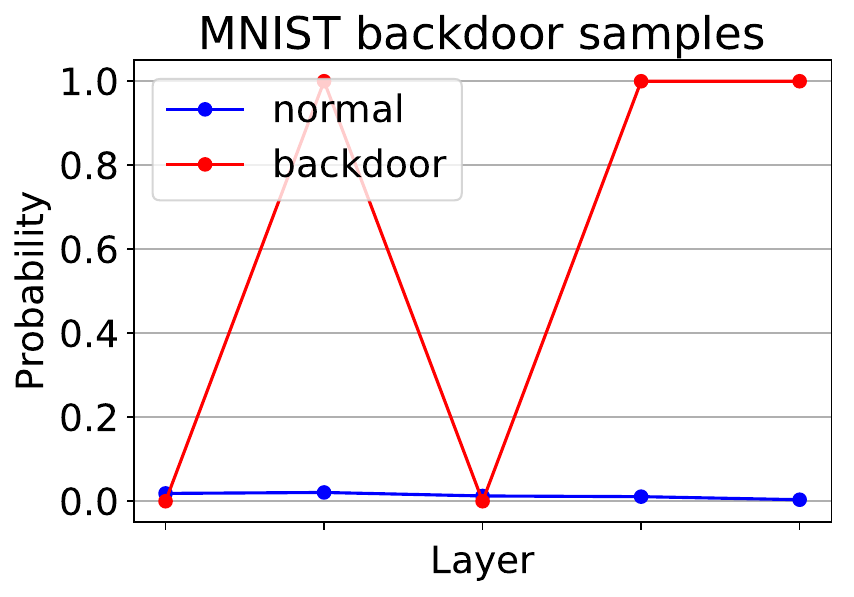}
			\end{minipage}
		}
		\subfloat{
			\begin{minipage}[t]{0.24\linewidth}
				\label{gtsrb_backdoor_switch}
				\centering
				\includegraphics[width=1\textwidth]{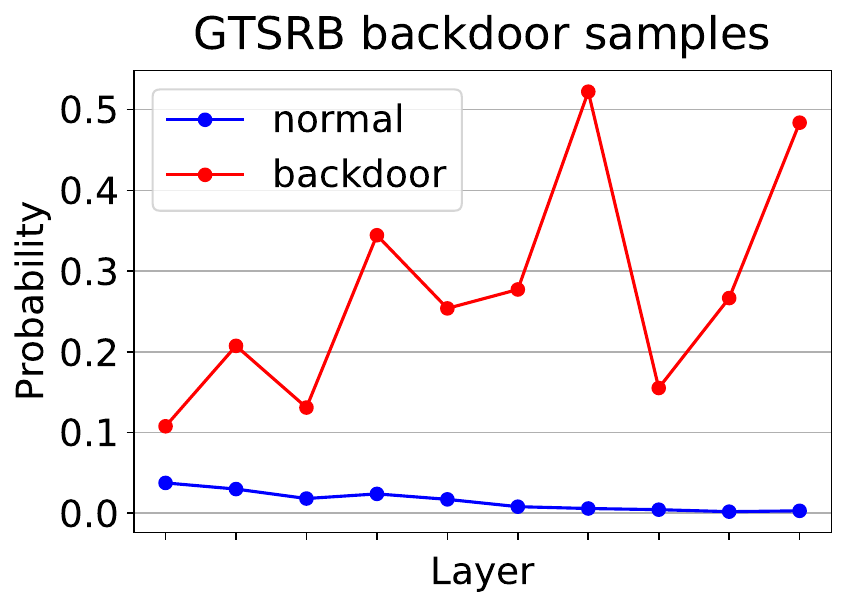}
			\end{minipage}		
		}
		\subfloat{
			\begin{minipage}[t]{0.24\linewidth}
				\label{face_wm_backdoor_switch}
				\centering
				\includegraphics[width=1\textwidth]{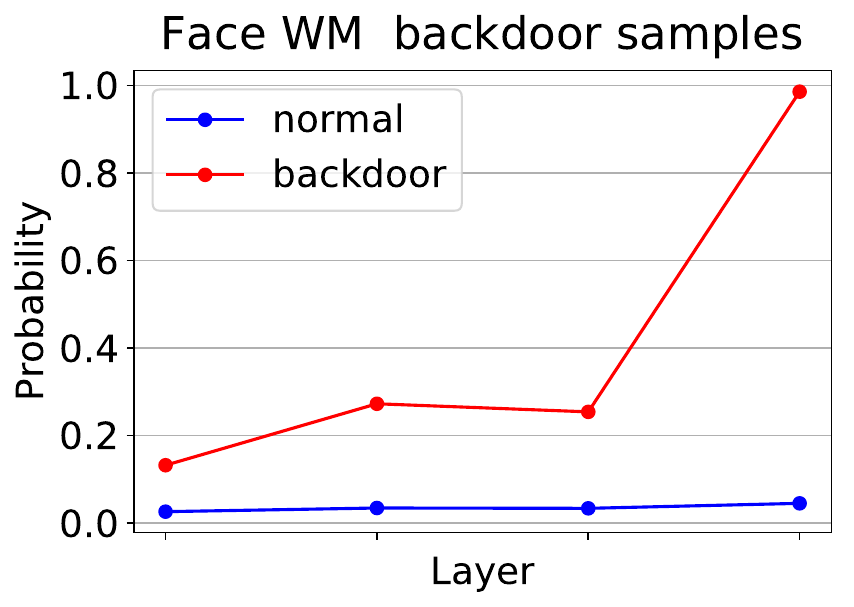}
			\end{minipage}
		}
		\subfloat{
			\begin{minipage}[t]{0.24\linewidth}
				\label{face_square_backdoor_switch}
				\centering
				\includegraphics[width=1\textwidth]{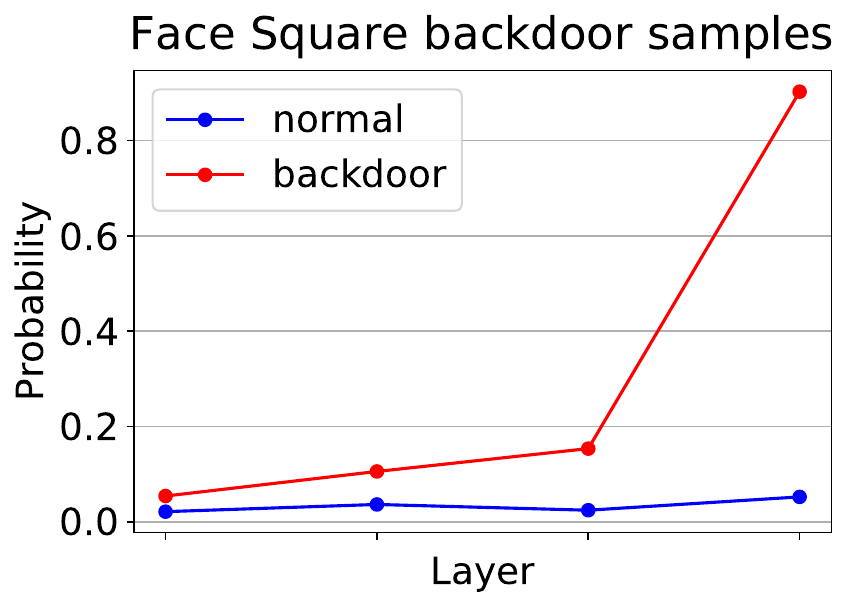}
			\end{minipage}		
		}
		\caption{Label switching probability of normal samples, AEs and BEs.}
		\label{fig:switch_prob}    	
	\end{figure*}

	\begin{figure*}[!t]
		\centering
		\vspace{-5pt}
		\subfloat
		{
			\begin{minipage}[t]{0.23\linewidth}
				\label{mnist_kd_score}
				\centering
				\includegraphics[width=1\textwidth]{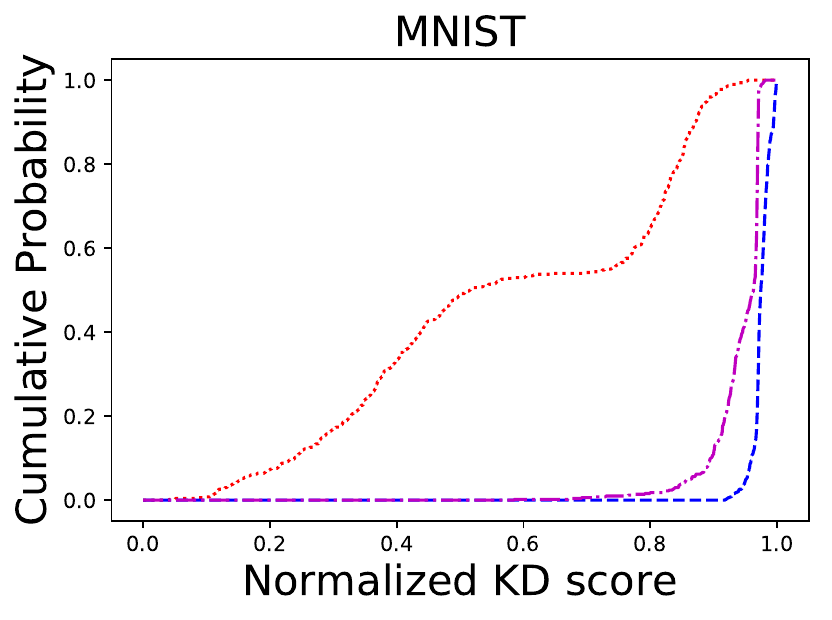}            
			\end{minipage}
		}
		\subfloat{
			\begin{minipage}[t]{0.23\linewidth}
				\label{gtsrb_kd_score}
				\centering
				\includegraphics[width=1\textwidth]{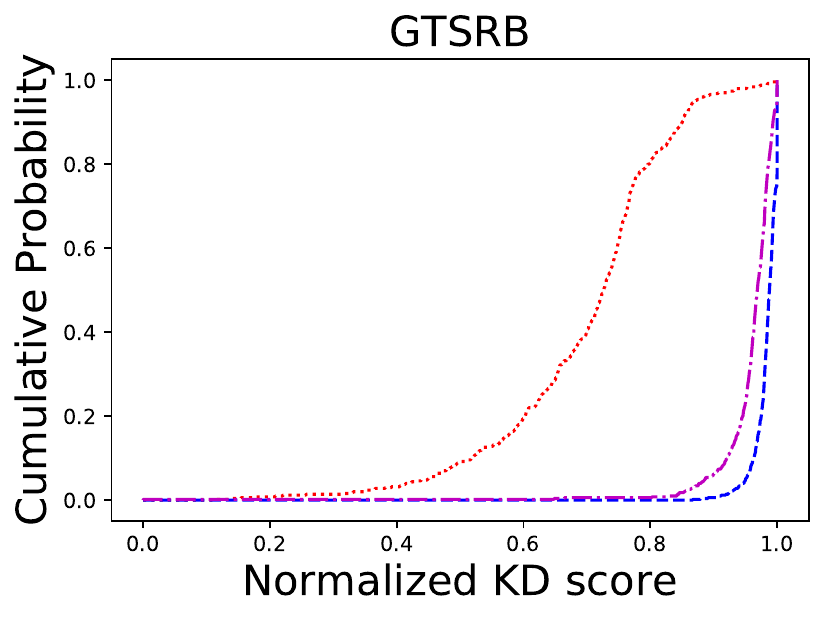}
			\end{minipage}
		}
		\subfloat
		{
			\begin{minipage}[t]{0.23\linewidth}
				\label{face_wm_kd_score}
				\centering
				\includegraphics[width=1\textwidth]{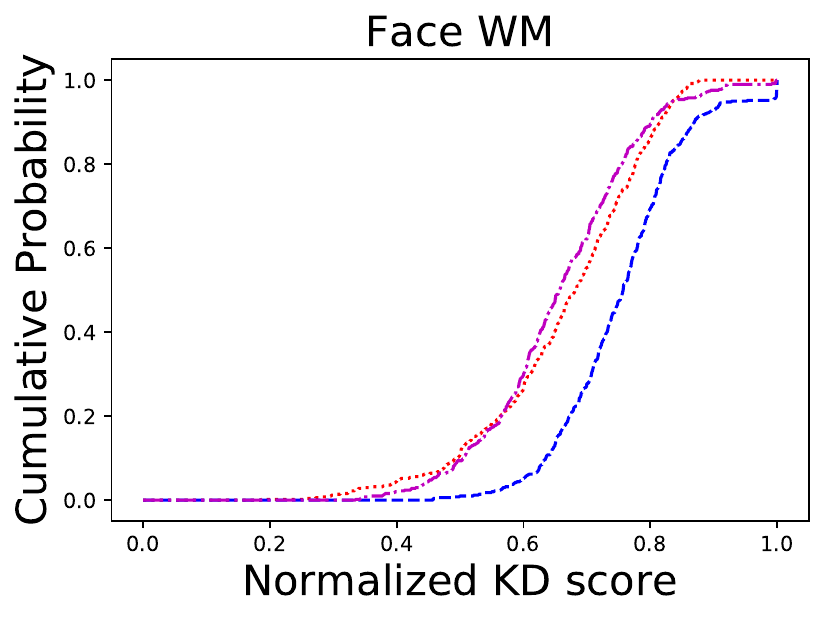}      
			\end{minipage}
		}
		\subfloat{
			\begin{minipage}[t]{0.23\linewidth}
				\label{face_square_kd_score}
				\centering
				\includegraphics[width=1\textwidth]{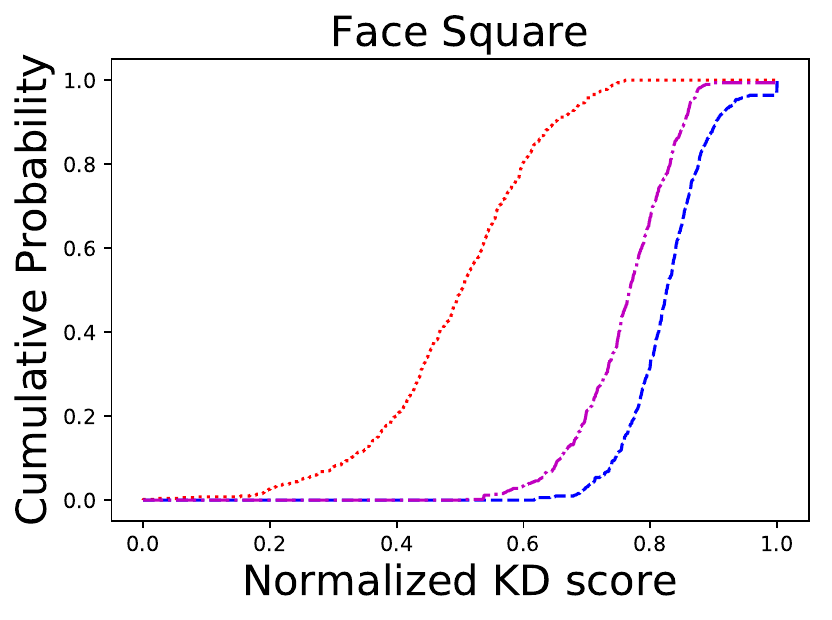}
			\end{minipage}		
		}
		\hfill
		\subfloat
		{
			\begin{minipage}[t]{0.23\linewidth}
				\label{mnist_lid_score}
				\centering
				\includegraphics[width=1\textwidth]{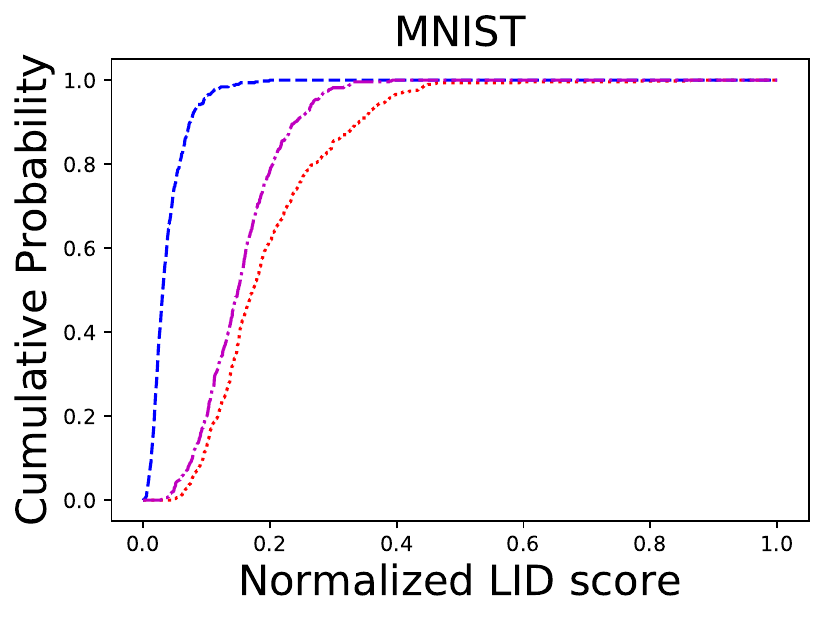}            
			\end{minipage}
		}
		\subfloat{
			\begin{minipage}[t]{0.23\linewidth}
				\label{gtsrb_lid_score}
				\centering
				\includegraphics[width=1\textwidth]{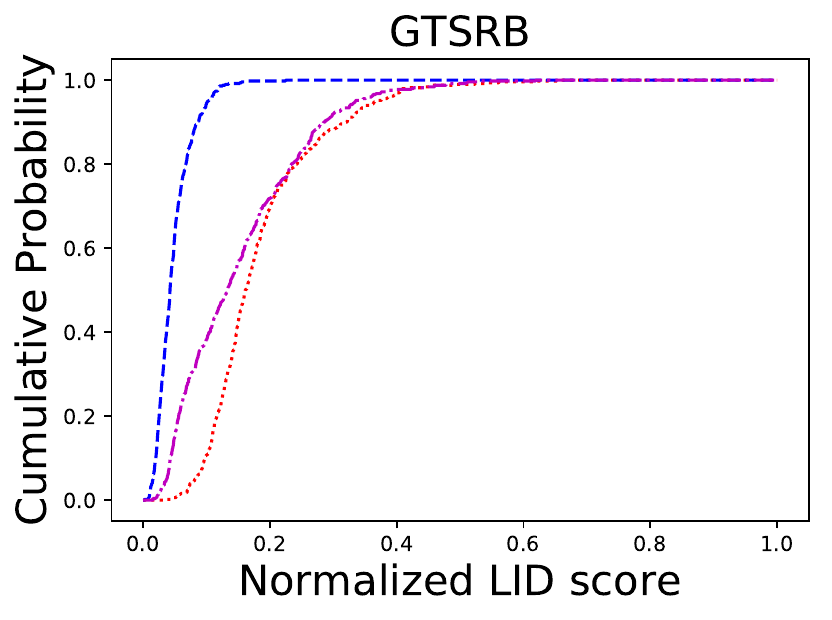}
			\end{minipage}
		}
		\subfloat
		{
			\begin{minipage}[t]{0.23\linewidth}
				\label{face_wm_lid_score}
				\centering
				\includegraphics[width=1\textwidth]{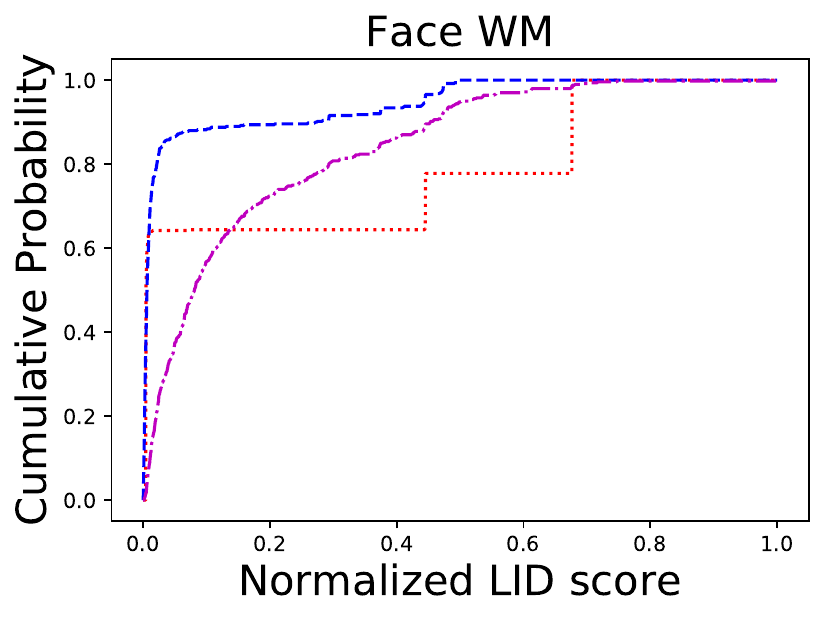}      
			\end{minipage}
		}
		\subfloat{
			\begin{minipage}[t]{0.23\linewidth}
				\label{face_square_lid_score}
				\centering
				\includegraphics[width=1\textwidth]{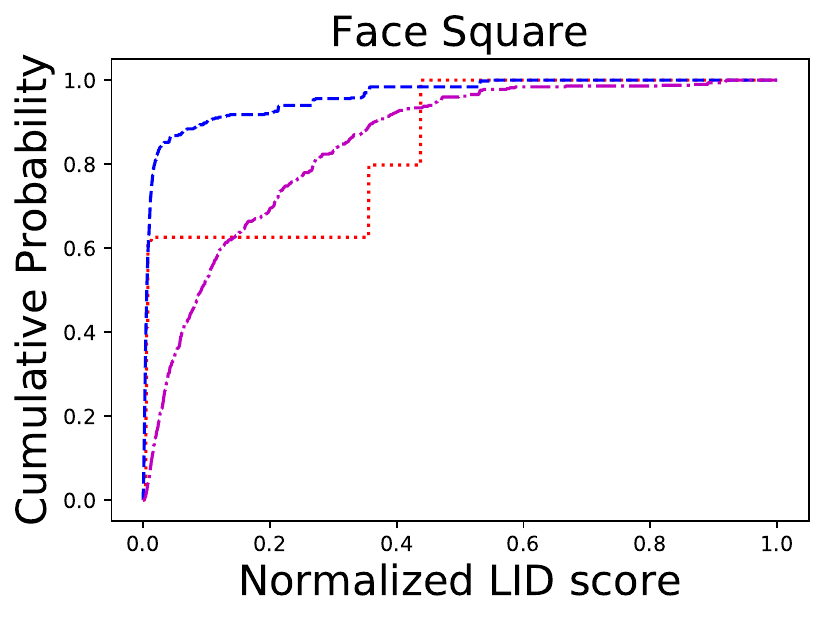}
			\end{minipage}		
		}
		\hfill
		\subfloat{
			\begin{minipage}[t]{0.5\linewidth}
				\label{legend_2}
				\centering
				\includegraphics[width=0.6\textwidth]{legend.png}
			\end{minipage}
		}
		\vspace{-5pt}
		\caption{Cumulative probability distribution of KD (first row) and LID (second row) values.}
		\label{fig:KdLidScore}    	
	\end{figure*}

Next, we consider the anomaly detection in the activation space. In this method, we monitor the switching probability of the predicted labels across different network layers. Figure \ref{fig:switch_prob} shows the results for different datasets (the first row is the comparison between normal samples and AEs; the second row is the comparison between normal samples and BEs). We get two observations. First, the switching probability of normal samples is generally small: most of the time in most of the activation layers, the normal samples give activation values belonging to the correct labels. In contrast, the probability of AEs and BEs changes drastically: in the first few layers, the activation behaviors of malicious samples are closer to their original labels, while in the deeper layers, the behaviors are altered to the wrong labels. This high switching probability serves as the indicator of AEs and BEs. Second, AEs and BEs have similar behaviors in the activation space. It is very hard to distinguish them using this method.

\begin{center}
\noindent \fbox{\parbox{85mm}
	{\textbf{Remark 2}: \emph{BEs and AEs have similar behaviors in the activation space, which are different from normal samples.}}
}
\end{center}

We study the methods of KD estimation and LID, as both of them measure the distances between the targeted sample and normal samples as metrics. Figure \ref{fig:KdLidScore} shows the cumulative probability distribution of normalized KD and LID values. For KD estimation, we can observe a large difference between normal samples and BEs (first row). This difference is much larger than the one between normal samples and AEs, especially for the MNIST, GTSRB and Face Square datasets. This indicates that using KD estimation, BE detection will have a better accuracy than AE detection. This will be further validated in Section \ref{sec:us-eff} and Table \ref{AUC_detection_result}.  
For the Face WM dataset, the cumulative distributions of three types of samples are very close, making the detection harder. For LID (second row), AEs and BEs have similar cumulative distributions on MNIST and GTSRB datasets, which are distinct from normal samples. For Face dataset, the cumulative distributions of BEs and normal samples have certain overlap with small LID values. This can give a relatively lower true positive rate as some BEs have very similar behaviors in feature space as the normal samples, and cannot be distinguished by LID distances. 

Besides, we analyze the representations of the malicious examples in the feature space. We use t-SNE to project the feature space into two principal components. Figure \ref{fig_AS_tsne} shows the t-sne visualization of the last layer of the feature space in the MNIST dataset. Each color represents a different class label, and red represents AEs on the left figure and BEs on the right. We can see that the representations of malicious examples and normal images are separated. The activations of BEs are completely separated into one cluster.

\begin{center}
\noindent \fbox{\parbox{85mm}
	{\textbf{Remark 3}: \emph{Both BEs and AEs have significant differences from normal samples in the feature space. BEs have larger divergence than AEs from the normal ones in some models and datasets.}}
}
\end{center}

\subsection{Usability versus Effectiveness}
\label{sec:us-eff}

\begin{figure*}[!t]
	\centering
	\subfloat
	{
		\begin{minipage}[t]{0.23\linewidth}
			\label{mnist_adv_AUC}
			\centering
			\includegraphics[width=1\textwidth]{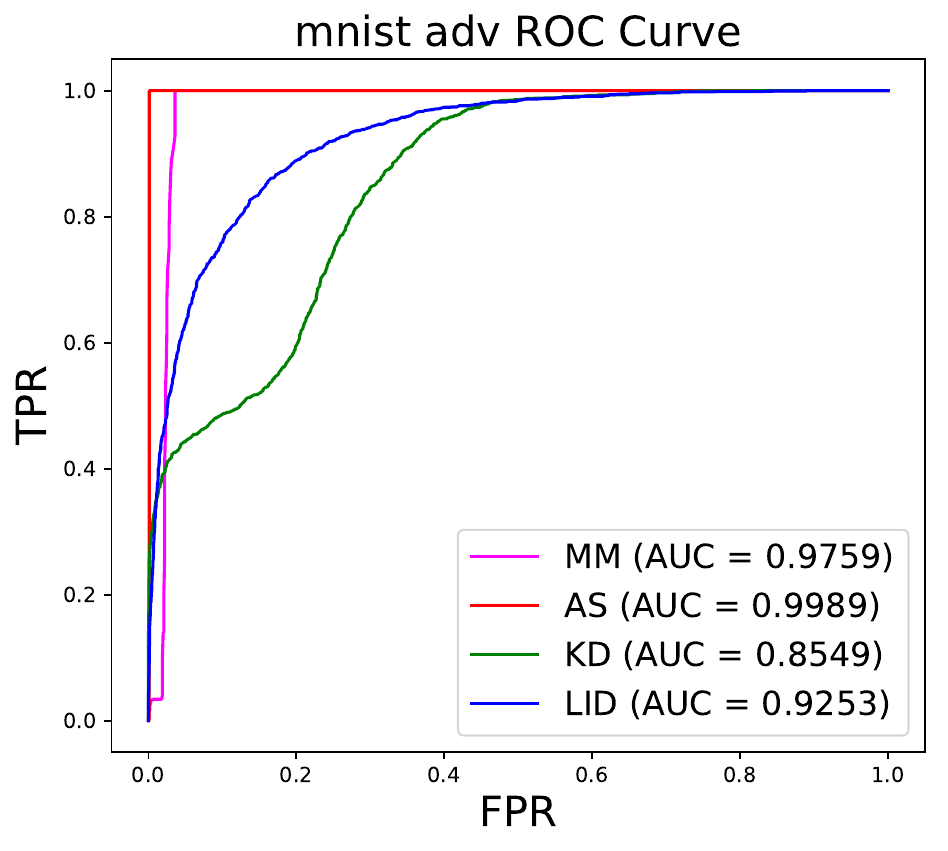}            
		\end{minipage}
	}
	\subfloat
	{
		\begin{minipage}[t]{0.23\linewidth}
			\label{gtsrb_adv_AUC}
			\centering
			\includegraphics[width=1\textwidth]{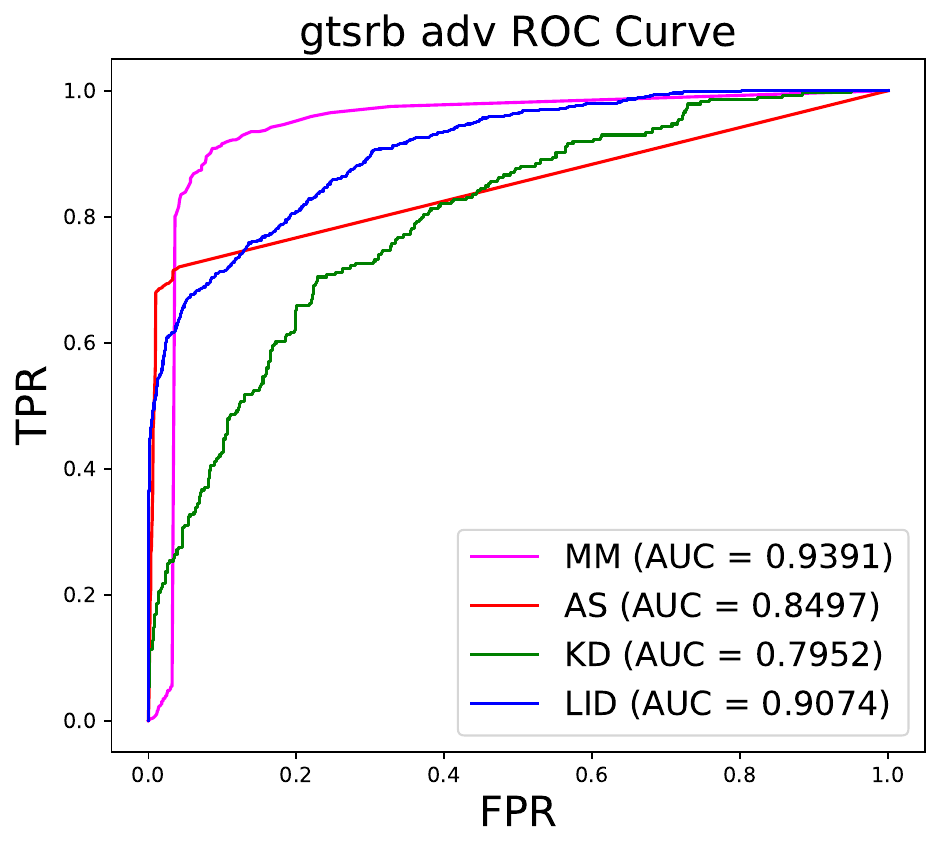}      
		\end{minipage}
	}
	\subfloat
	{
		\begin{minipage}[t]{0.23\linewidth}
			\label{face_wm_adv_AUC}
			\centering
			\includegraphics[width=1\textwidth]{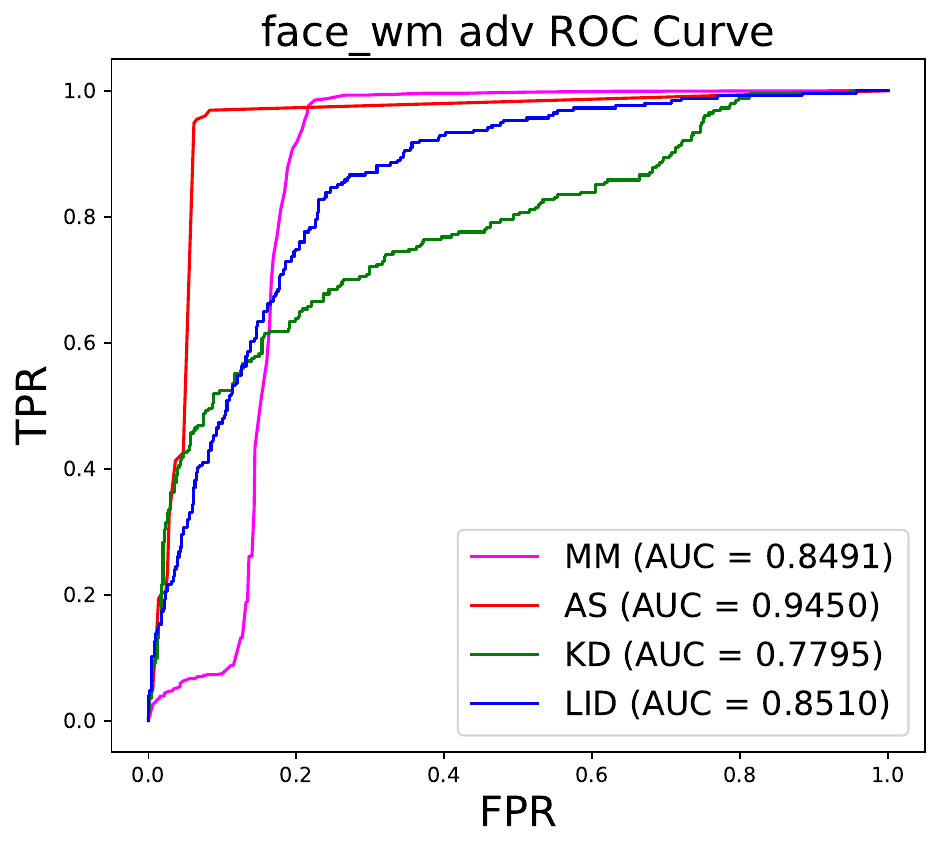}            
		\end{minipage}
	}
	\subfloat
	{
		\begin{minipage}[t]{0.23\linewidth}
			\label{face_square_adv_AUC}
			\centering
			\includegraphics[width=1\textwidth]{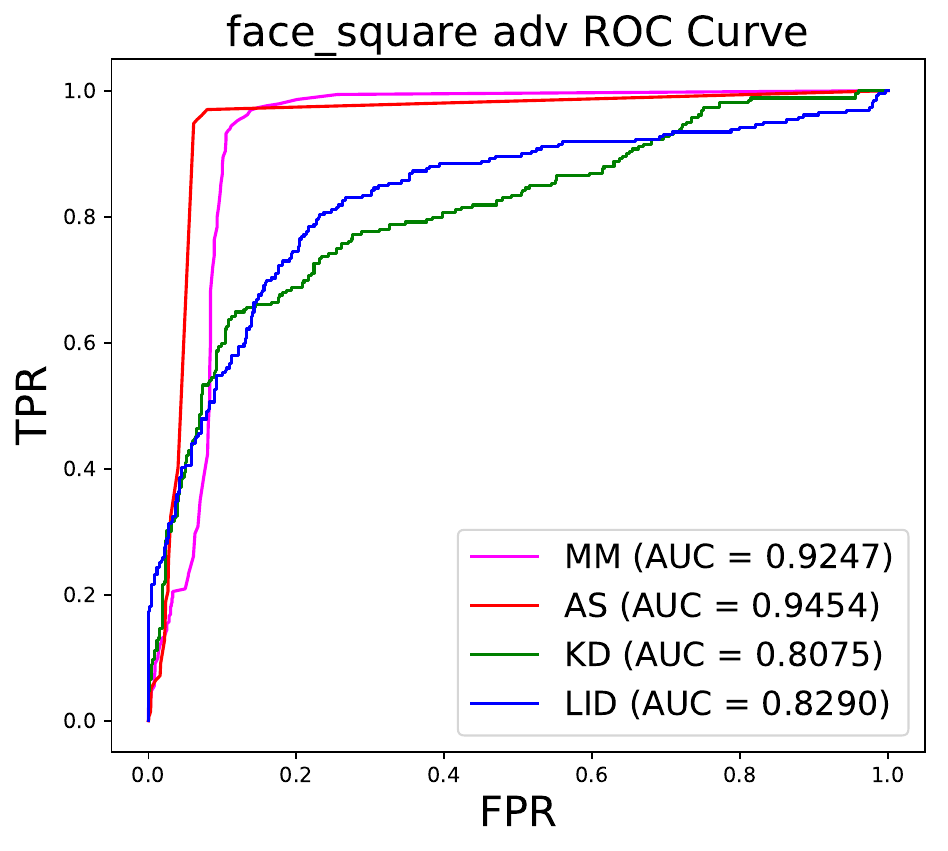}      
		\end{minipage}
	}
	\hfill
	\subfloat{
	\begin{minipage}[t]{0.23\linewidth}
		\label{mnist_backdoor_AUC}
		\centering
		\includegraphics[width=1\textwidth]{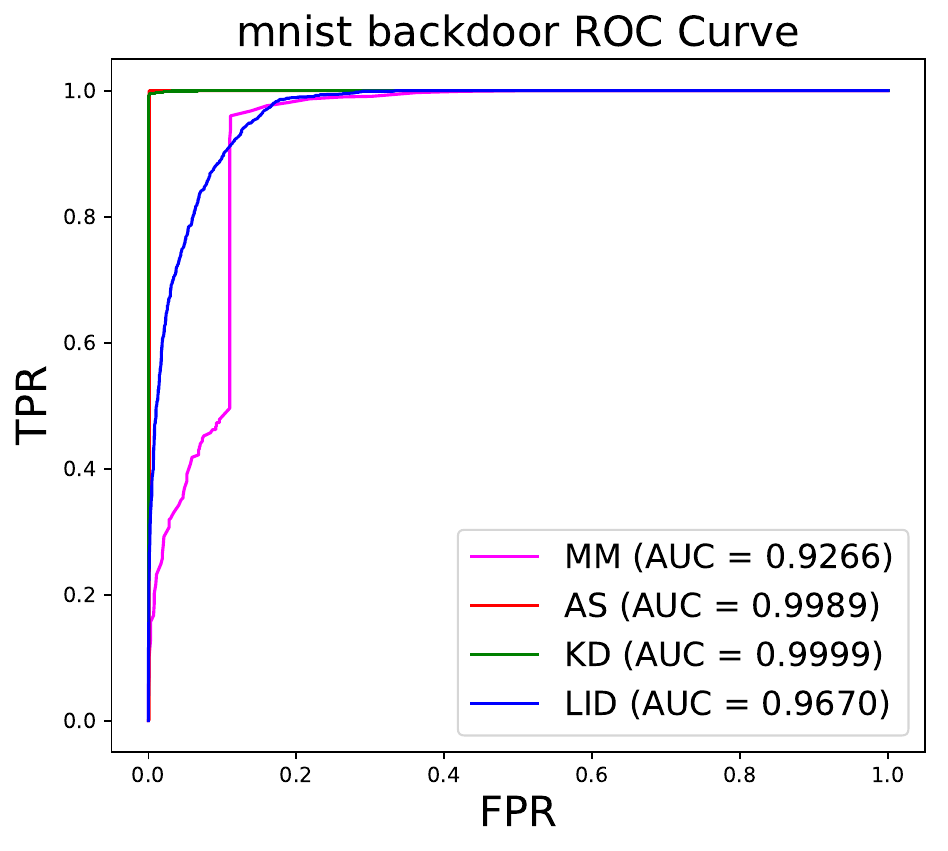}
	\end{minipage}
	}
	\subfloat{
	\begin{minipage}[t]{0.23\linewidth}
		\label{gtsrb_backdoor_AUC}
		\centering
		\includegraphics[width=1\textwidth]{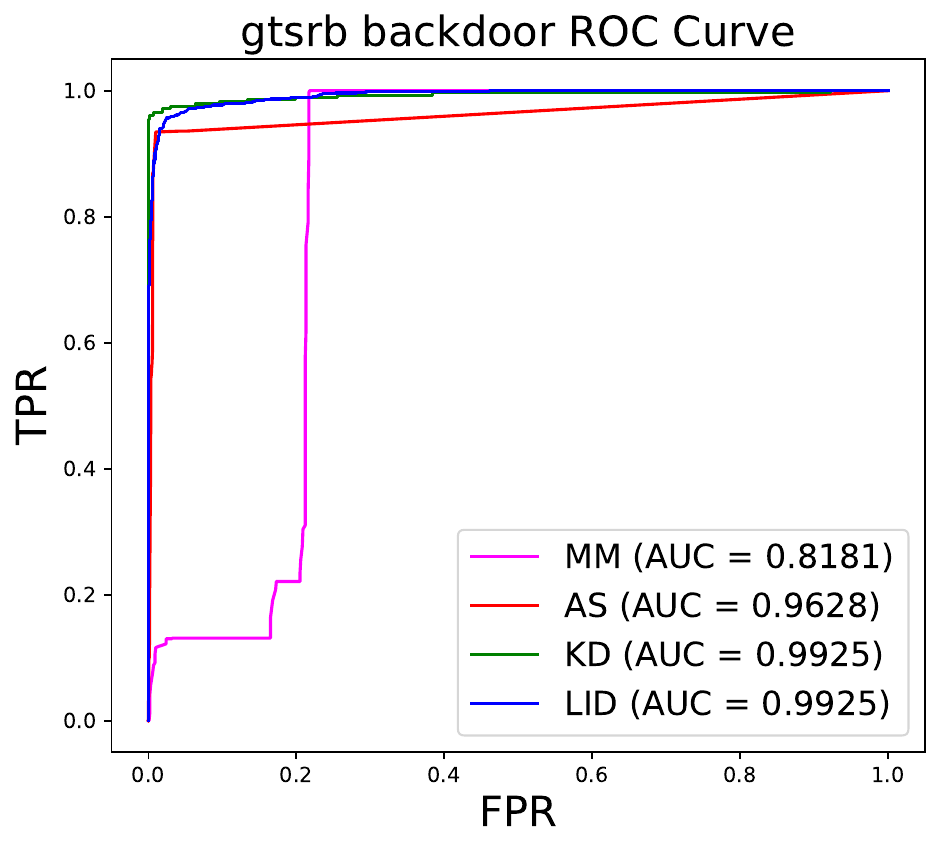}
	\end{minipage}		
	}
	\subfloat{
		\begin{minipage}[t]{0.23\linewidth}
			\label{face_wm_backdoor_AUC}
			\centering
			\includegraphics[width=1\textwidth]{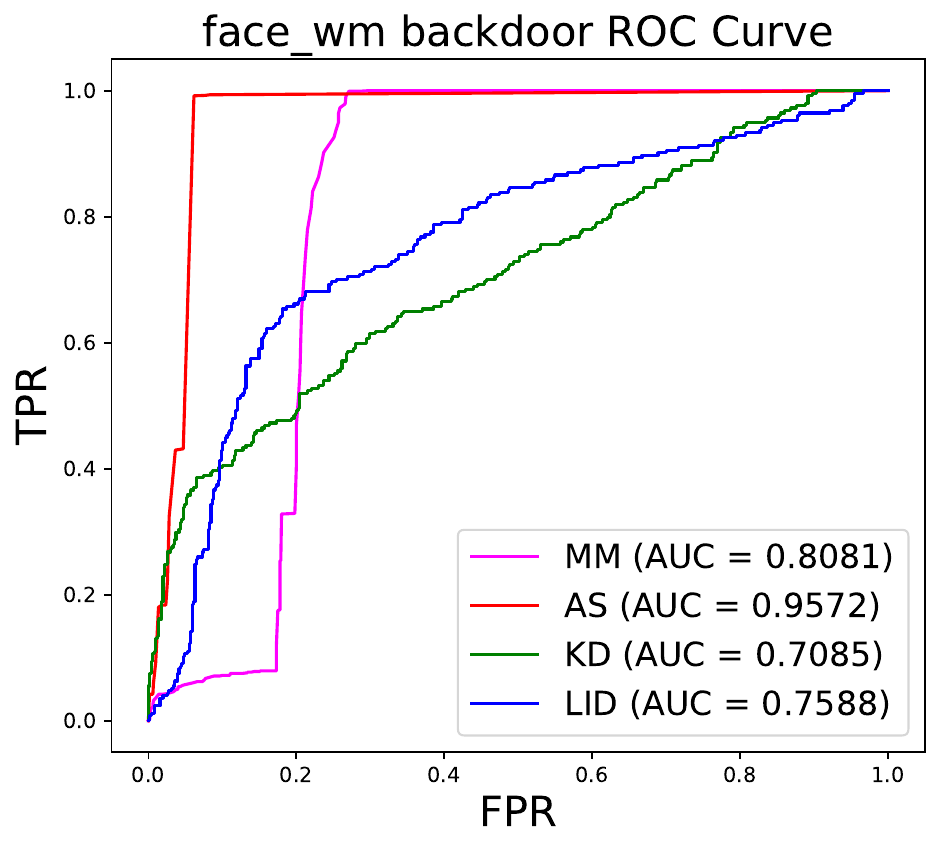}
		\end{minipage}
	}
	\subfloat{
		\begin{minipage}[t]{0.23\linewidth}
			\label{face_square_backdoor_AUC}
			\centering
			\includegraphics[width=1\textwidth]{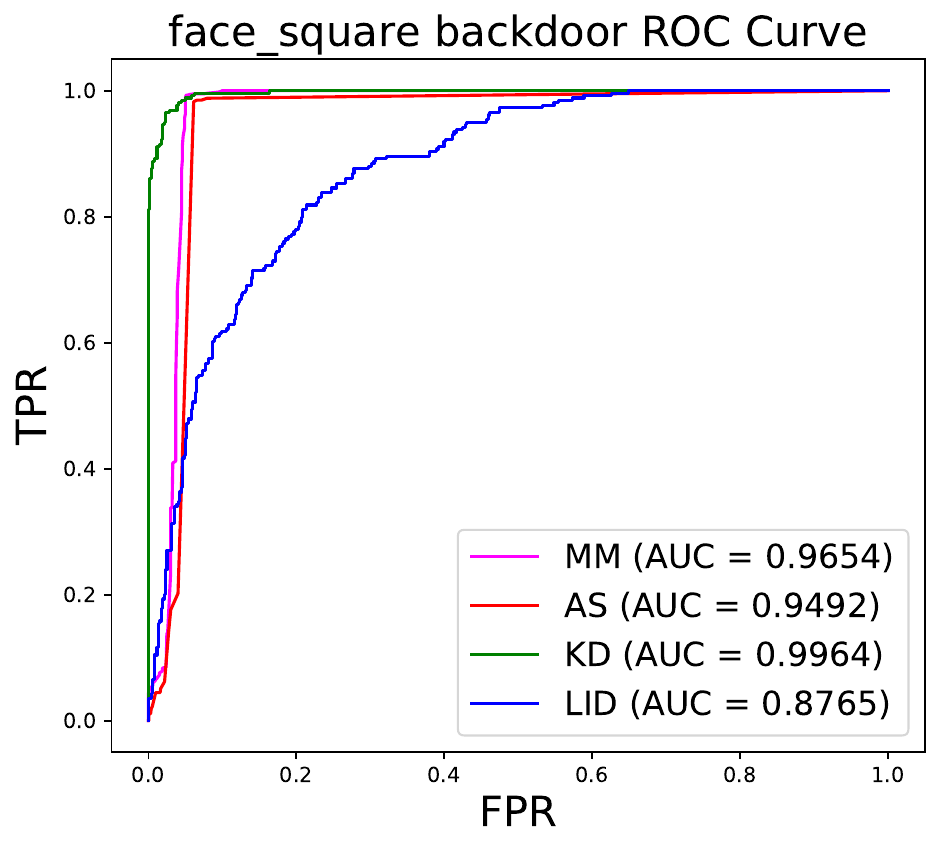}
		\end{minipage}		
	}
	\vspace{-5pt}
	\caption{ROC curve for detecting adversarial examples (first row) and backdoor examples (second row).}
	\label{fig:auc_result}    	
\end{figure*}

Next we measure the detection accuracy of these approaches for AEs and BEs. We consider both the true and false positive rates. We choose different threshold parameters in these approaches and draw the ROC curve, as shown in Figure \ref{fig:auc_result}. The corresponding AUC (Area Under the Curve) scores are summarized in Table \ref{AUC_detection_result}. 

	\begin{table}[ht]
		\small
		\caption{The AUC score result. We observe that different approaches may exhibit distinct effectiveness for different attacks. The best approach for each attack is highlighted in bold. In most cases, AS and KD give the best performance.}
		\vspace{-8pt}
		\begin{center}
			\renewcommand{\multirowsetup}{\centering}
			\resizebox{\linewidth}{!}{
			\begin{tabular}{c c c c c c}
				\toprule
				Dataset & Attack & MM & AS & KD & LID \\
				\midrule
				\multirow{2}{*}{\makecell{MNIST}} & C\&W   &  0.9759 & \textbf{0.9989} &  0.8549 & 0.9253 \\
				& Backdoor  & 0.9266 & 0.9989 &  \textbf{0.9999} & 0.9670 \\
				\hline
				\multirow{2}{*}{\makecell{GTSRB}} & C\&W  & \textbf{0.9391}  & 0.8497 &  0.7952 & 0.9074 \\
				& Backdoor  & 0.8181 & 0.9628 &  \textbf{0.9925} & \textbf{0.9925} \\
				\hline
				\multirow{2}{*}{\makecell{Face WM}} & C\&W & 0.8491     & \textbf{0.9450} &  0.7795 & 0.8510 \\
				& Backdoor & 0.8081 & \textbf{0.9572} &  0.7085 & 0.7588 \\
				\hline
				\multirow{2}{*}{\makecell{Face Square}} & C\&W & 0.9247 & \textbf{0.9454} &  0.8075 & 0.8290 \\
				& Backdoor & 0.9654 & 0.9492 &  \textbf{0.9964} & 0.8765 \\
				\bottomrule
			\end{tabular}}
			\label{AUC_detection_result}
		\end{center}
		\vspace{-5pt}
	\end{table} 	
	
We can observe that most approaches are effective at detecting both types of malicious samples with very high AUC scores. Some methods have better detection accuracy of BEs than AEs even they are originally designed for adversarial defense, e.g., KD and LID for MNIST, GTSRB and Face Square. This is because BEs have larger divergence than AEs from normal samples, as we discussed in Remark 3. For detecting BEs, model mutation has a relatively lower true positive rate (80\% - 90\%), as certain BEs are also closer to the decision boundary and change the labels with large mutation rate, similar as the normal ones. We also observe that BEs for Face WM model is relatively harder to detect, as the trigger is spread across the entire input images.

\begin{center}
	\noindent \fbox{\parbox{85mm}
		{\textbf{Remark 4}: \emph{Model Mutation, Activation Space, Kernel Density estimation and Local Intrinsic Dimensionality can effectively detect various types of BEs against different backdoor models. Some methods can achieve higher accuracy than AE detection.}}
	}
\end{center}

\subsection{Performance}
Finally we evaluate the runtime speed of those approaches. It is worth noting the performance of those methods were never considered in the original papers \cite{model_mutation,activation_spaces,detect_artifacts,detect_LID}. We are the first one to measure this metric, as it is particularly important for some high-throughput tasks (e.g., video analytic, surveillance, etc.) on resource-constrained devices. 

Table \ref{cost_time} shows the average inference time, and detection time of four methods for different models. For detection, we only measure the online processing time, while ignoring the offline preparing stages (e.g., training classifier, generating mutated models). We can observe that model mutation has the largest detection time. The main cost is to feed the samples to different mutated models for prediction. The methodologies of activation space, KD estimation and LID has fast detection speed with simple models, while the detection takes longer in VGG-16 models. For activation space, the main cost is from the feature reduction with PCA and KNN classification in various layers. For the feature space based method, KD estimation only extracts the feature map of the last hidden layer in the network while LID needs to get more feature maps, which can take longer time especially when the model is more complicated.

	\begin{table}[ht]
		\small
		\centering
		\caption{Cost time of MM, AS, KD and LID (millisecond).}
		\begin{tabular}{cccccc}
			\toprule
			\multirow{2}{*}{Datset}      & Orignal & \multirow{2}{*}{MM} & \multirow{2}{*}{AS}   & \multirow{2}{*}{KD}    & \multirow{2}{*}{LID}   \\ 
			& inference & & & & \\
			\midrule
			MNIST       & 1.5          &  230.1  & 5.7  & 2.7   & 1.8   \\
			GTSRB       & 1.6          &  245.7  & 10.4 & 3.4   & 4.9   \\
			Face WM     & 7.8          &  436.5  & 51.1 & 40.6  & 198.3 \\
			Face Square & 7.1          &  431.4  & 49.7 & 40.2 & 206.1 \\ 
			\bottomrule
		\end{tabular}
	    \label{cost_time}
	\end{table}
	
\begin{center}
	\noindent \fbox{\parbox{85mm}
		{\textbf{Remark 5}: \emph{The detection costs of these approaches are relatively large compared to the inference time. Detecting one sample can still be completed within 0.5 seconds. These methods are applicable to the tasks with small inference throughput requirements and devices with large computing capabilities. }}
	}
\end{center}

\subsection{Detection of More Advanced Attacks}
\begin{figure*}[t]
	\centering
		\subfloat[UAP]{
			\begin{minipage}[t]{0.13\linewidth}
				\label{Fig-GTSRB-INS}
				\centering
				\includegraphics[width=1\textwidth]{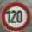}
			\end{minipage}
			\begin{minipage}[t]{0.13\linewidth}
				\label{Fig-GTSRB-UAP-INS}
				\centering
				\includegraphics[width=1\textwidth]{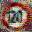}
			\end{minipage}
		}
		\hspace{5pt}
		\subfloat[IAB]{
			\begin{minipage}[t]{0.39\linewidth}
				\label{Fig-IABA-GTSRB}
				\centering
				\includegraphics[width=1\textwidth]{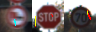}
			\end{minipage}
		}
		\hspace{5pt}
		\subfloat[HTB (source and target)]{
			\begin{minipage}[t]{0.13\linewidth}
				\label{Fig-PatchedSource}
				\centering
				\includegraphics[width=1\textwidth]{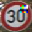}
			\end{minipage}
			\begin{minipage}[t]{0.13\linewidth}
				\label{Fig-PoisonedTarget}
				\centering
				\includegraphics[width=1\textwidth]{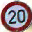}
			\end{minipage}
		}

		\caption{Malicious Examples from the advanced attacks.}
		\label{fig:INVI-attack-instances}    
	\end{figure*}

     Moreover, we conduct the evaluation on three more sophisticated attacks: (1) Universal Adversarial Perturbation Attack (UAP)~\cite{moosavi2017universal}: it adopts a universal perturbation for all normal samples to fool the target classifier; (2) Input-aware dynamic backdoor attack (IAB)~\cite{nguyen2020input}: this is an invisble backdoor attack that generates input-specific triggers. (3) Hidden Trigger Backdoor Attack (HTB)~\cite{saha2020hidden}: this generates invisible trigger to poison the training set and embed backdoors in the model. Figure \ref{fig:INVI-attack-instances} visualizes the corresponding adversarial and backdoor samples on the GTSRB dataset. Table \ref{Refool_setting} reports the performance of IAB and HTB backdoor attacks, including the accuracy of backdoor and normal samples compared to the original model accuracy.
     
	\begin{table}[ht]
		\footnotesize
		\caption{Attack success rate and classification accuracy for IAB and HTB Backdoor attacks.}
		\vspace{-8pt}
		\begin{center}
			\renewcommand{\multirowsetup}{\centering}
			\begin{tabular}{c|c|c|c|c}
				\toprule
				\multirow{3}{*}{Dataset} & \multirow{3}{*}{Attack} & \multicolumn{2}{c|}{Infected Model} & Clean Model \\
				\cline{3-5}
				& &\makecell{Attack \\Success Rate} & \makecell{Natural\\ Accuracy} & \makecell{Natural\\ Accuracy}\\
				\hline
				\multirow{2}{*}{\makecell{MNIST}} & IAB & 100\% & 99.21\% & \multirow{2}{*}{99.06\%} \\
			                                      & HTB	& 78.37\% & 98.30\% &\\
				\cline{1-5}
				\multirow{2}{*}{\makecell{GTSRB}}  & IAB & 89.88\% & 97.01\% &  \multirow{2}{*}{\makecell{96.94\%}}\\
			                                      & HTB	& 89.74\% & 96.57\% & \\
				\cline{1-5}
				\multirow{2}{*}{\makecell{PubFig}}  & IAB & 89.63\% & 89.52\% &  \multirow{2}{*}{\makecell{95.56\%}}\\
			                                      & HTB	& 80.31\% & 35.63\% & \\
				\bottomrule
			\end{tabular}
			\label{Refool_setting}
		\end{center}
		\vspace{-10pt}
	\end{table}

\begin{figure*}[!t]
	\centering
	\subfloat
	{
		\begin{minipage}[t]{0.25\linewidth}
			\label{MNIST-UAP-ROC}
			\centering
			\includegraphics[width=1\textwidth]{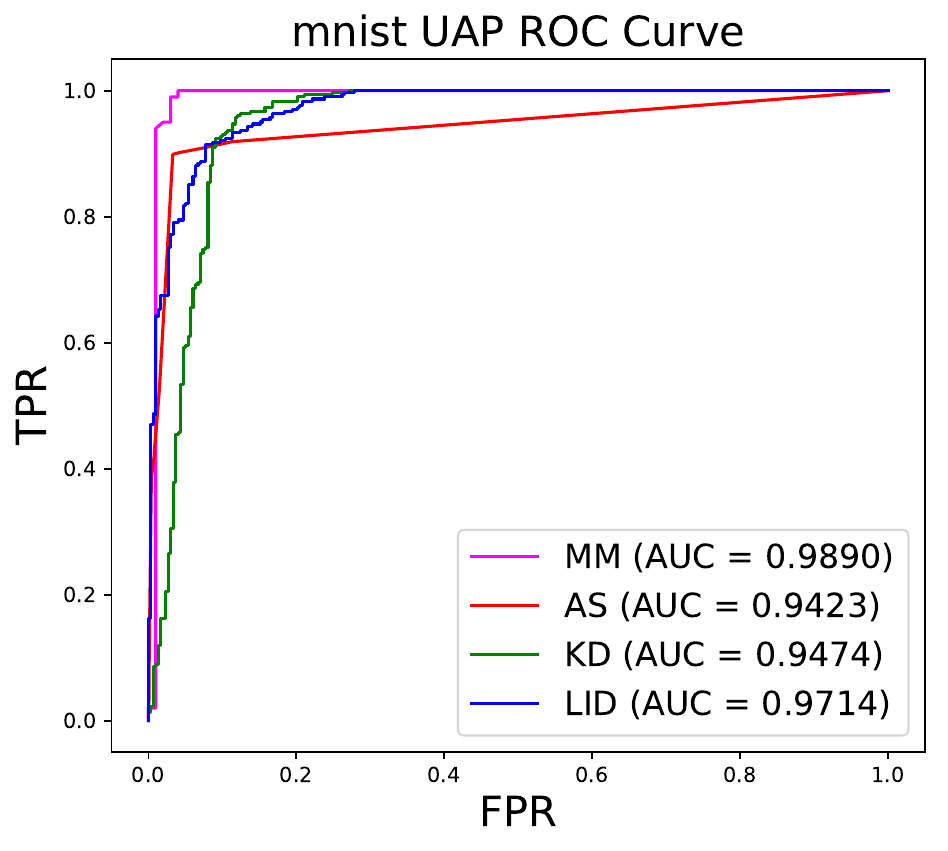}            
		\end{minipage}
	}
	\subfloat
	{
		\begin{minipage}[t]{0.25\linewidth}
			\label{MNIST-IABA-ROC}
			\centering
			\includegraphics[width=1\textwidth]{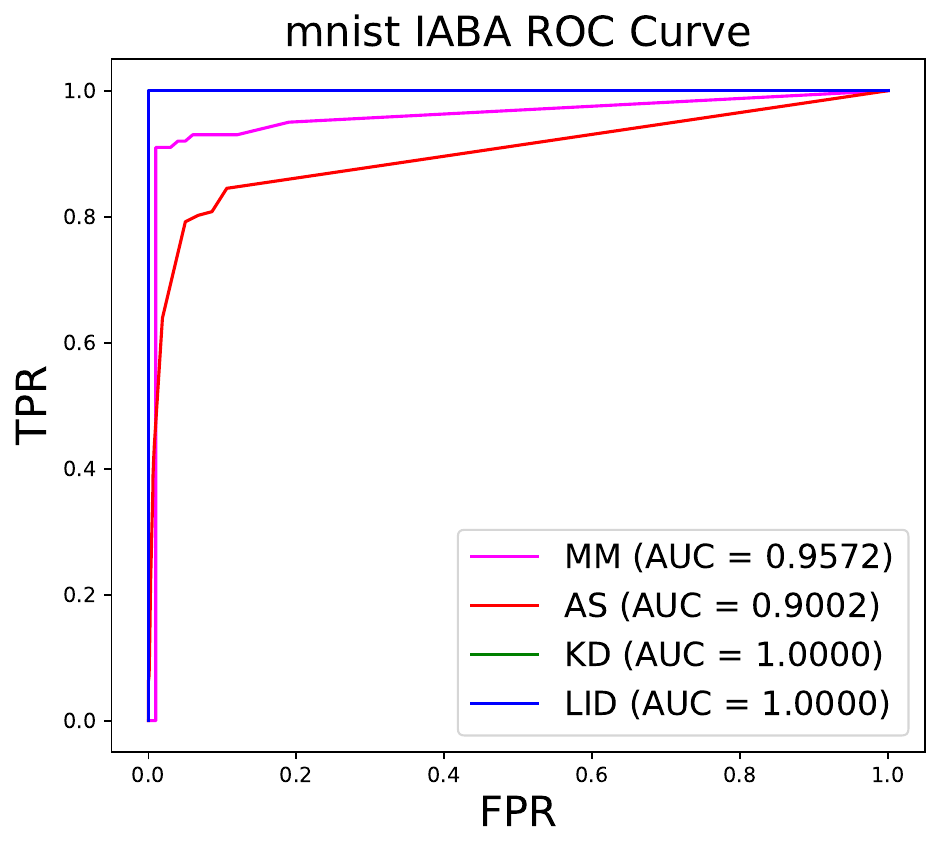}      
		\end{minipage}
	}
	\subfloat{
	\begin{minipage}[t]{0.25\linewidth}
		\label{MNIST-HTBA-ROC}
		\centering
		\includegraphics[width=1\textwidth]{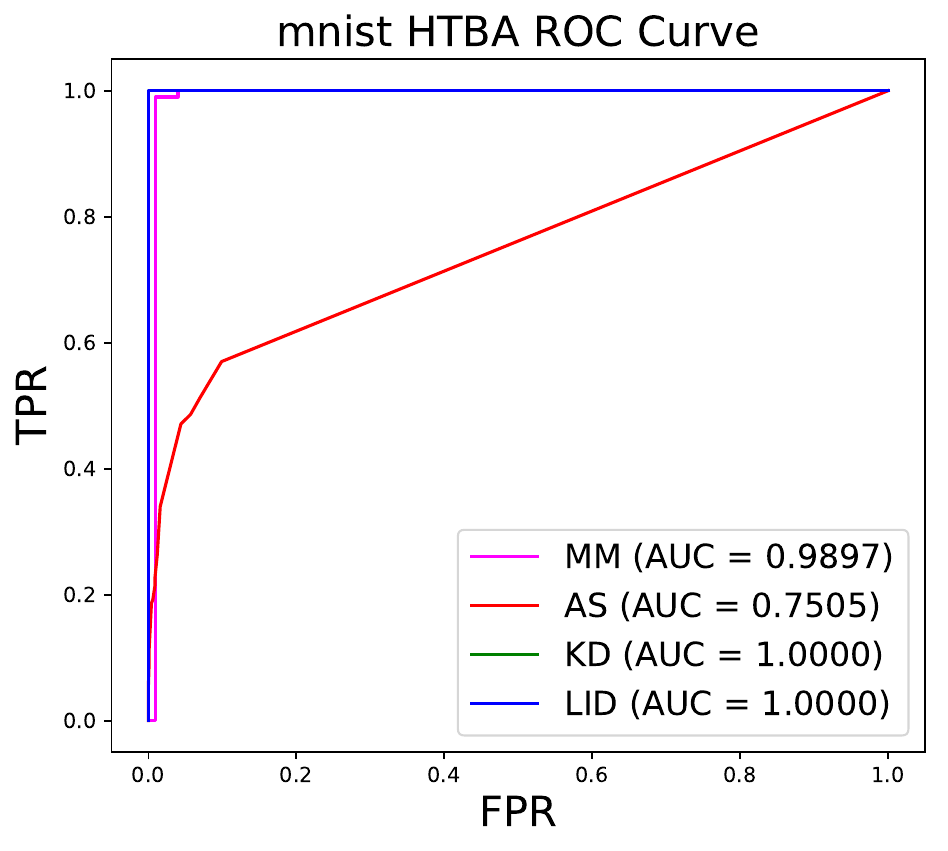}
	\end{minipage}		
	}
	\hfill
	\subfloat
	{
		\begin{minipage}[t]{0.25\linewidth}
			\label{GTSRB-UAP-ROC}
			\centering
			\includegraphics[width=1\textwidth]{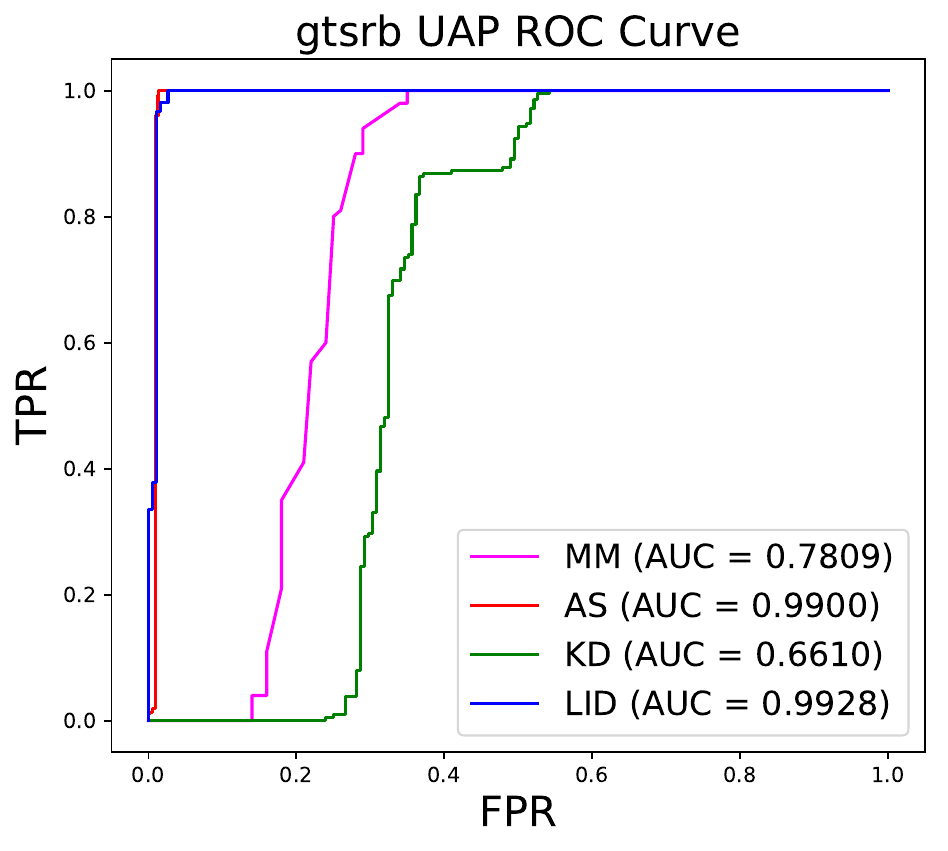}      
		\end{minipage}
	}
	\subfloat{
	\begin{minipage}[t]{0.25\linewidth}
		\label{GTSRB-IABA-ROC}
		\centering
		\includegraphics[width=1\textwidth]{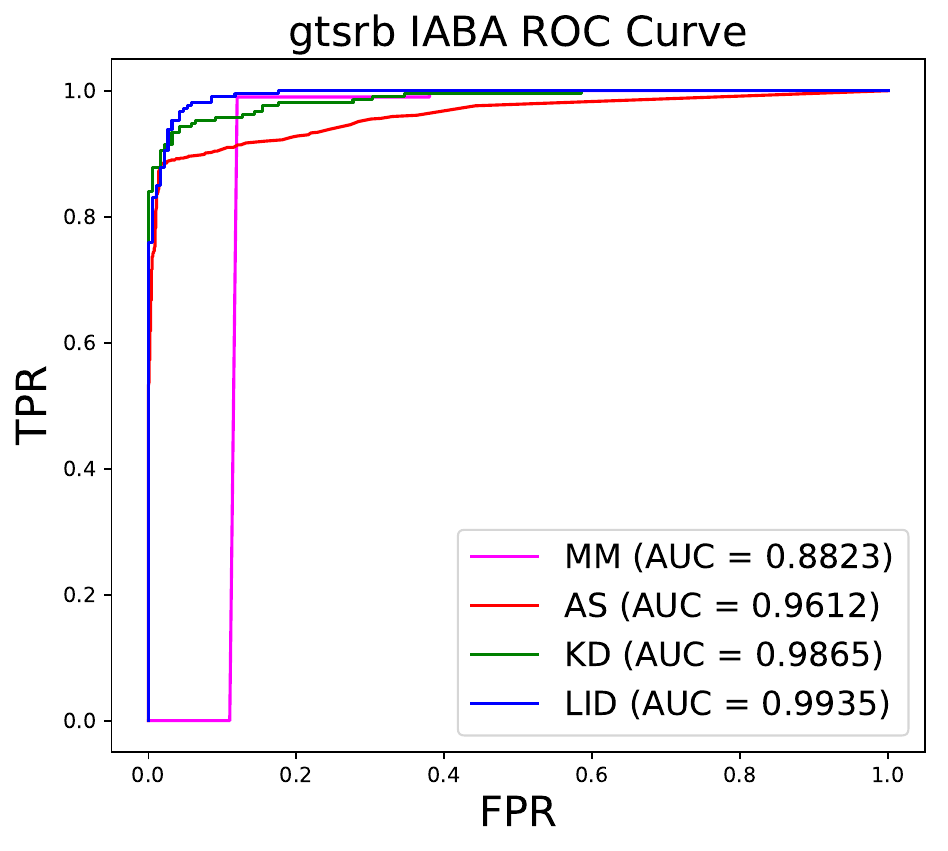}    
	\end{minipage}
	}
	\subfloat{
	\begin{minipage}[t]{0.25\linewidth}
		\label{GTSRB-HTBA-ROC}
		\centering
		\includegraphics[width=1\textwidth]{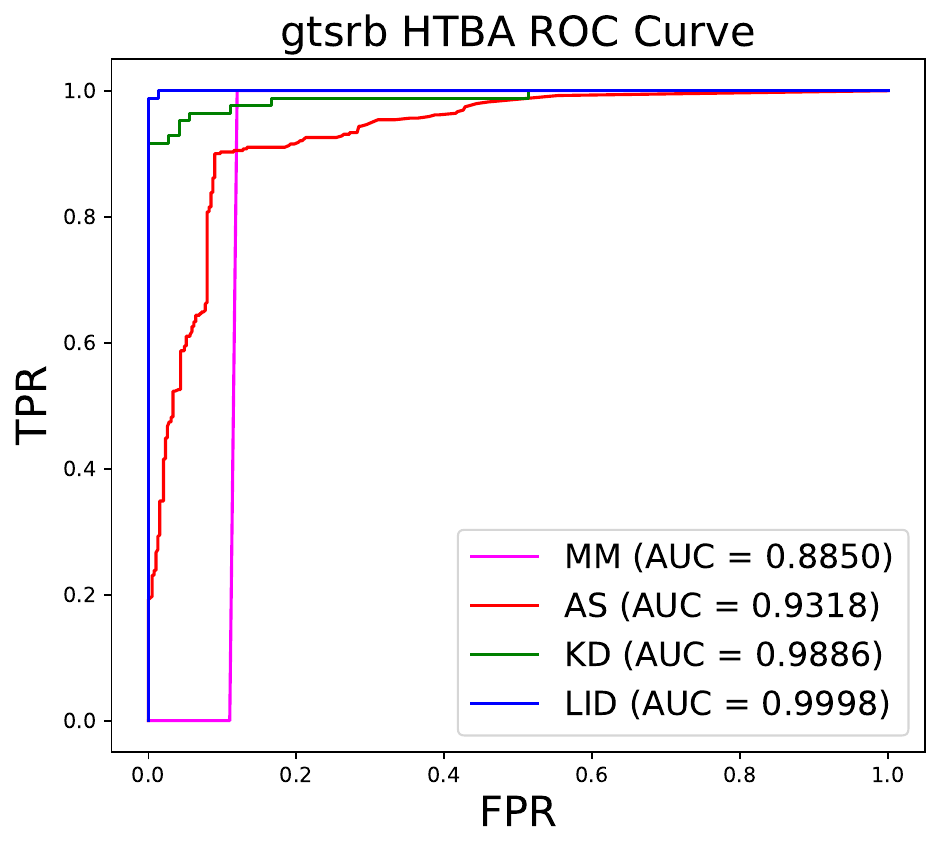}    
	\end{minipage}
	}
	\hfill
	\subfloat
	{
		\begin{minipage}[t]{0.25\linewidth}
			\label{PubFig-UAP-ROC}
			\centering
			\includegraphics[width=1\textwidth]{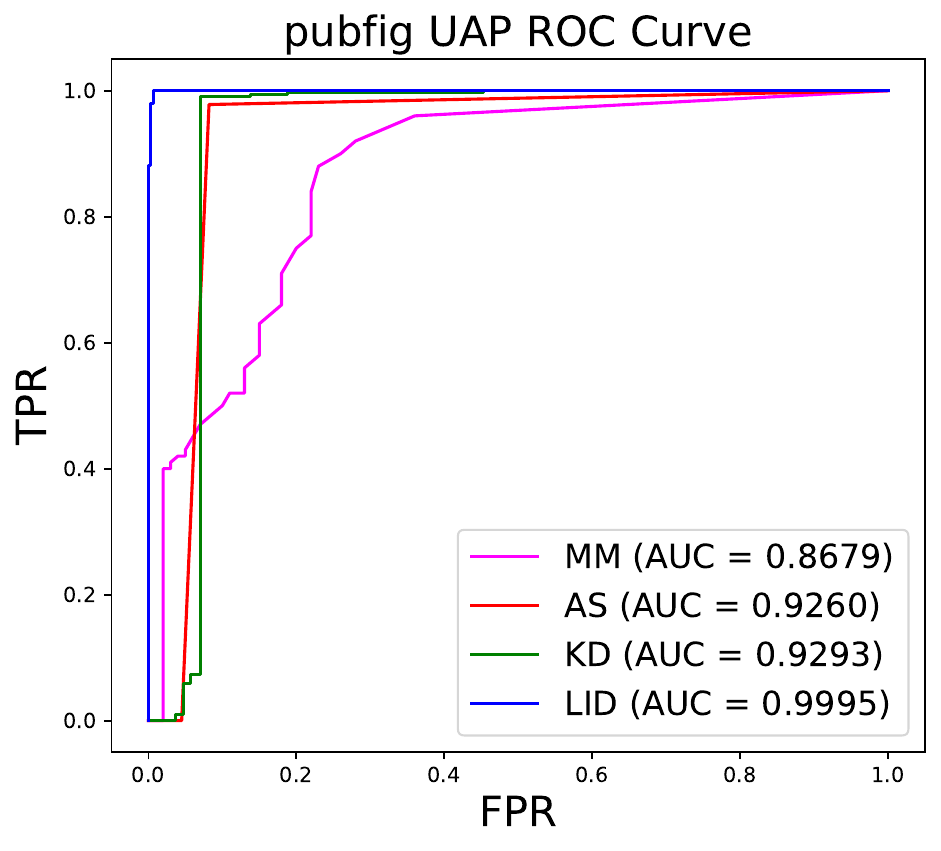}      
		\end{minipage}
	}
	\subfloat{
	\begin{minipage}[t]{0.25\linewidth}
		\label{PubFig-IABA-ROC}
		\centering
		\includegraphics[width=1\textwidth]{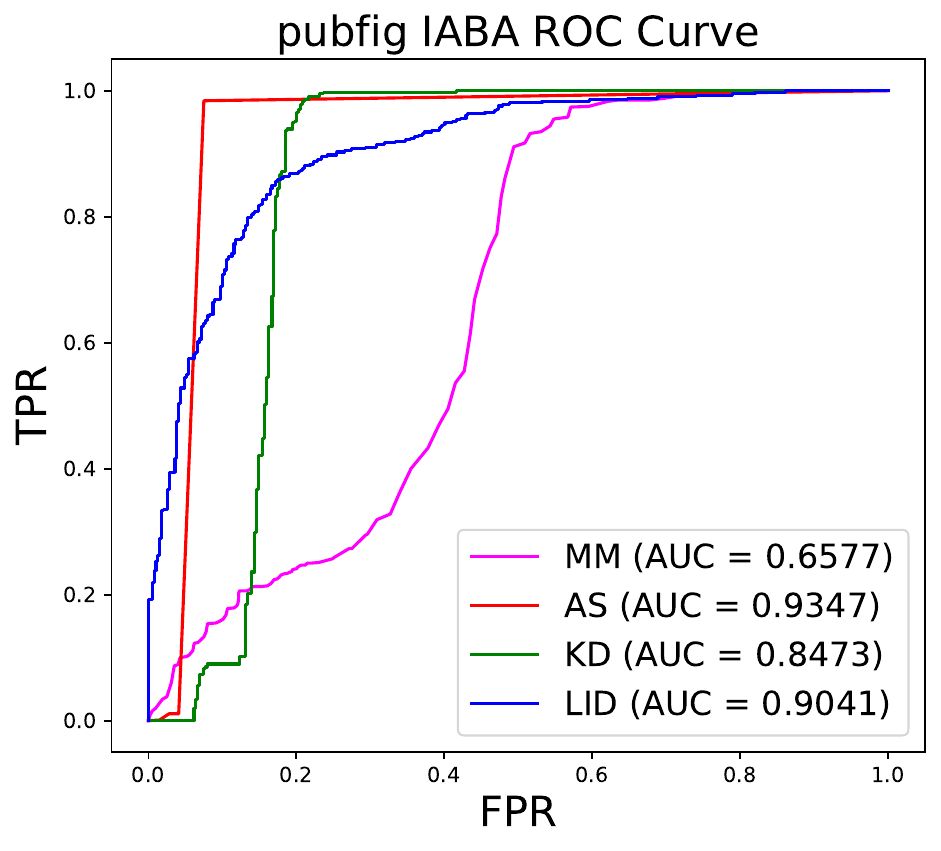}    
	\end{minipage}
	}
	\subfloat{
		\begin{minipage}[t]{0.25\linewidth}
			\label{PubFig-HTBA-ROC}
			\centering
			\includegraphics[width=1\textwidth]{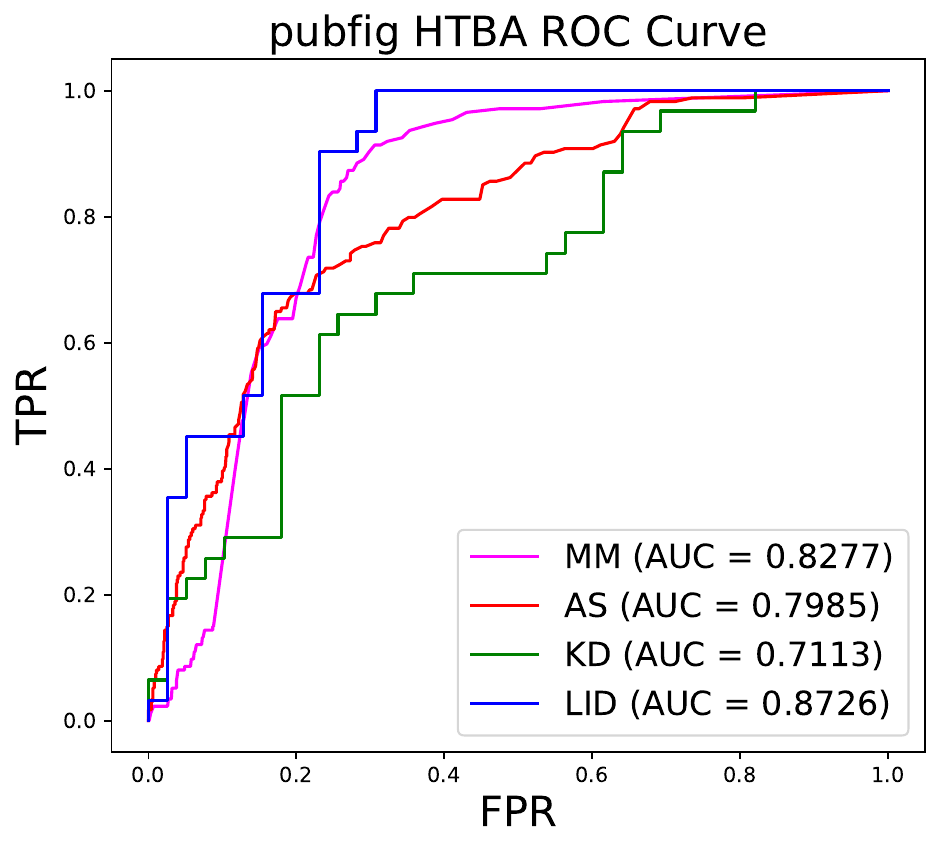}
		\end{minipage}
	}
	\vspace{-5pt}
	\caption{ROC curves for detecting UAP, IAB and HTB attacks.}
	\label{fig:INVI-AUC}    	
\end{figure*}

     We apply the considered detection approaches to those attacks. Table \ref{tab:invi-AUC} summarized the detection accuracy, and Figure \ref{fig:INVI-AUC} shows the corresponding ROC curves for each attack and solution. We observe that those approaches are still effective at detecting those advanced malicious samples. Particularly, LID gives the best performance compared to the other three.

\begin{table}[h]
  \centering
  \small
  \caption{AUC results for  UAP, IAB and HTB attacks.}
    \begin{tabular}{c c c c c c }
    \toprule
    Dataset & Attack & {MM} & {AS} & {KD} & {LID}\\
    \cline{1-6}
    \multirow{3}[0]{*}{MNIST} & UAP & \textbf{0.9890}   &  0.9423 & 0.9474  & 0.9714  \\
                      & IAB &   0.9572   &  0.9002     &    \textbf{1.00}   & \textbf{1.00}  \\
                      & HTB &   0.9897   &  0.7505     &    \textbf{1.00}  &  \textbf{1.00}  \\\cline{1-6}
    \multirow{3}[0]{*}{GTSRB} & UAP &  0.7809  &  0.9900 & 0.6610  & \textbf{0.9928} \\
                      & IAB &   0.8823   & 0.9612  & 0.9865 & \textbf{0.9935} \\
                      & HTB &   0.8850   &  0.9318     &    0.9886   & \textbf{0.9998} \\\cline{1-6}
    \multirow{3}[0]{*}{PubFig} & UAP &  0.8679  &  0.9260 & 0.9293  & \textbf{0.9995} \\
                      & IAB &   0.6577   & \textbf{0.9347}  & 0.8473 & 0.9041 \\
                      & HTB &   0.8277   &  0.7985     &    0.7113   & \textbf{0.8726}\\
                      \bottomrule
    \end{tabular}
  \label{tab:invi-AUC}%
\end{table}
	
\vspace{-3pt}
\begin{center}
	\noindent \fbox{\parbox{85mm}
		{\textbf{Remark 6}: \emph{The considered detection approaches are effective and general for more advanced attacks (e.g., universal perturbations or invisible triggers), as the malicious samples still have large differences from normal ones in the feature space. LID gives the best performance in particular.}}
	}
	
\end{center}

\section{Other Defenses}
In addition to the above four methods we have discussed and evaluated, we also test several other adversarial example detection algorithms in the backdoor scenario. They are relatively less effective, or in a lack of generality. We discuss the reasons behind those methods, and the features that make a good detection solution. 

	\begin{figure*}[!ht]
		\centering
		\subfloat{
			\begin{minipage}[t]{0.23\linewidth}
				\label{mnist_drop_2}
				\centering
				\includegraphics[width=1\textwidth]{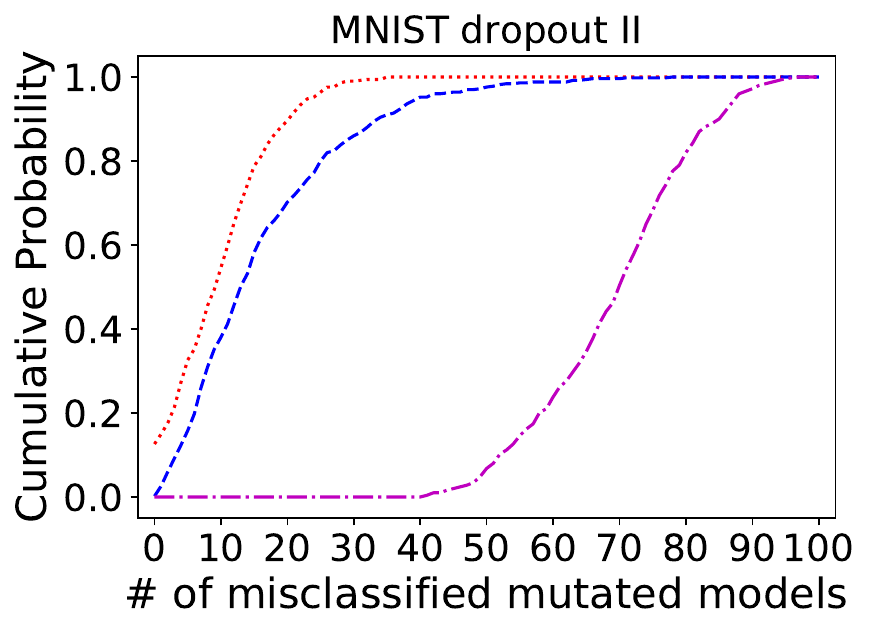}
			\end{minipage}
		}
		\subfloat{
			\begin{minipage}[t]{0.23\linewidth}
				\label{gtsrb_drop_2}
				\centering
				\includegraphics[width=1\textwidth]{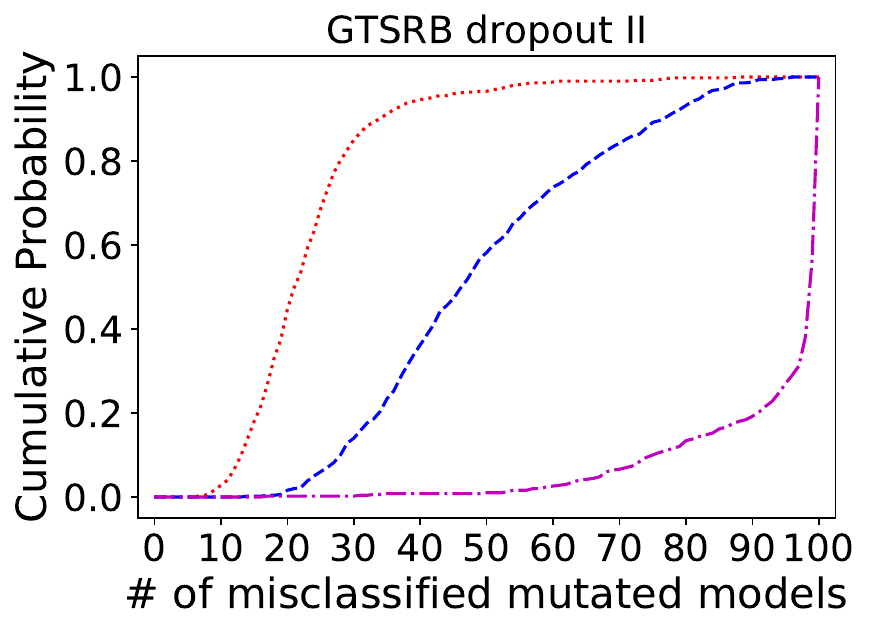}
			\end{minipage}
		}
		\subfloat{
			\begin{minipage}[t]{0.23\linewidth}
				\label{face_wm_drop_2}
				\centering
				\includegraphics[width=1\textwidth]{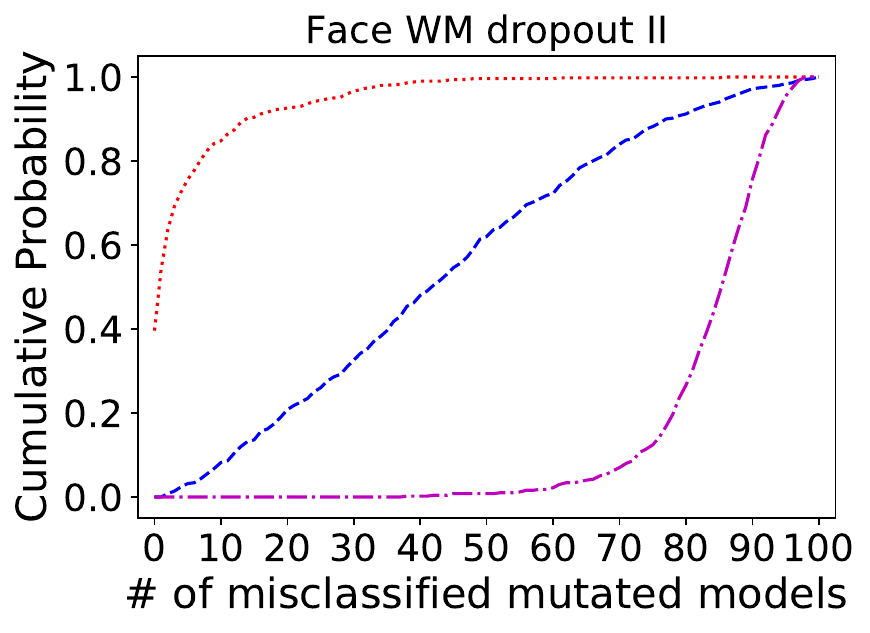}
			\end{minipage}
			
		}
		\subfloat{
			\begin{minipage}[t]{0.23\linewidth}
				\label{face_square_drop_2}
				\centering
				\includegraphics[width=1\textwidth]{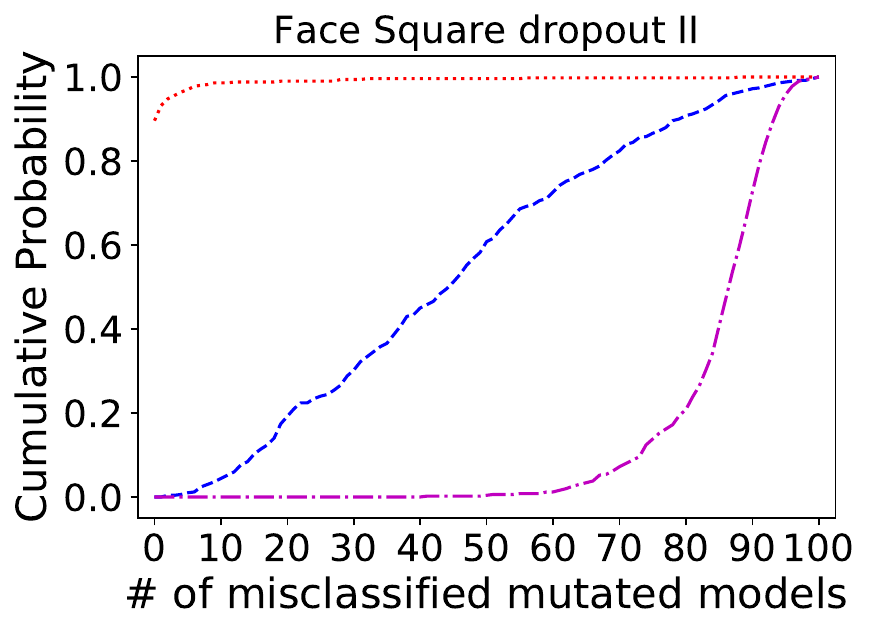}
			\end{minipage}
			
		}
		\hfill
		\subfloat{
			\begin{minipage}[t]{0.5\linewidth}
				\label{legend3}
				\centering
				\includegraphics[width=0.5\textwidth]{legend.png}
			\end{minipage}
		}
		\vspace{-5pt}
		\caption{Cumulative Distribution Function of three samples on the BU method. }
		\label{fig:BU_LCR}    
	\end{figure*}
	\begin{figure*}[!ht]
		\centering
		\subfloat{
			\begin{minipage}[t]{0.23\linewidth}
				\label{BU_mnist_backdoor_AUC}
				\centering
				\includegraphics[width=1\textwidth]{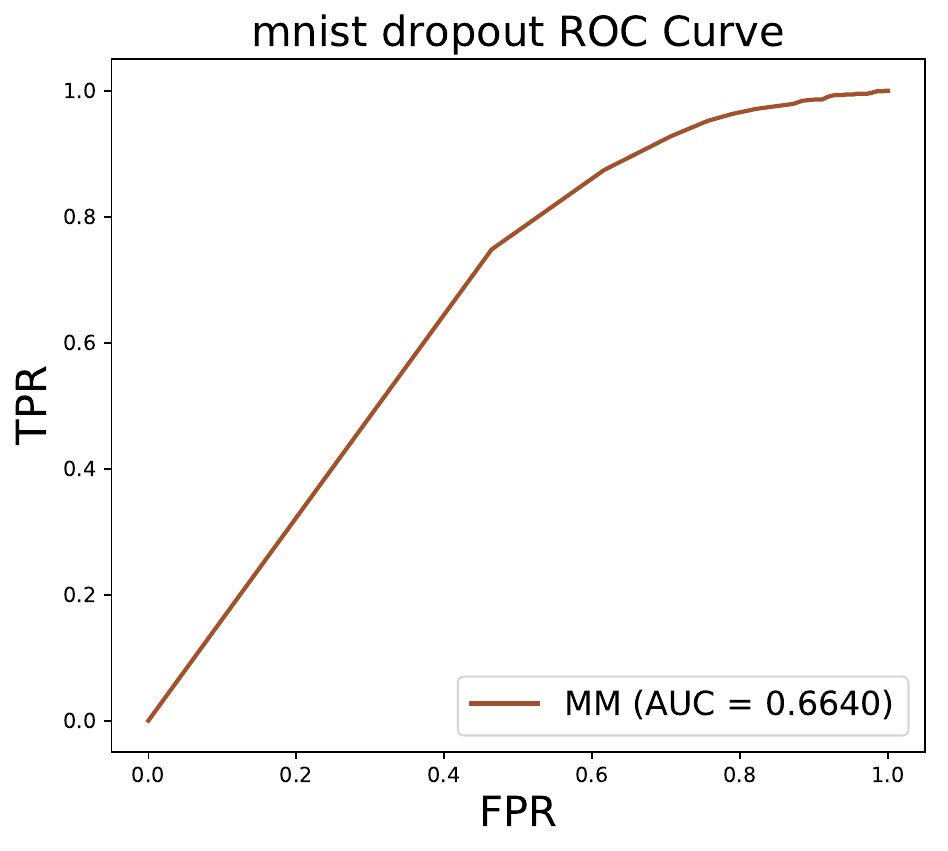}
			\end{minipage}
		}
		\subfloat{
			\begin{minipage}[t]{0.23\linewidth}
				\label{BU_gtsrb_backdoor_AUC}
				\centering
				\includegraphics[width=1\textwidth]{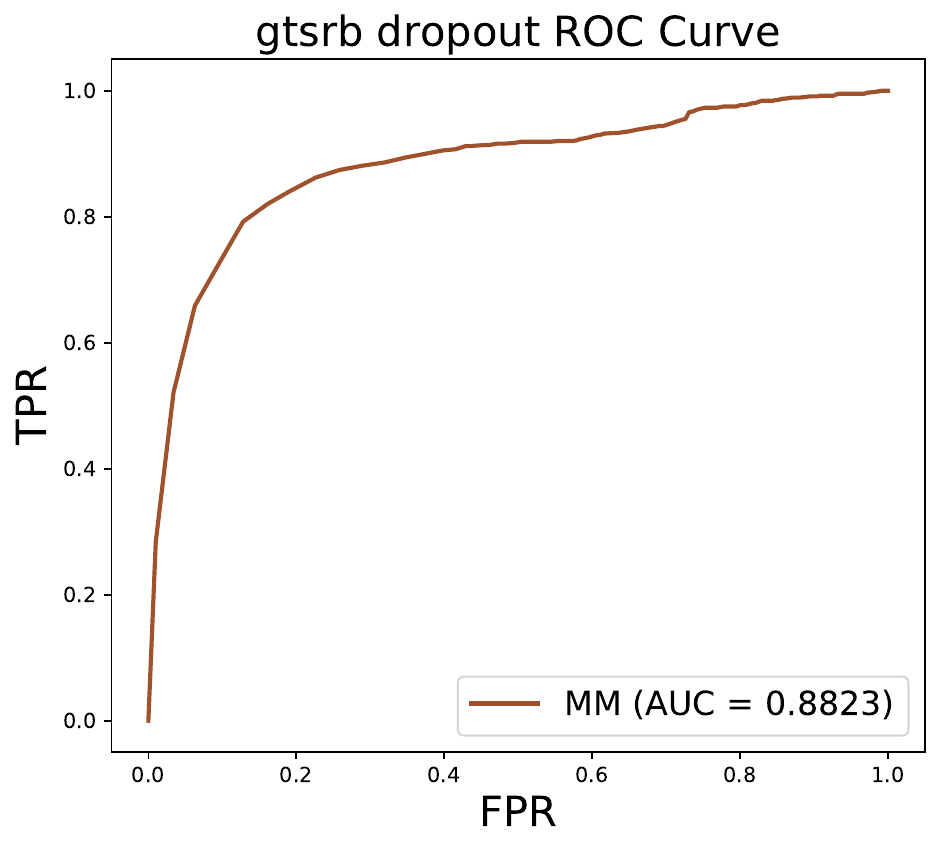}
			\end{minipage}		
		}
		\subfloat{
			\begin{minipage}[t]{0.23\linewidth}
				\label{BU_face_wm_backdoor_AUC}
				\centering
				\includegraphics[width=1\textwidth]{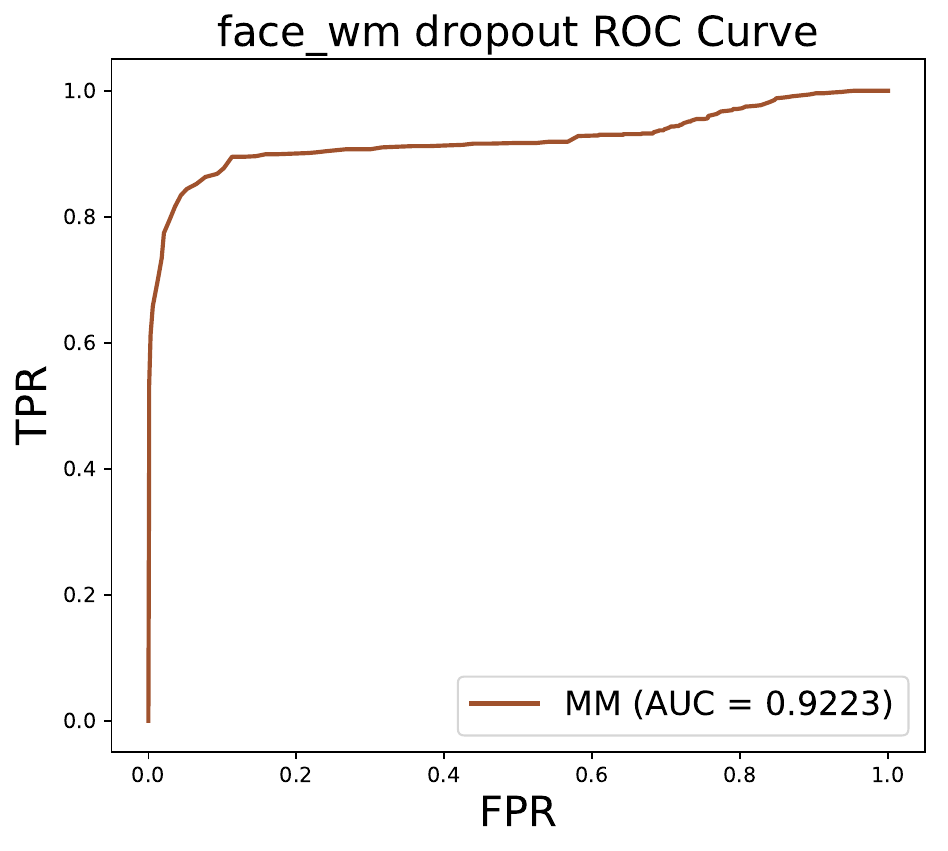}
			\end{minipage}
		}
		\subfloat{
			\begin{minipage}[t]{0.23\linewidth}
				\label{BU_face_square_backdoor_AUC}
				\centering
				\includegraphics[width=1\textwidth]{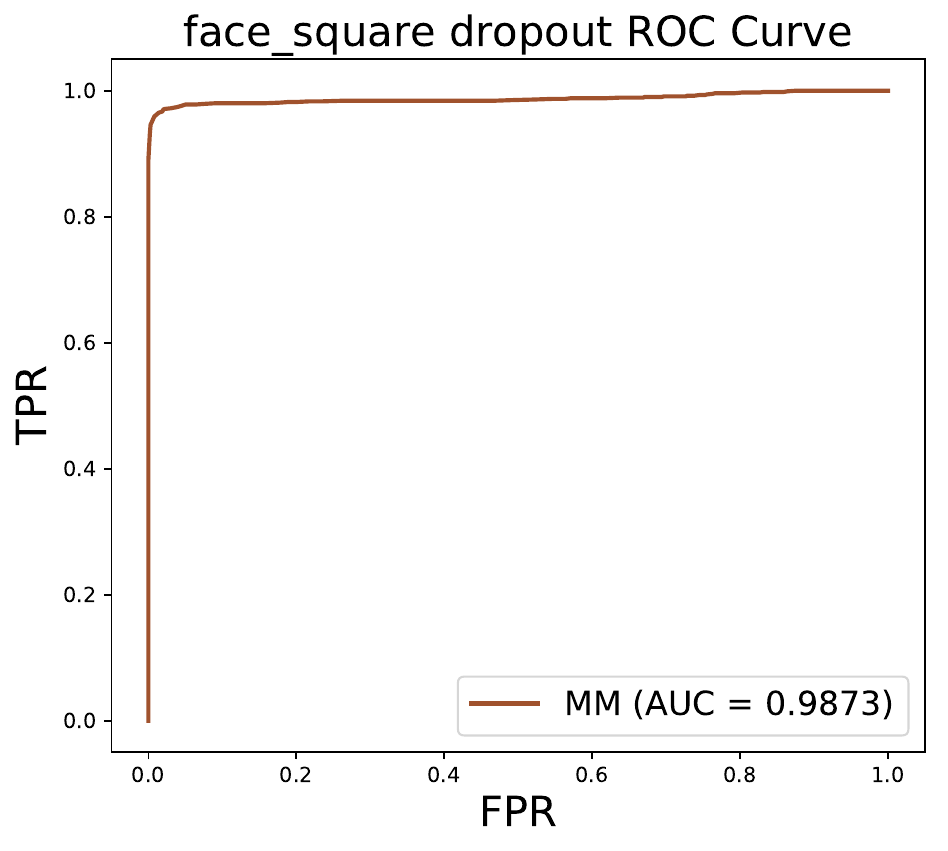}
			\end{minipage}		
		}
		\vspace{-5pt}
		\caption{ROC curve with BU method.}
		\label{fig:BU_auc_result}    	
		
	\end{figure*}
	
Bayesian Uncertainty estimates~\cite{detect_artifacts} is also based on the hypothesis that adversarial examples are are sensitive to model changes than normal samples, similar as the model mutation approach. Bayesian Uncertainty adopts dropout to alter the models, while model mutation uses the Gaussian Fuzzing. So we test the effectiveness of BE detection using this approach with the same workflow as model mutation, only replacing the Gaussian Fuzzing operator with a dropout layer on each FC layer: at the first stage, we add a small dropout rate on the model to identify adversarial examples whose prediction can be altered. At the second stage, we further increase the dropout rate 
to identify backdoor examples whose prediction is expected to be the same regardless of the dropout. Figure \ref{fig:BU_LCR} shows the cumulative probability distribution of different types of samples under Mutation \uppercase\expandafter{\romannumeral2}. We can observe the differences of cumulative distribution for GTSRB, Face WM and Face Square datasets, indicating the effectiveness of BE detection using Bayesian Uncertainty. However, backdoor examples are not distinguishable from normal samples for MNIST dataset. This is confirmed by the detection results in Figure \ref{fig:BU_auc_result}. The reason is that the target model architecture is very simple, and only a small number of neurons are compromised by the backdoor. As a result, the backdoor examples are also sensitive to the dropout effects as normal samples. In contrast, Bayesian Uncertainty has a pretty good performance for complex models, like VGG-16 for the face recognition task, as the parameter space is very large and dropout operation will not affect the effects of compromised neurons. 

\begin{center}
	\noindent \fbox{\parbox{85mm}
		{\textbf{Remark 7}: \emph{Bayesian Uncertainty Estimate with dropout can be used to detect backdoor examples in complicated models. It does not work well when the backdoor model is too simple.}}
	}    
\end{center}

	\begin{figure}[t]
		
		\subfloat{
			\begin{minipage}[t]{0.2\linewidth}
				\centering
				\includegraphics[width=1\textwidth]{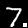}
			\end{minipage}
		}
		\hfill
		\subfloat{
			\begin{minipage}[t]{0.2\linewidth}
				\centering
				\includegraphics[width=1\textwidth]{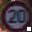}
			\end{minipage}
		}
		\hfill
		\subfloat{
			\begin{minipage}[t]{0.2\linewidth}
				\centering
				\includegraphics[width=1\textwidth]{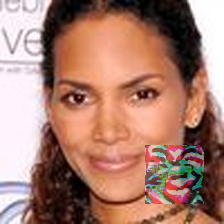}
			\end{minipage}
		}
		\hfill
		\subfloat{
			\begin{minipage}[t]{0.2\linewidth}
				\centering
				\includegraphics[width=1\textwidth]{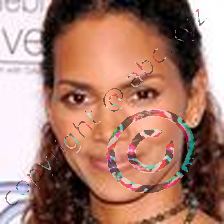}
			\end{minipage}
		}
		\hfill
		
		\centerline{Original Backdoor Images}
		
		\subfloat{
			\begin{minipage}[t]{0.2\linewidth}
				\centering
				\includegraphics[width=1\textwidth]{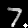}
			\end{minipage}
		}
		\hfill
		\subfloat{
			\begin{minipage}[t]{0.2\linewidth}
				\centering
				\includegraphics[width=1\textwidth]{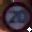}
			\end{minipage}
		}
		\hfill
		\subfloat{
			\begin{minipage}[t]{0.2\linewidth}
				\centering
				\includegraphics[width=1\textwidth]{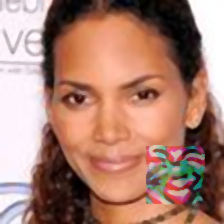}
			\end{minipage}
		}
		\hfill
		\subfloat{
			\begin{minipage}[t]{0.2\linewidth}
				\centering
				\includegraphics[width=1\textwidth]{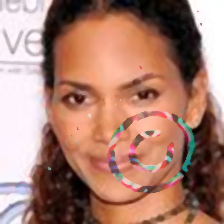}
			\end{minipage}
		}
		\hfill
		\centerline{Median Filter Images}
		\caption{Transformation with median filter. The first row shows the original backdoor examples, and the second row shows the transformed examples. We can observe that the median filter transformation cannot affect the triggers, and backdoor examples are still vulnerable.}
		\label{fig:squeezed_image}    
	\end{figure}

	
Region-Based classification~\cite{region_based} detects AEs based on the hypothesis that AEs are closer to the decision boundary, and most neighbour labels in the hypercube of AEs are the correct labels.  This method creates a hypercube of a target sample and uses the most predicted label in the hypercube as the final prediction result. Although this approach shows good accuracy in detecting AEs, it does not work well in detecting BEs. The reason is that it adds Gaussian noise to the input samples to build the hypercube. BEs with the trigger are much more robust against random noise than AEs. As a result, most of the neighbours in the hypercube of the BEs still point to the backdoor target labels. 

Feature Squeezing~\cite{FeatureSqueezing} measures the confidence distance from the target input and its squeezed input. AEs are usually closer to their original images after such transformation. Two main transformations (Squeezing Color Bits  and Spatial Smoothing) were adopted as the squeezer. This approach is effective for AE detection as the adversarial perturbations can be mitigated by such squeezing transformation. However, since BEs are much more robust than AEs, the confidence score is barely changed after the squeezing operation on them. Then Feature Squeezing fails to detect BEs with triggers. (Figure \ref{fig:squeezed_image} shows the BEs transformed with median filter). 

\begin{center}
	\noindent \fbox{\parbox{85mm}
		{\textbf{Remark 8}: \emph{Since BEs are more robust than AEs, input transformation based solutions generally fail to mitigate BEs, even they have been proved effective in defeating adversarial attacks.}}
	}
\end{center}

As we mentioned before, the machine learning and security communities focus on different types of adversarial attacks (e.g., adversarial examples, backdoor attacks) and their corresponding defense solutions. However, the connections between these threats are not well investigated, although they share certain similarities. There are only two works~\cite{pang2020tale,weng2020trade} exploring the relationships between adversarial and backdoor examples, from the perspective of attacks. We present the first study towards the defenses of these threats. We believe our work reveals the common features of adversarial and backdoor attacks, which can facilitate the design of defenses. In particular, our findings on Activation Space and Feature Space for detecting malicious examples can be effective for building secure deep learning applications and systems. The Activation Clustering method \cite{activation_clustering} for detecting poisoning data can also help to improve adversarial defenses such as adversarial training~\cite{wang2019improving,bai2021improving}.

\section{conclusion}
In this paper, we identify the connections between adversarial examples and backdoor examples in model sensitivity, feature space and activation space. Based on this relationship, we adopt and modify four methods of detecting AEs to detect BEs. Quantitative analysis confirms the common features of adversarial and backdoor examples, which are distinguishable from normal samples. Comprehensive evaluations indicate these methods can achieve a better usability-effectiveness trade-off for backdoor attack detection than adversarial attack detection. 

Although the connection between adversarial examples and backdoor attacks were preliminarily explored in \cite{pang2020tale,weng2020trade} from the attack behaviors, this paper presents the first study towards such connection from the perspective of detection. We identify eight remarks, which can shed light on the design of more advanced defense solutions against backdoor attacks. In the future, we will extend our work with the following three directions: (1) we will focus on unifying other detection methods, and other types of defenses (e.g., removing perturbation via input preprocessing, combining the activation space and feature space~\cite{li2021neural,bai2021improving}). (2) We will adopt the ensemble of multiple detection approaches for better accuracy. (3) We will also analyze and interpret the connection and unification of adversarial and backdoor examples in a theoretical way. (4) In this paper, we only evaluate the state-of-the-art backdoor attacks. In the future, we will consider adaptive attacks which can try to make them stealthy in the feature space and activation space to bypass our detectors.


\bibliographystyle{unsrt}  
\bibliography{references}

\begin{thebibliography}{10}

\bibitem{image_class}
Kaiming He, Xiangyu Zhang, Shaoqing Ren, and Jian Sun.
\newblock Deep residual learning for image recognition.
\newblock In {\em Proceedings of the IEEE conference on computer vision and
  pattern recognition}, pages 770--778, 2016.

\bibitem{face_recognition}
Alex Krizhevsky, Ilya Sutskever, and Geoffrey~E Hinton.
\newblock Imagenet classification with deep convolutional neural networks.
\newblock In {\em Advances in neural information processing systems}, pages
  1097--1105, 2012.

\bibitem{speech_recognition}
Wayne Xiong, Jasha Droppo, Xuedong Huang, Frank Seide, Mike Seltzer, Andreas
  Stolcke, Dong Yu, and Geoffrey Zweig.
\newblock Achieving human parity in conversational speech recognition.
\newblock {\em arXiv preprint arXiv:1610.05256}, 2016.

\bibitem{luong2015effective}
Minh-Thang Luong, Hieu Pham, and Christopher~D Manning.
\newblock Effective approaches to attention-based neural machine translation.
\newblock {\em arXiv preprint arXiv:1508.04025}, 2015.

\bibitem{adver_attack}
Christian Szegedy, Wojciech Zaremba, Ilya Sutskever, Joan Bruna, Dumitru Erhan,
  Ian Goodfellow, and Rob Fergus.
\newblock Intriguing properties of neural networks.
\newblock {\em arXiv preprint arXiv:1312.6199}, 2013.

\bibitem{badnets}
Tianyu Gu, Brendan Dolan-Gavitt, and Siddharth Garg.
\newblock Badnets: Identifying vulnerabilities in the machine learning model
  supply chain.
\newblock {\em arXiv preprint arXiv:1708.06733}, 2017.

\bibitem{self_drive}
Mariusz Bojarski, Davide Del~Testa, Daniel Dworakowski, Bernhard Firner, Beat
  Flepp, Prasoon Goyal, Lawrence~D Jackel, Mathew Monfort, Urs Muller, Jiakai
  Zhang, et~al.
\newblock End to end learning for self-driving cars.
\newblock {\em arXiv preprint arXiv:1604.07316}, 2016.

\bibitem{malware_detection}
Qinglong Wang, Wenbo Guo, Kaixuan Zhang, Alexander~G Ororbia, Xinyu Xing, Xue
  Liu, and C~Lee Giles.
\newblock Adversary resistant deep neural networks with an application to
  malware detection.
\newblock In {\em Proceedings of the 23rd ACM SIGKDD International Conference
  on Knowledge Discovery and Data Mining}, pages 1145--1153, 2017.

\bibitem{fan2018android}
Ming Fan, Jun Liu, Xiapu Luo, Kai Chen, Zhenzhou Tian, Qinghua Zheng, and Ting
  Liu.
\newblock Android malware familial classification and representative sample
  selection via frequent subgraph analysis.
\newblock {\em IEEE Transactions on Information Forensics and Security},
  13(8):1890--1905, 2018.

\bibitem{fan2019ctdroid}
Ming Fan, Xiapu Luo, Jun Liu, Chunyin Nong, Qinghua Zheng, and Ting Liu.
\newblock Ctdroid: leveraging a corpus of technical blogs for android malware
  analysis.
\newblock {\em IEEE Transactions on Reliability}, 69(1):124--138, 2019.

\bibitem{fan2016frequent}
Ming Fan, Jun Liu, Xiapu Luo, Kai Chen, Tianyi Chen, Zhenzhou Tian, Xiaodong
  Zhang, Qinghua Zheng, and Ting Liu.
\newblock Frequent subgraph based familial classification of android malware.
\newblock In {\em 2016 IEEE 27th International Symposium on Software
  Reliability Engineering (ISSRE)}, pages 24--35. IEEE, 2016.

\bibitem{fan2019graph}
Ming Fan, Xiapu Luo, Jun Liu, Meng Wang, Chunyin Nong, Qinghua Zheng, and Ting
  Liu.
\newblock Graph embedding based familial analysis of android malware using
  unsupervised learning.
\newblock In {\em 2019 IEEE/ACM 41st International Conference on Software
  Engineering (ICSE)}, pages 771--782. IEEE, 2019.

\bibitem{chen2017targeted}
Xinyun Chen, Chang Liu, Bo~Li, Kimberly Lu, and Dawn Song.
\newblock Targeted backdoor attacks on deep learning systems using data
  poisoning.
\newblock {\em arXiv preprint arXiv:1712.05526}, 2017.

\bibitem{finlayson2018adversarial}
Samuel~G Finlayson, Hyung~Won Chung, Isaac~S Kohane, and Andrew~L Beam.
\newblock Adversarial attacks against medical deep learning systems.
\newblock {\em arXiv preprint arXiv:1804.05296}, 2018.

\bibitem{adver_train}
Uri Shaham, Yutaro Yamada, and Sahand Negahban.
\newblock Understanding adversarial training: Increasing local stability of
  supervised models through robust optimization.
\newblock {\em Neurocomputing}, 307:195--204, 2018.

\bibitem{huang2015learning}
Ruitong Huang, Bing Xu, Dale Schuurmans, and Csaba Szepesv{\'a}ri.
\newblock Learning with a strong adversary.
\newblock {\em arXiv preprint arXiv:1511.03034}, 2015.

\bibitem{gu2014towards}
Shixiang Gu and Luca Rigazio.
\newblock Towards deep neural network architectures robust to adversarial
  examples.
\newblock {\em arXiv preprint arXiv:1412.5068}, 2014.

\bibitem{ross2018improving}
Andrew~Slavin Ross and Finale Doshi-Velez.
\newblock Improving the adversarial robustness and interpretability of deep
  neural networks by regularizing their input gradients.
\newblock In {\em Thirty-second AAAI conference on artificial intelligence},
  2018.

\bibitem{defen_distillation}
Nicolas Papernot, Patrick McDaniel, Xi~Wu, Somesh Jha, and Ananthram Swami.
\newblock Distillation as a defense to adversarial perturbations against deep
  neural networks.
\newblock In {\em 2016 IEEE Symposium on Security and Privacy (SP)}, pages
  582--597. IEEE, 2016.

\bibitem{Magnet}
Dongyu Meng and Hao Chen.
\newblock Magnet: a two-pronged defense against adversarial examples.
\newblock In {\em Proceedings of the 2017 ACM SIGSAC Conference on Computer and
  Communications Security}, pages 135--147, 2017.

\bibitem{lee2017generative}
Hyeungill Lee, Sungyeob Han, and Jungwoo Lee.
\newblock Generative adversarial trainer: Defense to adversarial perturbations
  with gan.
\newblock {\em arXiv preprint arXiv:1705.03387}, 2017.

\bibitem{guo2018countering}
Chuan Guo, Mayank Rana, Moustapha Cisse, and Laurens van~der Maaten.
\newblock Countering adversarial images using input transformations.
\newblock In {\em International Conference on Learning Representations}, 2018.

\bibitem{prakash2018deflecting}
Aaditya Prakash, Nick Moran, Solomon Garber, Antonella DiLillo, and James
  Storer.
\newblock Deflecting adversarial attacks with pixel deflection.
\newblock In {\em Proceedings of the IEEE conference on computer vision and
  pattern recognition}, pages 8571--8580, 2018.

\bibitem{xie2018mitigating}
Cihang Xie, Jianyu Wang, Zhishuai Zhang, Zhou Ren, and Alan Yuille.
\newblock Mitigating adversarial effects through randomization.
\newblock In {\em International Conference on Learning Representations}, 2018.

\bibitem{buckman2018thermometer}
Jacob Buckman, Aurko Roy, Colin Raffel, and Ian Goodfellow.
\newblock Thermometer encoding: One hot way to resist adversarial examples.
\newblock In {\em International Conference on Learning Representations}, 2018.

\bibitem{das2018shield}
Nilaksh Das, Madhuri Shanbhogue, Shang-Tse Chen, Fred Hohman, Siwei Li,
  Li~Chen, Michael~E Kounavis, and Duen~Horng Chau.
\newblock Shield: Fast, practical defense and vaccination for deep learning
  using jpeg compression.
\newblock In {\em Proceedings of the 24th ACM SIGKDD International Conference
  on Knowledge Discovery \& Data Mining}, pages 196--204, 2018.

\bibitem{Neural_cleanse}
Bolun Wang, Yuanshun Yao, Shawn Shan, Huiying Li, Bimal Viswanath, Haitao
  Zheng, and Ben~Y Zhao.
\newblock Neural cleanse: Identifying and mitigating backdoor attacks in neural
  networks.
\newblock In {\em 2019 IEEE Symposium on Security and Privacy (SP)}, pages
  707--723. IEEE, 2019.

\bibitem{Deepinspect}
Huili Chen, Cheng Fu, Jishen Zhao, and Farinaz Koushanfar.
\newblock Deepinspect: A black-box trojan detection and mitigation framework
  for deep neural networks.
\newblock In {\em Proceedings of the 28th International Joint Conference on
  Artificial Intelligence. AAAI Press}, pages 4658--4664, 2019.

\bibitem{defen_Tabor}
Wenbo Guo, Lun Wang, Xinyu Xing, Min Du, and Dawn Song.
\newblock Tabor: A highly accurate approach to inspecting and restoring trojan
  backdoors in ai systems.
\newblock {\em arXiv preprint arXiv:1908.01763}, 2019.

\bibitem{defen_GDM}
Ximing Qiao, Yukun Yang, and Hai Li.
\newblock Defending neural backdoors via generative distribution modeling.
\newblock In {\em Advances in Neural Information Processing Systems}, pages
  14004--14013, 2019.

\bibitem{defen_MNA}
Xiaojun Xu, Qi~Wang, Huichen Li, Nikita Borisov, Carl~A Gunter, and Bo~Li.
\newblock Detecting ai trojans using meta neural analysis.
\newblock {\em arXiv preprint arXiv:1910.03137}, 2019.

\bibitem{ABS}
Yingqi Liu, Wen-Chuan Lee, Guanhong Tao, Shiqing Ma, Yousra Aafer, and Xiangyu
  Zhang.
\newblock Abs: Scanning neural networks for back-doors by artificial brain
  stimulation.
\newblock In {\em Proceedings of the 2019 ACM SIGSAC Conference on Computer and
  Communications Security}, pages 1265--1282, 2019.

\bibitem{tan2019bypassing}
Te~Juin~Lester Tan and Reza Shokri.
\newblock Bypassing backdoor detection algorithms in deep learning.
\newblock {\em arXiv preprint arXiv:1905.13409}, 2019.

\bibitem{activation_clustering}
Bryant Chen, Wilka Carvalho, Nathalie Baracaldo, Heiko Ludwig, Benjamin
  Edwards, Taesung Lee, Ian Molloy, and Biplav Srivastava.
\newblock Detecting backdoor attacks on deep neural networks by activation
  clustering.
\newblock {\em arXiv preprint arXiv:1811.03728}, 2018.

\bibitem{Spectral_signatures}
Brandon Tran, Jerry Li, and Aleksander Madry.
\newblock Spectral signatures in backdoor attacks.
\newblock In {\em Advances in Neural Information Processing Systems}, pages
  8000--8010, 2018.

\bibitem{Differential_Privacy}
Min Du, Ruoxi Jia, and Dawn Song.
\newblock Robust anomaly detection and backdoor attack detection via
  differential privacy.
\newblock {\em arXiv preprint arXiv:1911.07116}, 2019.

\bibitem{Strip}
Yansong Gao, Change Xu, Derui Wang, Shiping Chen, Damith~C Ranasinghe, and
  Surya Nepal.
\newblock Strip: A defence against trojan attacks on deep neural networks.
\newblock In {\em Proceedings of the 35th Annual Computer Security Applications
  Conference}, pages 113--125, 2019.

\bibitem{Sentinet}
Edward Chou, Florian Tram{\`e}r, Giancarlo Pellegrino, and Dan Boneh.
\newblock Sentinet: Detecting physical attacks against deep learning systems.
\newblock {\em arXiv preprint arXiv:1812.00292}, 2018.

\bibitem{Grad_cam}
Ramprasaath~R Selvaraju, Michael Cogswell, Abhishek Das, Ramakrishna Vedantam,
  Devi Parikh, and Dhruv Batra.
\newblock Grad-cam: Visual explanations from deep networks via gradient-based
  localization.
\newblock In {\em Proceedings of the IEEE international conference on computer
  vision}, pages 618--626, 2017.

\bibitem{pang2020tale}
Ren Pang, Hua Shen, Xinyang Zhang, Shouling Ji, Yevgeniy Vorobeychik, Xiapu
  Luo, Alex Liu, and Ting Wang.
\newblock A tale of evil twins: Adversarial inputs versus poisoned models.
\newblock In {\em Proceedings of the 2020 ACM SIGSAC Conference on Computer and
  Communications Security}, pages 85--99, 2020.

\bibitem{weng2020trade}
Cheng-Hsin Weng, Yan-Ting Lee, and Shan-Hung~Brandon Wu.
\newblock On the trade-off between adversarial and backdoor robustness.
\newblock {\em Advances in Neural Information Processing Systems}, 33, 2020.

\bibitem{biggio2013evasion}
Battista Biggio, Igino Corona, Davide Maiorca, Blaine Nelson, Nedim
  {\v{S}}rndi{\'c}, Pavel Laskov, Giorgio Giacinto, and Fabio Roli.
\newblock Evasion attacks against machine learning at test time.
\newblock In {\em Joint European conference on machine learning and knowledge
  discovery in databases}, pages 387--402. Springer, 2013.

\bibitem{FGSM}
Ian~J Goodfellow, Jonathon Shlens, and Christian Szegedy.
\newblock Explaining and harnessing adversarial examples.
\newblock {\em arXiv preprint arXiv:1412.6572}, 2014.

\bibitem{kurakin2016adversarial}
Alexey Kurakin, Ian Goodfellow, and Samy Bengio.
\newblock Adversarial examples in the physical world.
\newblock {\em arXiv preprint arXiv:1607.02533}, 2016.

\bibitem{dong2017discovering}
Yinpeng Dong, Fangzhou Liao, Tianyu Pang, Xiaolin Hu, and Jun Zhu.
\newblock Discovering adversarial examples with momentum.
\newblock {\em arXiv preprint arXiv:1710.06081}, 2017.

\bibitem{BIM}
Alexey Kurakin, Ian Goodfellow, and Samy Bengio.
\newblock Adversarial examples in the physical world.
\newblock {\em arXiv preprint arXiv:1607.02533}, 2016.

\bibitem{moosavi2016deepfool}
Seyed-Mohsen Moosavi-Dezfooli, Alhussein Fawzi, and Pascal Frossard.
\newblock Deepfool: a simple and accurate method to fool deep neural networks.
\newblock In {\em Proceedings of the IEEE conference on computer vision and
  pattern recognition}, pages 2574--2582, 2016.

\bibitem{JSMA}
Nicolas Papernot, Patrick McDaniel, Somesh Jha, Matt Fredrikson, Z~Berkay
  Celik, and Ananthram Swami.
\newblock The limitations of deep learning in adversarial settings.
\newblock In {\em 2016 IEEE European symposium on security and privacy
  (EuroS\&amp;P)}, pages 372--387. IEEE, 2016.

\bibitem{su2019one}
Jiawei Su, Danilo~Vasconcellos Vargas, and Kouichi Sakurai.
\newblock One pixel attack for fooling deep neural networks.
\newblock {\em IEEE Transactions on Evolutionary Computation}, 23(5):828--841,
  2019.

\bibitem{CW_attack}
Nicholas Carlini and David Wagner.
\newblock Towards evaluating the robustness of neural networks.
\newblock In {\em 2017 ieee symposium on security and privacy (sp)}, pages
  39--57. IEEE, 2017.

\bibitem{Trojan_attack}
Yingqi Liu, Shiqing Ma, Yousra Aafer, Wen-Chuan Lee, Juan Zhai, Weihang Wang,
  and Xiangyu Zhang.
\newblock Trojaning attack on neural networks.
\newblock In {\em 25nd Annual Network and Distributed System Security
  Symposium, {NDSS} 2018, San Diego, California, USA, February 18-221, 2018}.
  The Internet Society, 2018.

\bibitem{Latent_attack}
Yuanshun Yao, Huiying Li, Haitao Zheng, and Ben~Y Zhao.
\newblock Latent backdoor attacks on deep neural networks.
\newblock In {\em Proceedings of the 2019 ACM SIGSAC Conference on Computer and
  Communications Security}, pages 2041--2055, 2019.

\bibitem{liu2020reflection}
Yunfei Liu, Xingjun Ma, James Bailey, and Feng Lu.
\newblock Reflection backdoor: A natural backdoor attack on deep neural
  networks.
\newblock In {\em European Conference on Computer Vision}, pages 182--199.
  Springer, 2020.

\bibitem{duan2020adversarial}
Ranjie Duan, Xingjun Ma, Yisen Wang, James Bailey, A~Kai Qin, and Yun Yang.
\newblock Adversarial camouflage: Hiding physical-world attacks with natural
  styles.
\newblock In {\em Proceedings of the IEEE/CVF Conference on Computer Vision and
  Pattern Recognition}, pages 1000--1008, 2020.

\bibitem{hosseini2018semantic}
Hossein Hosseini and Radha Poovendran.
\newblock Semantic adversarial examples.
\newblock In {\em Proceedings of the IEEE Conference on Computer Vision and
  Pattern Recognition Workshops}, pages 1614--1619, 2018.

\bibitem{bhattad2019unrestricted}
Anand Bhattad, Min~Jin Chong, Kaizhao Liang, Bo~Li, and David~A Forsyth.
\newblock Unrestricted adversarial examples via semantic manipulation.
\newblock {\em arXiv preprint arXiv:1904.06347}, 2019.

\bibitem{moosavi2017universal}
Seyed-Mohsen Moosavi-Dezfooli, Alhussein Fawzi, Omar Fawzi, and Pascal
  Frossard.
\newblock Universal adversarial perturbations.
\newblock In {\em Proceedings of the IEEE conference on computer vision and
  pattern recognition}, pages 1765--1773, 2017.

\bibitem{nguyen2020input}
Anh Nguyen and Anh Tran.
\newblock Input-aware dynamic backdoor attack.
\newblock {\em arXiv preprint arXiv:2010.08138}, 2020.

\bibitem{detect_artifacts}
Reuben Feinman, Ryan~R Curtin, Saurabh Shintre, and Andrew~B Gardner.
\newblock Detecting adversarial samples from artifacts.
\newblock {\em arXiv preprint arXiv:1703.00410}, 2017.

\bibitem{detect_LID}
Xingjun Ma, Bo~Li, Yisen Wang, Sarah~M Erfani, Sudanthi Wijewickrema, Grant
  Schoenebeck, Dawn Song, Michael~E Houle, and James Bailey.
\newblock Characterizing adversarial subspaces using local intrinsic
  dimensionality.
\newblock {\em arXiv preprint arXiv:1801.02613}, 2018.

\bibitem{activation_spaces}
Ziv Katzir and Yuval Elovici.
\newblock Detecting adversarial perturbations through spatial behavior in
  activation spaces.
\newblock In {\em 2019 International Joint Conference on Neural Networks
  (IJCNN)}, pages 1--9. IEEE, 2019.

\bibitem{input_transformation}
Shixin Tian, Guolei Yang, and Ying Cai.
\newblock Detecting adversarial examples through image transformation.
\newblock In {\em Thirty-Second AAAI Conference on Artificial Intelligence},
  2018.

\bibitem{FeatureSqueezing}
Weilin Xu, David Evans, and Yanjun Qi.
\newblock Feature squeezing: Detecting adversarial examples in deep neural
  networks.
\newblock In {\em Network and Distributed System Security Symposium}, 2018.

\bibitem{region_based}
Xiaoyu Cao and Neil~Zhenqiang Gong.
\newblock Mitigating evasion attacks to deep neural networks via region-based
  classification.
\newblock In {\em Proceedings of the 33rd Annual Computer Security Applications
  Conference}, pages 278--287, 2017.

\bibitem{model_mutation}
Jingyi Wang, Guoliang Dong, Jun Sun, Xinyu Wang, and Peixin Zhang.
\newblock Adversarial sample detection for deep neural network through model
  mutation testing.
\newblock In {\em 2019 IEEE/ACM 41st International Conference on Software
  Engineering (ICSE)}, pages 1245--1256. IEEE, 2019.

\bibitem{Fine_pruning}
Kang Liu, Brendan Dolan-Gavitt, and Siddharth Garg.
\newblock Fine-pruning: Defending against backdooring attacks on deep neural
  networks.
\newblock In {\em International Symposium on Research in Attacks, Intrusions,
  and Defenses}, pages 273--294. Springer, 2018.

\bibitem{ling2019deepsec}
Xiang Ling, Shouling Ji, Jiaxu Zou, Jiannan Wang, Chunming Wu, Bo~Li, and Ting
  Wang.
\newblock Deepsec: A uniform platform for security analysis of deep learning
  model.
\newblock In {\em 2019 IEEE Symposium on Security and Privacy (SP)}, pages
  673--690. IEEE, 2019.

\bibitem{ibm-adv}
IBM.
\newblock Adversarial robustness toolbox (art) v1.2.
\newblock \url{https://github.com/IBM/adversarial-robustness-toolbox}.

\bibitem{MNIST_DATA}
Yann LeCun, L{\'e}on Bottou, Yoshua Bengio, and Patrick Haffner.
\newblock Gradient-based learning applied to document recognition.
\newblock {\em Proceedings of the IEEE}, 86(11):2278--2324, 1998.

\bibitem{GTSRB_DATA}
Johannes Stallkamp, Marc Schlipsing, Jan Salmen, and Christian Igel.
\newblock Man vs. computer: Benchmarking machine learning algorithms for
  traffic sign recognition.
\newblock {\em Neural networks}, 32:323--332, 2012.

\bibitem{PubFig_data}
Nicolas Pinto, Zak Stone, Todd Zickler, and David Cox.
\newblock Scaling up biologically-inspired computer vision: A case study in
  unconstrained face recognition on facebook.
\newblock In {\em CVPR 2011 WORKSHOPS}, pages 35--42. IEEE, 2011.

\bibitem{papernot2018cleverhans}
Nicolas Papernot, Fartash Faghri, Nicholas Carlini, Ian Goodfellow, Reuben
  Feinman, Alexey Kurakin, Cihang Xie, Yash Sharma, Tom Brown, Aurko Roy,
  Alexander Matyasko, Vahid Behzadan, Karen Hambardzumyan, Zhishuai Zhang,
  Yi-Lin Juang, Zhi Li, Ryan Sheatsley, Abhibhav Garg, Jonathan Uesato, Willi
  Gierke, Yinpeng Dong, David Berthelot, Paul Hendricks, Jonas Rauber, and
  Rujun Long.
\newblock Technical report on the cleverhans v2.1.0 adversarial examples
  library.
\newblock {\em arXiv preprint arXiv:1610.00768}, 2018.

\bibitem{saha2020hidden}
Aniruddha Saha, Akshayvarun Subramanya, and Hamed Pirsiavash.
\newblock Hidden trigger backdoor attacks.
\newblock In {\em Proceedings of the AAAI Conference on Artificial
  Intelligence}, volume~34, pages 11957--11965, 2020.

\bibitem{wang2019improving}
Yisen Wang, Difan Zou, Jinfeng Yi, James Bailey, Xingjun Ma, and Quanquan Gu.
\newblock Improving adversarial robustness requires revisiting misclassified
  examples.
\newblock In {\em International Conference on Learning Representations}, 2019.

\bibitem{bai2021improving}
Yang Bai, Yuyuan Zeng, Yong Jiang, Shu-Tao Xia, Xingjun Ma, and Yisen Wang.
\newblock Improving adversarial robustness via channel-wise activation
  suppressing.
\newblock {\em arXiv preprint arXiv:2103.08307}, 2021.

\bibitem{li2021neural}
Yige Li, Xixiang Lyu, Nodens Koren, Lingjuan Lyu, Bo~Li, and Xingjun Ma.
\newblock Neural attention distillation: Erasing backdoor triggers from deep
  neural networks.
\newblock {\em arXiv preprint arXiv:2101.05930}, 2021.

\end{thebibliography}

\end{document}